\pdfoutput=1

\documentclass[11pt]{article}

\usepackage{acl}

\usepackage{times}
\usepackage{makecell}
\usepackage{tabularx}
\usepackage{latexsym}
\usepackage{arydshln}
\usepackage{amsthm}

\usepackage{subcaption}
\usepackage{soul}
\DeclareRobustCommand{\hlpink}[1]{{\sethlcolor{pink}\hl{#1}}}
\DeclareRobustCommand{\hlgreen}[1]{{\sethlcolor{pink}\hl{#1}}}
\definecolor{delectricblue}{RGB}{93, 117, 131}
\colorlet{lightdelectricblue}{delectricblue!30}

\newcommand{\hldb}[1]{%
    {%
    \sethlcolor{lightdelectricblue}%
    \hl{#1}%
    }%
}
\definecolor{lightblue}{rgb}{.90,.95,1}
\sethlcolor{lightblue}
\usepackage{xcolor}
\usepackage{booktabs}
\usepackage{tcolorbox}
\newtheorem*{definition}{Definition}
\usepackage{arydshln}
\usepackage{color,soul}
\newcommand{\greenup}[1]{\textcolor{red}{(+#1)}}
\newcommand{\reddown}[1]{\textcolor{blue}{(-#1)}}

\usepackage[T1]{fontenc}

\usepackage[utf8]{inputenc}
\usepackage{multirow}
\usepackage{microtype}

\usepackage{inconsolata}

\usepackage{graphicx}
\newtcolorbox{promptbox}{
  colback=white!20,
  colframe=gray!50!black,
  boxsep=10pt,
  left=1pt, right=1pt, top=1pt, bottom=1pt, boxrule=0.5pt
}
\newtcolorbox{mybox}[1]{colback=white!20,colframe=gray!50!black,fonttitle=\bfseries,title=#1}
\title{\textsc{LazyReview}\\ A Dataset for Uncovering Lazy Thinking in NLP Peer Reviews}

\author{ Sukannya Purkayastha$^1$, Zhuang Li$^2$, Anne Lauscher$^3$, Lizhen Qu$^4$, Iryna Gurevych$^1$ \\
         $^1$ Ubiquitous Knowledge Processing Lab, \\ 
         Department of Computer Science and Hessian Center for AI (hessian.AI), \\
         Technical University of Darmstadt \\ 
         $^2$ School of Computing Technologies, Royal Melbourne Institute of Technology, Australia \\
         $^3$ Data Science Group, University of Hamburg \\
         $^4$ Department of Data Science \& AI, Monash University, Australia \\
          \url{www.ukp.tu-darmstadt.de} 
         }%

\begin{document}
\maketitle
\begin{abstract}
Peer review is a cornerstone of quality control in scientific publishing. With the increasing workload, the unintended use of `quick' heuristics, referred to as \emph{lazy thinking}, has emerged as a recurring issue compromising review quality. Automated methods to detect such heuristics can help improve the peer-reviewing process. However, there is limited NLP research on this issue, and no real-world dataset exists to support the development of detection tools. This work introduces $\textsc{LazyReview}$, a dataset of peer-review sentences annotated with fine-grained \emph{lazy thinking} categories. Our analysis reveals that Large Language Models (LLMs) struggle to detect these instances in a zero-shot setting. However, instruction-based fine-tuning on our dataset significantly boosts performance by 10-20 performance points, highlighting the importance of high-quality training data. Furthermore, a controlled experiment demonstrates that reviews revised with \emph{lazy thinking} feedback are more comprehensive and actionable than those written without such feedback. We will release our dataset and the enhanced guidelines that can be used to train junior reviewers in the community.\footnote{Code available here: \url{https://github.com/UKPLab/acl2025-lazy-review}}
\end{abstract}

\section{Introduction}

Peer Reviewing is widely regarded as one of the most effective ways to assess the quality of scientific papers~\cite{ware2015stm}. It is a distributed procedure where the experts (\textit{reviewers}) independently evaluate whether a submitted manuscript adheres to the standards of the field. With the Mathew effect~\cite{merton1968matthew} in science (``rich get richer''), where the researchers receive benefits throughout their career for having papers at prestigious venues, it is of utmost importance to ensure sound practices in the reviewing process.

With the growing load of paper submissions, reviewers often face an overwhelming workload~\cite{landhuis2016scientific, 10.3389/phrs.2022.1605407} to assess multiple manuscripts \emph{quickly}. When presented with a cognitively hard task (e.g., reviewing) coupled with information overload and limited time, humans often end up using simple decision rules, also known as \emph{heuristics}~\cite{tversky1982judgment}. Though simple and efficient, these heuristics can often lead to errors and unfair evaluations~\cite{raue2018use}. 
\begin{figure}[!t]
    \centering
    \includegraphics[width=0.48\textwidth]{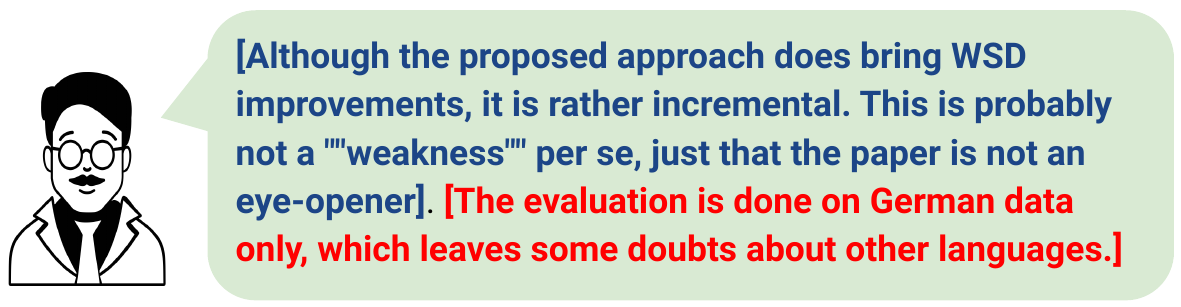}
  \vspace{-9mm}
    \caption{Illustration of lazy thinking in ARR-22 reviews sourced from \textsc{NLPeer}~\cite{dycke-etal-2023-nlpeer}. The first review segment belongs to the class `The results are not novel.' The last segment pertains to, `The approach is tested only on [not English], so unclear if it will generalize to other languages.' as per ARR-22 guidelines.}
    \vspace{-5mm}
    \label{fig:lazy_thinking}
\end{figure}
The usage of such heuristics to dismiss research papers in the Natural Language Processing (NLP) Community has been termed as \emph{lazy thinking}~\cite{Rogers_Augenstein_2021}. One such example is shown in Fig~\ref{fig:lazy_thinking}. Here, the reviewer dismisses the paper in the first review segment for not being an ``eye-opener''. However, they do not provide any references regarding similar prior work or feedback to improve the paper. This corresponds to the \emph{lazy thinking} class `The results are not novel'. 

In 2021, when ACL Rolling Review (ARR) was adopted as one of the main reviewing platforms for major NLP conferences, these heuristics were added to the guidelines~\cite{Rogers_Augenstein_2021}, aiming to discourage reviewers from relying on such approaches.\footnote{\url{https://aclrollingreview.org/}} However, in their ACL 2023 report, \citet{rogers-etal-2023-report} identified the usage of these heuristics as one of the leading factors (\textbf{24.3}\%) of author-reported issues in peer reviews. Therefore, in this work, we focus on this pertinent issue, \emph{lazy thinking} in NLP peer reviewing and heed \citet{kuznetsov2024can}'s call for automated methods to signal such occurrences in order to improve the overall reviewing quality.



To have a finer look at the problem of \emph{lazy thinking} in NLP peer reviews, we introduce \textsc{LazyReview}, a dataset with \textbf{500} expert-annotated review segments and \textbf{1276} silver annotated review segments from the best-performing model tagged with \emph{lazy thinking} classes with a review segment consisting of 1 or more review sentences. We develop this dataset over three rounds of annotation and incrementally sharpen the guidelines guided by inter-annotator agreements. We further provide positive examples, i.e., annotated examples for each class, to enhance the understanding of annotators for this task and reach annotation consensus sooner. 
Annotating review segments by a new batch of annotators who were not involved in developing the guidelines resulted in Cohen's $\kappa$~\cite{landis1977measurement} values of \textbf{0.32}, \textbf{0.36}, and \textbf{0.48}, respectively. The steady increase in agreement across rounds justifies the effectiveness of the developed guidelines.

With this dataset, we further test the zero-shot capabilities of LLMs in identifying \emph{lazy thinking}, which can be leveraged downstream to identify rule-following behaviour~\cite{sun2024beyond} in peer-review scenarios. Despite likely exposure to peer reviews via platforms like OpenReview during pre-training, LLMs struggle to accurately identify the type of \emph{lazy thinking} when presented with the current guidelines. To enhance their comprehension of \emph{lazy thinking}, we employ instruction-tuning on the LLMs using our dataset, leading to significant performance gains—around \textbf{10-20} accuracy points compared to their zero-shot and few-shot performances. 
Finally, we perform a controlled experiment where human reviewers rewrite peer reviews with(out) using annotations of \emph{lazy thinking} from our dataset. Human preference-based evaluations reveal that reviews written with the \emph{lazy thinking} guidance are more comprehensive and actionable than their vanilla counterparts. 

\noindent \textbf{Contributions.} We make the following contributions: \textbf{(1)} Introduce \textsc{LazyReview}, a dataset annotated with \emph{lazy thinking} classes for a new task in model development and evaluation. \textbf{(2)} Enhance annotation guidelines to improve both automated and human annotations. \textbf{(3)} Demonstrate that positive examples boost annotation quality and in-context learning. \textbf{(4)} Show that instruction tuning on our dataset enhances model performance. \textbf{(5)} Find that annotated \emph{lazy thinking} classes improve review quality.

\section{\textsc{LazyReview}: A Dataset for detecting Lazy Thinking in Peer Reviews}\label{sec:background}
In line with \citet{Rogers_Augenstein_2021}, we first define \emph{lazy thinking} as follows.
\begin{definition}
    `Lazy thinking, in the context of NLP research paper reviews, refers to the practice of dismissing or criticizing research papers based on superficial heuristics or preconceived notions rather than thorough analysis. It is characterized by reviewers raising concerns that lack substantial supporting evidence and are often influenced by prevailing trends within the NLP community.'
\end{definition}

\citet{Rogers_Augenstein_2021} enlist a total of 14 types of \textit{lazy thinking} heuristics in the ARR 2021 guidelines adapted from the study in \citet{rogers-augenstein-2020-improve}. We show some of the classes as described in the guidelines in Table~\ref{tab:arr_guidelines}.\footnote{The full table is provided in Table~\ref{tab:full_arr} in Appendix~\S\ref{sec:guidelines}} Re-iterating the example in Fig~\ref{fig:lazy_thinking}, we note that such claims about novelty need to be backed up with supporting evidence and hence constitute a classic example of \emph{lazy thinking} as per the guidelines. 

In this section, we describe the creation of our dataset, $\textsc{LazyReview}$ guided by the ARR 2022~\cite{Rogers_Augenstein_2021} and the EMNLP 2020 guidelines~\cite{LiuCohnEtAl_2020_Advice_on_Reviewing_for_EMNLP}. We describe the dataset curation process followed by an analysis of the dataset. This is the first dataset of annotated instances with fine-grained \emph{lazy thinking} classes for NLP peer reviews.

\begin{table*}[t]
    \centering
    \small{
    \begin{tabularx}{0.9\textwidth}{XXX}%
    \toprule
    \hldb{\textbf{Heuristics}} & \hldb{\textbf{Description}} & \hldb{\textbf{Example review segments}} \\
    \midrule
    \emph{The results are not surprising} & Many findings seem obvious in retrospect, but this does not mean that the community is already aware of them and can use them as building blocks for future work. & Transfer learning does not look to bring significant improvements. Looking at the variance, the results with and without transfer learning overlap. This is not surprising.  \\ \midrule
    \emph{The results are not novel	} & Such broad claims need to be backed up with references. & The approach the authors propose is still useful but not very novel. \\ \midrule
    \emph{The paper has language errors} & As long as the writing is clear enough, better scientific content should be more valuable than better journalistic skills. & The paper would be easy to follow with English proofreading even though the overall idea is  understandable. \\ \midrule   
    \end{tabularx}}
    \vspace{-3.5mm}
    \caption{Descriptions for some of the \textit{lazy thinking} classes sourced from ARR 2022 guidelines~\cite{Rogers_Augenstein_2021}. We present some examples corresponding to these classes from our dataset, \textsc{LazyReview}.}
    \vspace{-6mm}
    \label{tab:arr_guidelines}
\end{table*}

\subsection{Review Collection and Sampling} \label{sec:rev_collection}
We use the ARR 2022 reviews from the \textsc{NLPeer} Dataset~\cite{dycke-etal-2023-nlpeer} in which the reviews are collected using the 3Y-pipeline~\cite{dycke-etal-2022-yes} and as such has clear licenses attached. The dataset comprises reviews from five venues: CONLL 2016, ACL 2017, COLING 2020, ARR 2022, and F1000RD (an open-source science journal), with 11K review reports sourced from 5K papers. Focusing on the \emph{lazy thinking} definition in the ARR 2022 guidelines~\cite{Rogers_Augenstein_2021}, we consider only the ARR-22 reviews, using 684 reviews from 476 papers with 11,245 sentences in total. Each review is divided into sections like `paper summary', `summary of strengths', `summary of weaknesses', and `comments, typos, and suggestions', with reviewers completing the relevant sections based on their evaluation. However, there is no standardized format for expressing concerns, so we use automated methods to extract relevant review segments, as detailed later.

\subsection{Review Formatting} We utilize GPT-4~\cite{openai2024gpt4technicalreport} to prepare the instances (review segments) for subsequent annotations. We instruct GPT-4 to extract the review segments from the `Summary of Weaknesses' section of the review that can be classified into one of the lazy thinking classes, as outlined in the ARR guidelines\footnote{The prompt for GPT-4 is provided in Appendix~\S\ref{sec:gpt_prompt}}. Because \emph{lazy thinking} is often a factor in paper rejection and is specifically defined within the `Summary of Weaknesses’ section according to the ARR guidelines~\cite{Rogers_Augenstein_2021}, we limit our focus to this section. Additionally, ARR serves as the primary reviewing platform for major NLP conferences, such as ACL, EMNLP, and NAACL. As a result, the reviews encompass a wide range of topics and discussions across various prominent NLP conferences.
We obtain \textbf{1,776} review \textbf{segments} after this step, with each segment having varied lengths from 1 to 5 review sentences, as described later in Sec~\ref{sec:dataset_analysis}. To validate the quality of the extracted segments, we sample a set of $100$ segments from this pool and task three annotators to independently annotate each segment within the context of the entire review to decide on their candidacy towards \emph{lazy thinking}. The final label is chosen based on a majority vote. We obtain Cohen's $\kappa$ of \textbf{0.82} and obtain precision, recall and F1 scores as 0.74, 1.00 and 0.85, respectively. 
Intuitively, this means that GPT-4 samples more candidates than required, giving us a larger pool of candidates for annotation. We therefore introduce another class to our labelling schema, \textbf{`None'} to annotate non-\emph{lazy thinking} candidates.

\subsection{Annotation Protocol} \label{sec:annotation_proto}
Annotators are given the full review and the target review segment (highlighted) to classify according to incrementally developed lazy thinking guidelines based on ARR 2022. They also indicate their confidence levels: \textbf{low}, \textbf{medium}, or \textbf{high}. To aid in understanding, we added two additional classes: \textbf{`None'} for no lazy thinking, and \textbf{`Not Enough Information'} for instances lacking specificity or needing the paper for proper classification.

Two Ph.D. students, both fluent in English and experienced in NLP peer reviewing, are tasked with annotating review segments iteratively, given that peer reviewing requires specialized expertise. A senior PostDoc with an NLP background acts as a third annotator to resolve any disagreements between the initial annotations. The guidelines evolve over multiple rounds. Once the guidelines are refined and finalized, we recruit a new batch of annotators to re-validate them by asking them to annotate the same review segments. These new annotators follow the same set of guidelines used in the earlier rounds to ensure consistency. After this validation process, the original Ph.D. annotators are retained to annotate additional instances.


\subsection{Evolving Guidelines} \label{sec:guidelines}
We incrementally improve the guidelines guided by the inter-annotator agreements. Given the high subjectivity of this domain, we consider the guidelines to be precise once we achieve a substantial agreement on annotating the instances further. \looseness=-1

\noindent \textbf{\hl{Round 1: ARR 2022 Guidelines.}} In this round, the annotators are provided with the existing ARR~\cite{Rogers_Augenstein_2021} guidelines and asked to annotate the lazy thinking classes. We sample a set of \textbf{50} instances from the review segments extracted by GPT-4 as described in Sec~\S\ref{sec:rev_collection}. After the first round of annotation, we obtain a Cohen's kappa, $\kappa$ of \textbf{0.31}. This is a substantially low agreement, revealing an ambiguity in the reviewing guidelines. The confusion matrix for the first round of annotation is shown in Fig~\ref{fig:r1_mat} (cf. \S\ref{sec:ann_conf}). We find that there is a high agreement for the `None' class, which implies that the annotators can easily detect a review segment that is not problematic. However, they struggle to determine the fine-grained label of the problematic review segments. Further analysis of the confidence level of the annotators reveals that for most of the cases, the annotators have low confidence, as shown in Fig~\ref{fig:r1_conf} (cf. \S\ref{sec:ann_conf}), which points towards ambiguity in the guidelines. 

\noindent \textbf{\hl{Round 2: ARR 2022 and EMNLP guidelines.}} We further explored the initial reviewing guidelines released during EMNLP 2020~\cite{LiuCohnEtAl_2020_Advice_on_Reviewing_for_EMNLP}. We identify some additional classes that are now missing from the ARR guidelines, namely \emph{`Non-mainstream Approaches'} (rejecting papers for not using current trending approaches) and \emph{`Resource Paper'} (favoring resource papers lesser for ACL conferences). Additionally, we extend descriptions of some of the classes such as, \emph{`This has no precedent in existing literature'}, \emph{`The method is too simple'} using the guidelines in \citet{LiuCohnEtAl_2020_Advice_on_Reviewing_for_EMNLP}. Moreover, we extend the name of some of the class labels based on the EMNLP 2020 guidelines, such as \emph{`The paper has language errors'} with \emph{`Writing Style'}; \emph{`The topic is too niche'} with  \emph{`Narrow topics'}, which have similar meanings. We show the extended descriptions for those classes in Table~\ref{tab:arr_guidelines_r2} (cf. Appendix~\S\ref{sec:guidelines}). We annotate the same set of instances as in Round 1 and eventually calculate agreements. We obtain Cohen's $\kappa$ of \textbf{0.38}, which is significantly higher than the previous round (0.31). We observe higher agreements for the classes having extended names such as, `The paper has language errors' and `Niche Topics', as illustrated in Fig~\ref{fig:r2_mat} (cf. \S\ref{sec:ann_conf}). The confidence level of the annotators also substantially increased from low to medium in this round, as can be seen in Fig~\ref{fig:r2_conf} (cf. \S\ref{sec:ann_conf}).\looseness=-1

\noindent \textbf{\hl{Round 3: Round 2 guidelines with positive examples.}}  To promote quick learning through ``worked examples"~\cite{doi:10.3102/00346543070002181}, we refine our annotation round by fixing the guidelines and incorporating positive examples for each lazy thinking class. We evaluated several techniques to select representative examples, measuring effectiveness using Cohen's $\kappa$. The methods include (a) \hldb{Random}: selecting the shortest or longest review segments, and (b) \hldb{Diversity}: encoding segments with SciBERT~\cite{beltagy-etal-2019-scibert}, clustering them using K-Means, and choosing the cluster center with the lowest cosine distance. After pairwise comparisons, the \hldb{random shortest} method was preferred, achieving a Cohen's $\kappa$ of \textbf{0.86}. For round 3, we sampled \textbf{50} new review segments and included annotated examples from the previous round, resulting in a Cohen's $\kappa$ of \textbf{0.52}, indicating substantial agreement despite the task's subjectivity, as shown in Fig~\ref{fig:r3_mat} (cf. \S\ref{sec:ann_conf}). We also obtain higher annotator confidence levels as shown in Fig~\ref{fig:r3_conf} (cf. \S\ref{sec:ann_conf}).\footnote{The examples used in this round are in Table~\ref{tab:full_arr_positive} in \S\ref{sec:guidelines}}

\noindent \textbf{\hl{Final Validation.}} To validate the guidelines developed over three rounds, we hired a new group of English-speaking PhD students with NLP peer review expertise to annotate the same set of instances. These annotators were not involved in the development of the guidelines. Re-annotation across rounds 1, 2, and 3 resulted in inter-annotator agreement ($\kappa$) values of \textbf{0.32}, \textbf{0.36}, and \textbf{0.48}, respectively. This steady increase in $\kappa$ aligns with previous results of \textbf{0.31}, \textbf{0.38}, and \textbf{0.52}, validating the iterative improvement of the guidelines based on stronger inter-annotator agreement. Agreement scores higher than 0.4 are considered moderate due to the inherent subjectivity of the task and are consistent with agreement levels reported in previous studies within the peer-reviewing domain~\cite{kennard-etal-2022-disapere, purkayastha-etal-2023-exploring}. 
We retain the annotators from the initial rounds who developed the guidelines to annotate a total of \textbf{500 review segments} as a part of our dataset, \textsc{LazyReview}. The overall annotation time amounted to 20 hours. This corresponds to 25 minutes per example, ranging from less than a minute for annotating shorter segments up to half an hour for longer ones.   
\begin{figure}[!t]
\centering
    \includegraphics[width=0.7\linewidth]{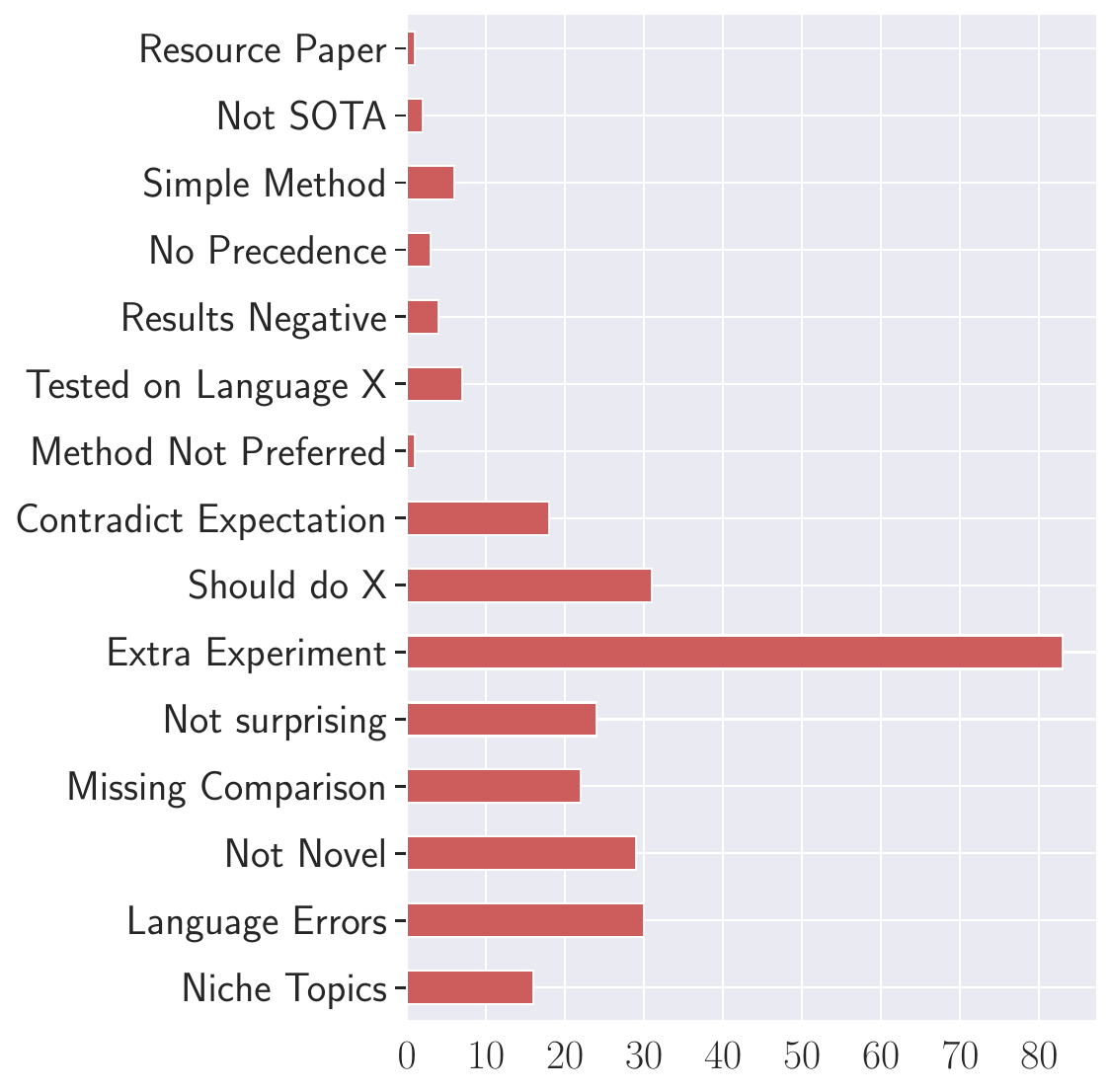}
        \vspace{-3.5mm}
    \caption{Distribution of \emph{lazy thinking} labels in our dataset, \textsc{LazyReview}.}
    \vspace{-7mm}
    \label{fig:dataset_analysis}
\end{figure}

\subsection{Dataset Analysis}\label{sec:dataset_analysis}
Our dataset comprises \textbf{500} expert-annotated review segments tagged with \textbf{18} classes. Out of all the labels, \textbf{16} classes in our dataset correspond to explicit \emph{lazy thinking}. The distribution of our dataset is shown in Fig~\ref{fig:dataset_analysis} corresponding to these 16 classes. The most frequent lazy thinking is \hldb{`Extra Experiments'}, where the reviewers ask the authors to conduct additional experiments without proper justification. This trend reflects the current expectations in the NLP field, which has rapidly shifted towards machine learning models and methods that often emphasize extensive empirical evaluations~\cite{gururaja-etal-2023-build}. The next most frequent classes are \hldb{`Not Enough Novelty'} and \hldb{`Language Errors'}. The ACL review report does not state the individual distribution of these classes but constitutes around 24.3\% of all reported issues. We further analyze the sentence length of the review segments within these classes. This is illustrated in Fig~\ref{fig:sent_length} of ~\S\ref{sec:seg_length}. We observe that most of these review segments have a length of 1 sentence, underscoring the use of shorter arguments to dismiss papers. The \emph{lazy thinking} class `Extra Experiment' is the most common with variable segment lengths.





\section{Experiments} \label{sec:experiments}
We use the \textsc{LazyReview} dataset to assess the performance of various open-source LLMs in detecting \emph{lazy thinking} in NLP peer reviews. 

\subsection*{Experimental Setup} \label{sec:models} 
\noindent \hl{\textbf{Tasks.}} \label{sec:tasks} We propose two formulations for detecting lazy thinking in peer reviews: (i) \textbf{Coarse-grained classification}: a binary task to determine if an input segment $x$ is lazy thinking, and (ii) \textbf{Fine-grained classification}: a multi-class task to classify $x$ into one of the specific lazy thinking classes, $c_i \in C$.

\noindent \textbf{\hl{Models.}} Since the guidelines for NLP conferences are mainly instructions, we explore various open-source instruction-tuned LLMs for this task. We experiment with the \textbf{chat versions} of LLaMa-2 7B and 13B~\cite{touvron2023llama2openfoundation} (abbr. LLaMa, LLaMa\textsubscript{L}), Mistral 7B instruct~\cite{jiang2023mistral7b} (abbr. Mistral), Qwen-Chat 1.5 7B~\cite{bai2023qwentechnicalreport} (abbr. Qwen), Yi-1.5 6B instruct~\cite{ai2024yiopenfoundationmodels} (abbr. Yi-1.5), Gemma-1.1 7B instruct~\cite{gemmateam2024gemmaopenmodelsbased} (abbr. Gemma), and SciT\"ulu 7B~\cite{wadden2024sciriffresourceenhancelanguage} (abbr. SciT\"ulu).\footnote{The justification for the choice of these models, along with implementation details, is in  Appendix~\S\ref{sec:model_details}}

\noindent \textbf{\hl{Prompts.}} Following our annotation protocol, we prompt the LLMs with guidelines and instructions for each round. For \textbf{coarse-grained classification}, the model determines whether the input is an instance of \emph{lazy thinking} or not. For \textbf{fine-grained one}, the model selects the best-fitting lazy thinking class from the provided options. We test two input types: the target segment (T) alone, or the combination of the review and target segment (RT).\footnote{Full details are in Appendix~\S\ref{sec:prompt_details}.}

\noindent \textbf{\hl{Metrics.}}
To evaluate LLM outputs, we use both strict and relaxed measures because LLMs sometimes do not produce exact label phrases. The strict measure, as defined by \citet{helwe-etal-2024-mafalda}, uses regular expressions to check for matches between the gold label and predictions, reporting accuracy and macro-F1 (string matching). The relaxed measure employs GPT-3.5 to judge whether the provided answer is semantically equivalent to the ground-truth class, outputting a ``yes'' or ``no'' decision. This approach follows previous work on evaluating free-form LLM outputs \cite{wadden2024sciriffresourceenhancelanguage, holtermann2024evaluatingelementarymultilingualcapabilities}, and reports accuracy based on the number of ``yes'' responses. Both metrics determine whether predictions are correct or incorrect. We also performed an alignment study on 50 responses from every model to validate the reliability of the evaluators. We find that the string-matching-based evaluator underestimates the correct predictions, whereas the GPT-based evaluator overestimates them, rendering a correct balance of lower and upper bounds for the model predictions.\footnote{Details about the study are in Appendix~\S\ref{sec:metrics}.}

\vspace{-2mm}
\subsection*{RQ1: How effective are the improved guidelines in enhancing zero-shot performance of LLMs?}
Since the first two rounds of our annotation study are dedicated to fixing the guidelines for annotations, we evaluate the understanding of the LLMs on the same \textbf{50} instances on both annotation rounds, i.e., rounds 1 and 2, respectively. This validates whether the improved guidelines actually influence the zero-shot performance of LLMs.

\noindent \textbf{\hl{Modelling Approach.}} We prompt the LLMs for round 1, with the definition of \emph{lazy thinking} classes, as shown in Table~\ref{tab:arr_guidelines}, representing the existing ARR guidelines. For round 2, we prompt the LLMs with the new guidelines as described in Sec~\S\ref{sec:guidelines} where we added new classes and extended descriptions of the existing classes in the ARR guidelines using the EMNLP 2020 Blog~\cite{LiuCohnEtAl_2020_Advice_on_Reviewing_for_EMNLP}.

\noindent \textbf{\hl{Results.}} 
We present results for fine-grained and coarse-grained classification of zero-shot LLMs in Table~\ref{tab:ann_round_llm_fine}.\footnote{Full results in Tables~\ref{tab:ann_round_llm_fine-huge} and \ref{tab:ann_round_llm_coarse} of Appendix~\S\ref{sec:full_results}.} Using only the target segment (T) as input generally outperforms using both the target segment and review (RT) across most models, likely due to spurious correlations from longer inputs, as noted by \citet{feng-etal-2023-less}. All models improve from round 1 (R1) to round 2 (R2), with Gemma gaining 4.5 points in string accuracy (S.A). SciT\"ulu, however, benefits from the broader context of RT, possibly due to its pre-training on long-context tasks. Coarse-grained classification scores are consistently higher than fine-grained, showing LLMs can serve as basic \emph{lazy thinking} detectors. 

\begin{table}[!t]
\centering
\resizebox{!}{0.18\textwidth}{\begin{tabular}{lllllllll}
\hline
\multirow{3}{*}{\textbf{Models}} & \multicolumn{4}{c}{\textbf{Fine-gr.}}                                                                      & \multicolumn{4}{c}{\textbf{Coarse-gr.}}                                                \\ \cmidrule(lr){2-5} \cmidrule(lr){6-9}
                                 & \multicolumn{2}{c}{\textbf{R1}}                              & \multicolumn{2}{c}{\textbf{R2}}                              & \multicolumn{2}{c}{\textbf{R1}}                              & \multicolumn{2}{c}{\textbf{R2}}          \\ \cline{2-3} \cmidrule(lr){4-5} \cmidrule(lr){6-7} \cmidrule(lr){8-9}
                                 & \textbf{S.A} & \multicolumn{1}{l}{\textbf{G.A}} & \textbf{S.A} & \multicolumn{1}{l}{\textbf{G.A}} & \textbf{S.A} & \multicolumn{1}{l}{\textbf{G.A}} & \textbf{S.A} & \textbf{G.A} \\ \cmidrule(lr){1-1} \cmidrule(lr){2-3} \cmidrule(lr){4-5} \cmidrule(lr){6-7} \cmidrule(lr){8-9}
\textsc{Random}                    & 7.11           & -                                   & \hldb{4.34}           & -                                   & 40.7           & -                                   & 40.7           & -              \\
\textsc{Majority}                  & 11.1           & -                                   & \hldb{7.34}           & -                                   & 51.4           & -                                   & 51.4           & -              \\ \hline
Gemma + \textsc{T}                 & 22.2           & 52.2                                & \hlgreen{26.7}           & \hlgreen{58.1}                                & 44.3           & 51.1                                & \hlgreen{46.1}           & \hlgreen{54.4}           \\
Gemma + \textsc{RT}                & 12.2           & 46.7                                & \hldb{11.6}           & \hlgreen{51.1}                                & 48.1           & 47.4                                & \hlgreen{50.4}           & \hlgreen{49.1}           \\ \hline
LLaMa + \textsc{T}                   & 12.2           & 15.6                                & \hlgreen{22.2}           & \hlgreen{30.6}                                & 57.7           & 70.0                                & \hlgreen{60.0}           & \hlgreen{75.0}           \\
LLaMa + \textsc{RT}                  & 12.2           & 25.6                                & \hlgreen{13.2}           & \hlgreen{33.7}                                & 53.3           & 55.1                                & \hlgreen{60.0}           & \hlgreen{67.7}           \\ \hline
LLaMa\textsubscript{L} + \textsc{T}                    & 26.7           & 44.4                                & \hldb{26.7}           & \hlgreen{45.3}                                & 60.2           & 73.1                                & \hlgreen{62.2}           & \hlgreen{75.4}           \\
LLaMa\textsubscript{L} + \textsc{RT}                   & 15.6           & 41.1                                & \hlgreen{17.6}           & \hldb{40.4}                                & 68.6           & 69.4                                & \hlgreen{70.2}           & \hlgreen{70.2}           \\ \hline
Mistral + \textsc{T}                   & 27.8           & 47.8                                & \hlgreen{28.8}           & \hlgreen{51.1}                                & 57.8           & 64.8                                & \hlgreen{58.8}           & \hlgreen{66.3}           \\
Mistral + \textsc{RT}                & 12.2           & 28.9                                & \hlgreen{16.6}           & \hlgreen{35.9}                                & 55.4  & 53.8                                & \hlgreen{57.4}           & \hlgreen{56.0}           \\ \hline
Qwen + \textsc{T}                      & 21.1           & 46.7                                & \hlgreen{22.7}           & \hlgreen{50.0}                                & \textbf{68.9}  & \textbf{74.1}                                & \hlgreen{\textbf{70.4}}  & \hlgreen{\textbf{76.1}}           \\
Qwen + \textsc{RT}                 & 12.2           & 43.3                                & \hlgreen{13.3}           & \hldb{42.6}                                & 53.3           & 53.3                                & \hlgreen{56.5}           & \hlgreen{56.5}           \\ \hline
Yi-1.5 + \textsc{T}                   & \textbf{35.3}  & \textbf{56.7}                       & \hlgreen{\textbf{37.6}}  & \hlgreen{\textbf{60.0}}                       & 64.4           & 71.1                                & \hlgreen{68.7}           & \hlgreen{73.4}           \\
Yi-1.5 + \textsc{RT}                  & 34.4           & 51.1                                & \hldb{32.8}           & \hlgreen{52.2}                                & 63.3           & 65.1                                & \hlgreen{68.3}           & \hlgreen{70.4}           \\ \hline
SciT\"ulu + \textsc{T}                & 14.4           & 18.1                                & \hlgreen{25.3}           & \hlgreen{29.4}                                & 57.8           & 57.8                                & \hlgreen{58.3}           & \hlgreen{58.3}           \\
SciT\"ulu + \textsc{RT}               & 15.6           & 17.3                                & \hlgreen{18.3}           & \hlgreen{23.7}                                & 55.6           & 55.6                                & \hlgreen{58.7}           & \hlgreen{58.7}   \\ \hline       
\end{tabular}}
\vspace{-2.5mm}
\caption{LLM performance across annotation rounds in terms of string-matching (S.A) and GPT-based (G.A) \textbf{accuracy} for fine-grained (Fine-gr.) and coarse-grained (Coarse-gr.) tasks. `T' uses only the target sentence, `RT' combines review and target. R1 and R2 represent `Round 1' and `Round 2'. Increments from R1 to R2 are highlighted in \hlgreen{red}; decreases or no change in \hldb{gray}.}
\vspace{-6.5mm}
\label{tab:ann_round_llm_fine}
\end{table}

\subsection*{RQ2: How effective are positive examples in improving the performance of LLMs?} \label{sec:incontext}
To test the effect of using positive examples on the performance of LLMs, we 
leverage in-context learning to test the models with the same \textbf{50} instances as used in round 3 of our annotation study. 

\noindent \textbf{\hl{Modelling Approach.}} We prompt LLMs using positive examples from previous rounds and Round 2 guidelines, combining target segments (TE) and the combination of review and target segments (RTE) as in-context learning examples. Since LLMs have a fixed context window, we employ two setups for selecting examples: (i) \textbf{Static} selection of fixed random examples for all inferences, and (ii) \textbf{Dynamic} selection on per test case basis using following methods: (a) \textit{BM25}: uses BM25 to select 
examples, (b) \textit{Top K}: uses embedding space similarity, and (c) \emph{Vote-K}: penalizing already selected examples. We tested with 1, 2, and 3 exemplars.\footnote{We leverage OpenICL~\cite{wu-etal-2023-openicl} for retrieving exemplars and GPT2-XL as the embedding generator. }
\begin{table}[!t]
\centering
\resizebox{!}{0.18\textwidth}{\begin{tabular}{lllll}
\hline
\multicolumn{1}{l}{\multirow{2}{*}{\textbf{Models}}} & \multicolumn{2}{c}{\textbf{Fine-gr.}}                               & \multicolumn{2}{c}{\textbf{Coarse-gr.}}          \\  \cmidrule(lr){2-3} \cmidrule(lr){4-5}
\multicolumn{1}{c}{}                                 & \multicolumn{1}{l}{\textbf{S.A}} & \multicolumn{1}{l}{\textbf{G.A}} & \multicolumn{1}{l}{\textbf{S.A}} & \textbf{G.A}           \\ \cmidrule(lr){1-1} \cmidrule(lr){2-3}  \cmidrule(lr){4-5}
\textsc{Random}                                      & 2.46                             &   -                                & 43.3                             & -              \\
\textsc{Majority}                                    & 5.11                             &   -                                & 52.3                             &   -            \\ \hline
Gemma + \textsc{TE}                                  & \hlgreen{24.4}$_{5.5}$                             & \hlgreen{41.1}$_{8.9}$                              & \hlgreen{75.6}$_{20.0}$                             & \hlgreen{88.9}$_{31.7}$          \\ \hdashline
Gemma + \textsc{RTE}                                 & \hlgreen{17.3}$_{2.9}$                             & \hlgreen{32.8}$_{0.2}$                              & \hlgreen{71.1}$_{5.5}$                             & \hlgreen{82.2}$_{16.6}$          \\ \hline
LLaMa + \textsc{TE}              & \hlgreen{15.6}$_{4.5}$ & \hlgreen{38.9}$_{3.3}$ & \hlgreen{84.4}$_{4.4}$ & \hlgreen{89.1}$_{3.0}$ \\ \hdashline
LLaMa + \textsc{RTE}             & \hlgreen{14.2}$_{2.0}$ & \hlgreen{30.8}$_{1.9}$ & \hlgreen{75.3}$_{2.0}$ & \hlgreen{81.1}$_{4.4}$ \\ \hline
LLaMa\textsubscript{L} + \textsc{TE}   & \hlgreen{24.4}$_{13.3}$ & \hlgreen{41.1}$_{5.5}$ & \hlgreen{73.1}$_{1.9}$ & \hlgreen{71.1}$_{10.0}$ \\ \hdashline
LLaMa\textsubscript{L} + \textsc{RTE}  & \hlgreen{18.8}$_{8.1}$ & \hlgreen{34.4}$_{2.2}$ & \hlgreen{70.3}$_{1.5}$ & \hlgreen{61.1}$_{9.9}$ \\ \hline
Mistral + \textsc{TE}            & \hlgreen{30.0}$_{1.8}$ & \hlgreen{55.6}$_{1.2}$ & \hlgreen{86.7}$_{12.3}$ & \hlgreen{86.7}$_{11.5}$ \\ \hdashline
Mistral + \textsc{RTE}           & \hlgreen{27.8}$_{5.6}$ & \hlgreen{52.2}$_{1.1}$ & \hlgreen{68.8}$_{6.6}$  & \hlgreen{68.8}$_{6.6}$  \\ \hline
Qwen + \textsc{TE}               & \hlgreen{\textbf{31.1}}$_{2.2}$ & \hlgreen{\textbf{56.4}}$_{12.0}$ & \hlgreen{86.7}$_{4.0}$ & \hlgreen{86.7}$_{4.0}$ \\ \hdashline
Qwen + \textsc{RTE}              & \hlgreen{27.8}$_{1.1}$ & \hlgreen{44.2}$_{0.9}$ & \hlgreen{62.2}$_{2.0}$ & \hlgreen{62.2}$_{2.0}$ \\ \hline

Yi-1.5 + \textsc{TE}             & \hlgreen{30.0}$_{3.3}$ & \hlgreen{54.9}$_{1.1}$ & \hlgreen{74.5}$_{3.2}$ & \hlgreen{73.8}$_{1.5}$ \\ \hdashline
Yi-1.5 + \textsc{RTE}            & \hlgreen{24.4}$_{3.2}$ & \hlgreen{52.7}$_{1.4}$ & \hlgreen{70.1}$_{2.0}$ & \hlgreen{72.4}$_{2.3}$ \\ \hline
SciT\"ulu + \textsc{TE}          & \hlgreen{23.3}$_{1.1}$ & \hlgreen{44.8}$_{2.6}$ & \hlgreen{72.2}$_{21.1}$ & \hlgreen{72.2}$_{20.0}$ \\ \hdashline
SciT\"ulu + \textsc{RTE}         & \hlgreen{19.7}$_{0.8}$ & \hlgreen{41.1}$_{2.2}$ & \hlgreen{\textbf{88.8}}$_{17.7}$ & \hlgreen{\textbf{91.1}}$_{22.3}$ \\ \hline
\end{tabular}}
\vspace{-2mm}
\caption{Performance of LLMs for round 3 in terms of the metrics used in Table~\ref{tab:ann_round_llm_fine} for fine-grained (Fine-gr.) and coarse-grained (Coarse-gr.) tasks. `E' denotes adding in-context exemplars to input types: `T' (target sentence) and `RT' (review + target sentence).  \hlgreen{Red}: Increments with exemplars. Subscripts represent increments as compared to the zero-short versions.}
\vspace{-7.5mm}
\label{tab:ann_round_llm_fine_icl}
\end{table}

\noindent \textbf{\hl{Results.}} We plot results in Fig~\ref{fig:models_icl} and \ref{fig:models_icl_coarse} for fine-grained and coarse-grained classification (cf. \S\ref{sec:incontext_app}), finding that the \hldb{static strategy} outperforms all other setups, especially for Mistral and Qwen. Increasing exemplars does not enhance performance, supporting \citet{srivastava-etal-2024-nice} that random examples yield better results than similarity-based methods. This also can be attributed to the ability of LLMs to learn the format of the task from exemplars rather than learning to perform the task~\cite{lu-etal-2024-emergent}. We adopt a \hldb{static} in-context learning method using \hldb{1 exemplar} for the rest of the paper. 

We show the results using \hldb{1 static example} in Table~\ref{tab:ann_round_llm_fine_icl}. \hldb{ICL-based methods surpass zero-shot models}, with notable gains in coarse classification (20 points for Gemma, 21 points for SciT\"ulu string-based accuracy (S.A.)). We also obtain positive increments across the board in fine-grained classifications. SciT\"ulu excels in the coarse-grained classification, possibly due to the domain specific pre-training. Qwen leads in fine-grained classification, likely due to the multi-lingual pre-training data and high-quality data filtering performed for the pre-training phase~\cite{bai2023qwentechnicalreport}.\footnote{Further analysis on the performance of Qwen is in Appendix~\ref{sec:further_analysis}.}


\subsection*{RQ3: How effective is instruction-tuning in improving the performance of LLMs?}
Building on previous work in aligning LLMs with new tasks~\cite{ivison2023camelschangingclimateenhancing, wadden2024sciriffresourceenhancelanguage}, we apply instruction-based fine-tuning for lazy-thinking detection. Using a bottom-up approach, we determine the data requirements using 3-fold cross-validations for maximum performance on a small validation set. We then identify optimal data mixes for the test set. We subsequently use a similar mix to train the models and then compare their performance to their non-instruction tuned counterparts from the previous annotation rounds and obtain silver annotations from the best model.

\noindent \textbf{\hl{Modelling Approach.}} We apply instruction-based finetuning for the same models as detailed in Sec~\S\ref{sec:models}. To optimize our limited computational resources, we employ LoRa~\cite{hulora} for parameter-efficient finetuning and utilize \textit{open-instruct}~\cite{wang2023-how-far-can}.\footnote{\url{https://github.com/allenai/open-instruct}} We use the same prompt~(cf.~\S\ref{sec:models}) as instruction along with the \textbf{round 2} guidelines (combination of guidelines of ARR and EMNLP blog) while performing this experiment. We use multiple data sources to create data mixes: (i) \textsc{T\"ulu V2}~\cite{ivison2023camelschangingclimateenhancing} with 326,154 samples, (ii) \textsc{SciRIFF}~\cite{wadden2024sciriffresourceenhancelanguage} with 154K demonstrations across 54 tasks, and (iii) \textsc{LazyReview} with 1000 samples evenly split between coarse and fine-grained classification. \textsc{LazyReview} is divided into 70\% training (700 examples), 10\% validation (100 examples), and 20\% testing (200 samples) in a 3-fold cross-validation setup. We train on 700 examples in the \textsc{No Mix} setting. We use an equal number of instances (700) from each data source to balance the other mixtures. We create \textsc{SciRIFF Mix} (1400 samples) and \textsc{T\"ulu Mix} (1400 samples) by combining \textsc{LazyReview} with the other datasets and a \textsc{Full Mix} (2100 samples) integrating all three.  We test the performance on the same cross-validated test sets for each mix. We use a proportion of \textbf{0.3} for the `T' (using only target segment) setup and \textbf{0.7} for the `RT' setup (using a combination of review and target segment) from the different mixes to optimize performance and data efficiency based on a hyper-parameter search on different validation sets.\footnote{Details on hyper-parameters are in~\S\ref{sec:training}, with tuning methods described in~\S\ref{sec:hyper-parameter}.}

\begin{table}[!t]
\resizebox{!}{0.25\textwidth}{\begin{tabular}{lllllll}
\hline
\multirow{2}{*}{\textbf{Models}} & \multicolumn{3}{c}{\textbf{Fine-gr.}}                           & \multicolumn{3}{c}{\textbf{Coarse-gr.}}                     \\ \cmidrule(lr){2-4} \cmidrule(lr){5-7}
                        & \textbf{R1}   & \textbf{R2}   & \multicolumn{1}{l}{\textbf{R3}} & \textbf{R1} & \textbf{R2} & \multicolumn{1}{l}{\textbf{R3}} \\ \cmidrule(lr){1-1} \cmidrule(lr){2-4} \cmidrule(lr){5-7}
\textsc{Rand.}           & 7.11          & 4.34          & 2.46                            & 40.7        & 40.7        & 43.3                            \\
\textsc{Maj.}         & 11.1          & 7.34          & 5.11                            & 51.4        & 51.4        & 52.3                            \\ \hline
Gemma                   & 22.2          & 26.7          & 24.4                            & 48.1        & 50.4        & 75.6                            \\
\quad $it$ + \textsc{T}     & \hlgreen{31.4}$_{9.2}$          & \hlgreen{38.8}$_{12.1}$          & \hlgreen{34.6}$_{10.2}$                            & \hlgreen{57.8}$_{9.7}$        & \hlgreen{61.4}$_{10.0}$        & \hlgreen{81.2}$_{5.6}$                            \\
\quad $it$ + \textsc{RT}     & \hlgreen{28.2}$_{6.0}$          & \hlgreen{35.7}$_{9.0}$          & \hlgreen{32.8}$_{8.4}$                            & \hlgreen{55.6}$_{7.5}$        & \hlgreen{59.4}$_{9.0}$        & \hlgreen{78.8}$_{3.2}$                            \\ \hline
LLaMa                   & 12.2          & 22.2          & 15.6                            & 57.7        & 60.0        & 84.4                            \\
\quad $it$ + \textsc{T}     & \hlgreen{43.8}$_{31.6}$          & \hlgreen{47.8}$_{25.6}$          & \hlgreen{44.7}$_{29.1}$                            & \hlgreen{62.7}$_{5.0}$        & \hlgreen{65.4}$_{5.0}$        & \hlgreen{85.4}$_{1.0}$                            \\
\quad $it$ + \textsc{RT}      & \hlgreen{43.2}$_{31.0}$          & \hlgreen{45.3}$_{23.1}$          & \hlgreen{41.8}$_{26.2}$                            & \hlgreen{61.2}$_{3.5}$        & \hlgreen{63.1}$_{3.1}$        & \hldb{81.3}$_{3.1}$                            \\ \hline
LLaMa\textsubscript{L}  & 26.7          & 26.7          & 24.4                            & 68.6        & 70.2        & 73.1                            \\
\quad $it$ + \textsc{T}     & \hlgreen{45.8}$_{19.1}$          & \hlgreen{47.8}$_{21.1}$          & \hlgreen{50.5}$_{26.1}$                            & \hlgreen{74.3}$_{5.7}$        & \hlgreen{74.6}$_{4.4}$        & \hlgreen{75.3}$_{2.2}$                            \\
\quad $it$ + \textsc{RT}     & \hlgreen{41.2}$_{14.5}$          & \hlgreen{45.2}$_{18.5}$          & \hlgreen{47.3}$_{22.9}$                            & \hlgreen{70.2}$_{1.6}$        & \hlgreen{71.8}$_{1.8}$        & \hlgreen{73.3}$_{0.2}$                            \\ \hline
Mistral                 & 27.8          & 28.8          & 30.0                            & 57.8        & 58.8        & 86.7                            \\
\quad $it$ + \textsc{T}     & \hlgreen{35.4}$_{7.6}$          & \hlgreen{37.4}$_{8.6}$          & \hlgreen{42.4}$_{12.4}$                            & \hlgreen{60.2}$_{2.4}$        & \hlgreen{62.2}$_{3.4}$        & \hldb{86.4}$_{0.3}$                            \\
\quad $it$ + \textsc{RT}     & \hlgreen{31.2}$_{3.4}$          & \hlgreen{35.2}$_{6.4}$          & \hlgreen{37.8}$_{7.8}$                            & \hlgreen{65.3}$_{7.5}$        & \hlgreen{68.2}$_{9.4}$        & \hlgreen{88.2}$_{1.5}$                            \\ \hline
Qwen                    & 21.1          & 22.7          & 31.1                            & 68.9        & 70.4        & 86.7                            \\
\quad $it$ + \textsc{T}     & \hlgreen{\textbf{45.9}}$_{24.8}$          & \hlgreen{48.4}$_{25.7}$          & \hlgreen{\textbf{59.4}}$_{28.3}$                            & \hlgreen{\textbf{75.4}}$_{6.5}$        & \hlgreen{\textbf{76.3}}$_{5.9}$        & \hlgreen{88.4}$_{1.7}$                            \\
\quad $it$ + \textsc{RT}     & \hlgreen{41.2}$_{20.1}$          & \hlgreen{42.4}$_{19.7}$          & \hlgreen{47.8}$_{16.7}$                            & \hlgreen{73.2}$_{4.3}$        & \hlgreen{74.1}$_{3.7}$        & \hldb{86.3}$_{0.4}$                            \\ \hline
Yi-1.5              & 35.3          & 37.6          & 30.0                            & 64.4        & 68.7        & 74.5                            \\
\quad $it$ + \textsc{T}     & \hlgreen{45.1}$_{9.8}$          & \hlgreen{47.8}$_{10.2}$          & \hlgreen{47.9}$_{17.9}$                            & \hlgreen{69.5}$_{5.1}$        & \hlgreen{74.2}$_{5.5}$        & \hlgreen{78.4}$_{3.9}$                            \\
\quad $it$ + \textsc{RT}     & \hlgreen{43.2}$_{7.9}$          & \hlgreen{45.3}$_{7.7}$          & \hlgreen{46.3}$_{16.3}$                            & \hlgreen{67.2}$_{2.8}$        & \hlgreen{69.4}$_{0.7}$        & \hldb{73.2}$_{1.3}$                            \\ \hline
SciT\"ulu               & 15.6          & 25.3          & 23.3                            & 57.8        & 58.3        & 88.8                            \\
\quad $it$ + \textsc{T}     & \hlgreen{45.7}$_{30.1}$          & \hlgreen{\textbf{48.6}}$_{23.3}$          & \hlgreen{54.3}$_{31.0}$                            & \hlgreen{66.3}$_{8.5}$        & \hlgreen{68.4}$_{10.1}$        & \hlgreen{\textbf{91.2}}$_{2.4}$                            \\
\quad $it$ + \textsc{RT}    & \hlgreen{41.4}$_{25.8}$          & \hlgreen{42.6}$_{17.3}$          & \hlgreen{51.4}$_{28.1}$                            & \hlgreen{62.4}$_{4.6}$        & \hlgreen{65.6}$_{7.3}$        & \hldb{87.2}$_{1.6}$    \\ \hline                       
\end{tabular}}
\vspace{-2.5mm}
\caption{Performance of LLMs after instruction tuning ($it$) for fine-grained classification using target segment (T) and the combination of review and target segment (RT) in terms of \textbf{string-matching accuracy} (St. (Acc)). The first row of each model states the best results obtained previously as detailed in Tables~\ref{tab:ann_round_llm_fine} and \ref{tab:ann_round_llm_fine_icl} respectively. Subscripts represent increment or decrement compared to the non-instruction tuned versions. Increments compared to the first row of each model are highlighted in \hlgreen{red} and decrements or no change in \hldb{gray}.}
\vspace{-7mm}
\label{tab:instruct_tune_rounds}
\end{table}
\noindent \textbf{\hl{Results.}} We compare instruction-tuned models to their zero and few-shot counterparts for fine-grained and coarse-grained classification using \hldb{3-fold cross-validation}, as shown in Tables~\ref{tab:test_set} and \ref{tab:test_set_coarse} (cf. \S\ref{sec:3_fold}). Instruction tuning significantly enhances model performance. The LLaMa models and SciT\"ulu excel with the \textsc{SciRIFF Mix}, benefiting from pre-training on \textsc{SciRIFF} and \textsc{T\"ulu}. Gemma and Qwen achieve their best results with the \textsc{T\"ulu Mix}, with Qwen outperforming Gemma (45.5 vs. 31.6 S.A. accuracy), likely due to Qwen's multilingual training on 2.4T tokens~\cite{bai2023qwentechnicalreport} compared to Gemma's English-only training~\cite{gemmateam2024gemmaopenmodelsbased}. Mistral and Yi perform best with the full dataset mix, with Yi surpassing Mistral (35.7 vs. 26.8 S.A.), possibly due to its larger vocabulary size. Qwen and Yi lead in fine-grained classification, while SciT\"ulu excels in coarse classification, consistent with earlier findings. Instruction tuning's effectiveness relies on data composition; blending general-purpose instruction data from \textsc{T\"ulu} with scientific data from \textsc{SciRiFF} optimizes LLM performance~\cite{shi2023specialist} on this task. However, including tasks from all sources (\textsc{Full Mix}) can occassionally underperform, likely due to negative task transfer~\cite{pmlr-v202-jang23a, wadden2024sciriffresourceenhancelanguage}.

We train models using optimal data mixes from our cross-validation experiments to obtain \textbf{silver annotations} for the rest of the 1,276 review segments. We test the performance on the \hldb{annotated instances from previous rounds} (details in Appendix \S\ref{sec:training}). We ensure no leakage between the annotation rounds and the training data. The instruction-tuned models' results for fine-grained classification (Table \ref{tab:instruct_tune_rounds}), demonstrate significant improvements over zero-shot and few-shot baselines (best results) from previous rounds (full results in Tables~\ref{tab:ann_round_llm_fine-huge} and \ref{tab:ann_round_llm_coarse}). Qwen achieves the best performance, with gains up to 31 pp. for SciT\"ulu. Coarse-grained classification also shows positive increments, though improvements are smaller (Table \ref{tab:instruct_tune_rounds}). This highlights the effectiveness of instruction-tuning in guiding models towards accurate outputs~\cite{wadden2024sciriffresourceenhancelanguage, ivison2023camelschangingclimateenhancing}. We use the instruction-tuned Qwen model to label the rest of the 1,276 review segments and release these silver annotations as a part of our dataset. Detailed analyses on this portion are provided in Appendix~\S\ref{sec:silver}.

\subsection*{RQ4: How effective are \emph{lazy thinking} guidelines in improving review quality?}
\noindent \textbf{\hl{Setup.}} To improve review quality by addressing \emph{lazy thinking}, we focus on assessing peer reviews created with and without \emph{lazy thinking} annotations. In a controlled experiment, we form two treatment groups, each with two PhD students experienced in NLP peer reviewing. We sample 50 review reports from \textsc{NLPeer} for which we have explicit \emph{lazy thinking} annotations. One group rewrites these reviews based on the current ARR guidelines~\cite{Rogers_Augenstein_2021}, while the other group rewrites the same reviews using the same guidelines with our \emph{lazy thinking} annotations included. Following previous studies~\cite{yuan2022can, sun2024reviewflow}, we evaluate the reviews based on \hldb{Constructiveness}, ensuring actionable feedback is provided, and \hldb{Justified}, requiring clear reasoning for arguments. Additionally, we introduce \hldb{Adherence} to assess how well the reviews align with the given guidelines. A senior PostDoc and a PhD compare the reviews from both groups and the original reviews pairwise. We split the reviews into equal portions~(25 reviews each) while keeping 10 reviews for calculating agreements.\footnote{Annotation instructions are provided in Appendix~\S\ref{sec:instruction}}

\noindent \textbf{\hl{Results.}} The `win-tie-loss' results in Table~\ref{tab:win-tie-loss} show that reviews using \emph{lazy thinking} signals outperform original reviews across all measures, with 90\% win rates for adherence and 85\% for constructiveness and justification. This suggests that \emph{lazy thinking} annotations help to provide actionable, evidence-backed feedback. When compared to guideline-based rewrites (75\% for adherence and 70\% for constructiveness and justified), the win rates are lower. This is likely because reviewers often exercised greater caution when rewriting reviews based on guidelines by shifting concerns from the weakness section to the comments section to soften criticism rather than fully rewriting the reviews. This approach resulted in more constructive and justified feedback. 
We also train a Bradley-Terry preference ranking model~\cite{19ff28b9-64f9-3656-ba40-08326a05748e} on adherence preference data, which reveals strengths of 1.6 for \emph{lazy thinking} rewrites, -1.5 for original reviews, and 0.4 for guideline-based rewrites. This translates to a 95.6\% win rate for \emph{lazy thinking} rewrites over original texts and 76.8\% over guideline-based rewrites, further confirming the effectiveness of these annotations. We obtain Krippendorff's $\alpha$ scores of 0.62, 0.68, and 0.72 for constructiveness, justified, and adherence respectively. 
\begin{table}[!t]
\centering
\resizebox{0.8\columnwidth}{!}{%
\small{\begin{tabular}{llll}
\hline
\multirow{2}{*}{Type}      & Constr.      & Justi.       & Adh.         \\ \cline{2-4} 
                           & W/T/L & W/T/L & W/T/L \\ \hline
Orig. vs \emph{lazy}                   &  \textbf{85}/5/10            &  \textbf{85}/10/5            & \textbf{90}/5/5             \\
Orig $w.$ gdl vs \emph{lazy}               &  70/5/25            &  70/5/25            &   75/5/20           \\ \hline
\end{tabular}}
}
\vspace{-2.5mm}
\caption{Pair-wise comparison of rewrites based on Win (W), Tie (T), and Loss (L) rates across metrics. The first row compares \emph{lazy thinking} rewrites (\emph{lazy}) with original reviews (Orig), while the second compares \emph{lazy} rewrites with guideline-based rewrites (Orig $w.$ gdl).}
\vspace{-6.5mm}
\label{tab:win-tie-loss}
\end{table}
\section{Related Work}
\noindent \textbf{\hl{Rule Following.}} Rules are essential for everyday reasoning~\cite{geva2021did, wang2024can}, with much research focusing on evaluating rule-following behavior in generated outputs. Prior studies have examined inferential logic in question-answering~\cite{wang2022lsat, wang2022logic, ijcai2024p399, sun2024beyond} and detecting forbidden tokens in red-teaming~\cite{mu2023can}. However, identifying rule adherence in peer-review reports requires more abstract reasoning than in typical rule-based scenarios which makes our work more challenging.

\noindent \textbf{\hl{Review Quality Analysis.}} As the number of publications continues to steeply grow, interest in assessing review quality has grown significantly within the academic community~\cite{kuznetsov2024can}. Previous studies have examined various aspects of review reports, including the importance of politeness and professional etiquette~\cite{bharti2023politepeer, bharti2024please}, thoroughness~\cite{severin2022journal, guo-etal-2023-automatic}, and comprehensiveness~\cite{yuan2022can}. There have been efforts to automate the peer-reviewing process using large language models~\cite{du-etal-2024-llms, zhou-etal-2024-llm}. However, there has been no focus on analyzing the usage of heuristics within the reviews. This work focuses on \emph{lazy thinking}, a common issue affecting the quality of peer reviews in Natural Language Processing (NLP). While \citet{rogers-augenstein-2020-improve} have qualitatively analyzed heuristics in NLP conferences, we are the first to present a dedicated dataset for identifying such practices and to propose automated methods for detecting low-quality reviews, thereby enhancing the effectiveness of the peer-review process in NLP.

\section{Conclusion} Aiming to address the pressing issue of \emph{lazy thinking} in review reports, we have presented \textsc{LazyReview}, a novel resource for detecting \emph{lazy thinking} in peer reviews. We conduct extensive experiments and establish strong baselines for this task. Additionally, our controlled experiments demonstrate that providing explicit \emph{lazy thinking} signals to reviewers can significantly improve review quality. We hope our work will inspire further efforts to improve the overall quality of peer reviews.



\section*{Limitations}
    In this work, we introduce \textsc{LazyReview}, a novel resource for detecting \emph{lazy thinking} in NLP peer reviews. While the resource is aimed at promoting sound reviewing practices and enhancing overall review quality, it has some limitations. First, we adopt the definition and categories of \emph{lazy thinking} from the ARR guidelines~\cite{Rogers_Augenstein_2021} and the EMNLP Blog~\cite{LiuCohnEtAl_2020_Advice_on_Reviewing_for_EMNLP}, which are specific to NLP conference reviews. Therefore, the resource \textbf{does not encompass} peer-reviewing venues beyond ARR, and it would need adaptation to suit other domains and conferences such as ICLR with differing definitions of \emph{lazy thinking}. Second, from a task and modeling perspective, we focus on the under-explored area of detecting \emph{lazy thinking} in peer reviews. Although we conduct thorough experiments and analyses, the findings may not be fully generalizable to other classification tasks in the scientific domain, so results should be interpreted with caution. Additionally, ACL ARR guidelines currently characterize \emph{lazy thinking} specifically within the weakness section of a review; future research could explore how such patterns manifest themselves in other parts of the review as well. In addition, lazy thinking patterns may extend into subsequent author–reviewer discussions. However, as our starting dataset, \textsc{NLPeer}, does not include these interactions, we refrain from exploring this aspect, leaving it as a potential direction for future investigation. Furthermore, this study focuses exclusively on reviews authored prior to the widespread adoption of large language models (i.e., before 2023). We hypothesize that the \emph{lazy thinking} framework can be helpful in discriminating between human-written and AI-generated reviews.

\section*{Ethics Statement}
Our dataset, \textsc{LazyReview} will be released under the CC-BY-NC 4.0 licenses. The underlying ARR reviews have been collected from \textsc{NLPeer}~\cite{dycke-etal-2023-nlpeer}, which is also licensed under the  CC-BY-NC-SA 4.0 license. The analysis and automatic annotation do not require processing any personal or sensitive information. We prioritize the mental health of the annotators and ensure that each annotator takes a break every hour or whenever they feel unwell. 

\section*{Acknowledgements}
This work has been funded by the German Research Foundation (DFG) as part of the Research Training Group KRITIS No. GRK 2222. We gratefully acknowledge the support of Microsoft with a grant for access to OpenAI GPT models via the Azure cloud (Accelerate Foundation Model Academic Research). The work of Anne Lauscher is funded under the Excellence Strategy of the German Federal Government and the Federal States. 

We want to thank Viet Pham, Yu Liu, Hao Yang, and Haolan Zhan for their help with the annotation efforts. We are grateful to Qian Ruan and Gholamreza (Reza) Haffari for their initial feedback on a draft of this paper.

\bibliography{acl_latex}

\appendix
\clearpage
\newpage
\section{Appendix} \label{sec:appendix}
\begin{figure*}[!htb]
\centering
\includegraphics[height=0.5cm,keepaspectratio]{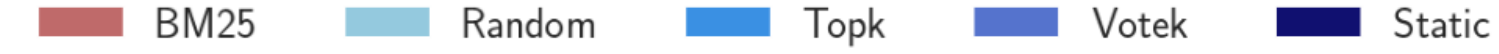}\\
    \begin{subfigure}[b]{0.23\textwidth}
        \includegraphics[width=\textwidth]{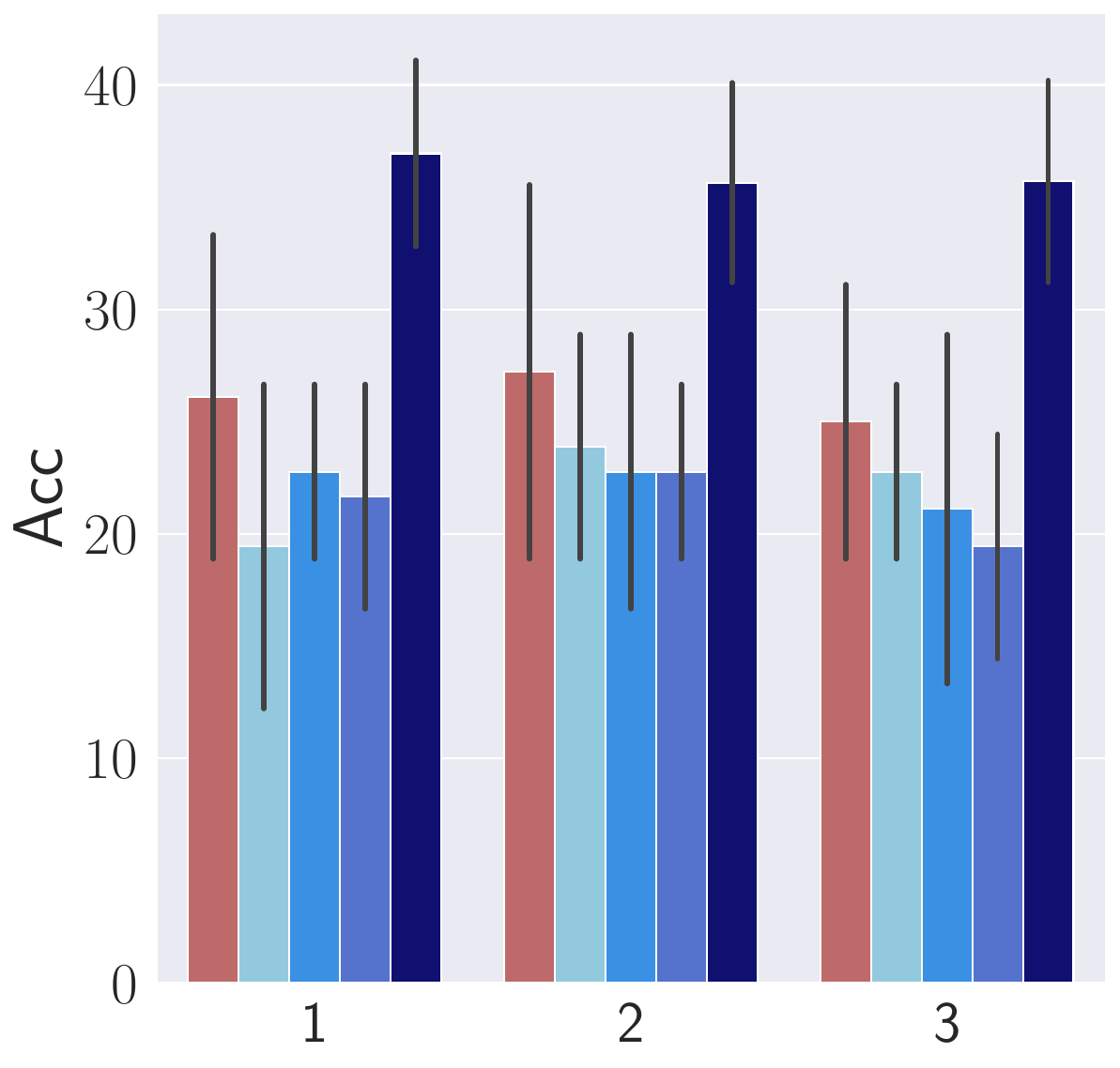}
        \caption{Gemma 7B}
        \label{fig:gemma}
        \end{subfigure}%
        \hspace{1em}%
        \begin{subfigure}[b]{0.23\textwidth}
        \includegraphics[width=\textwidth]{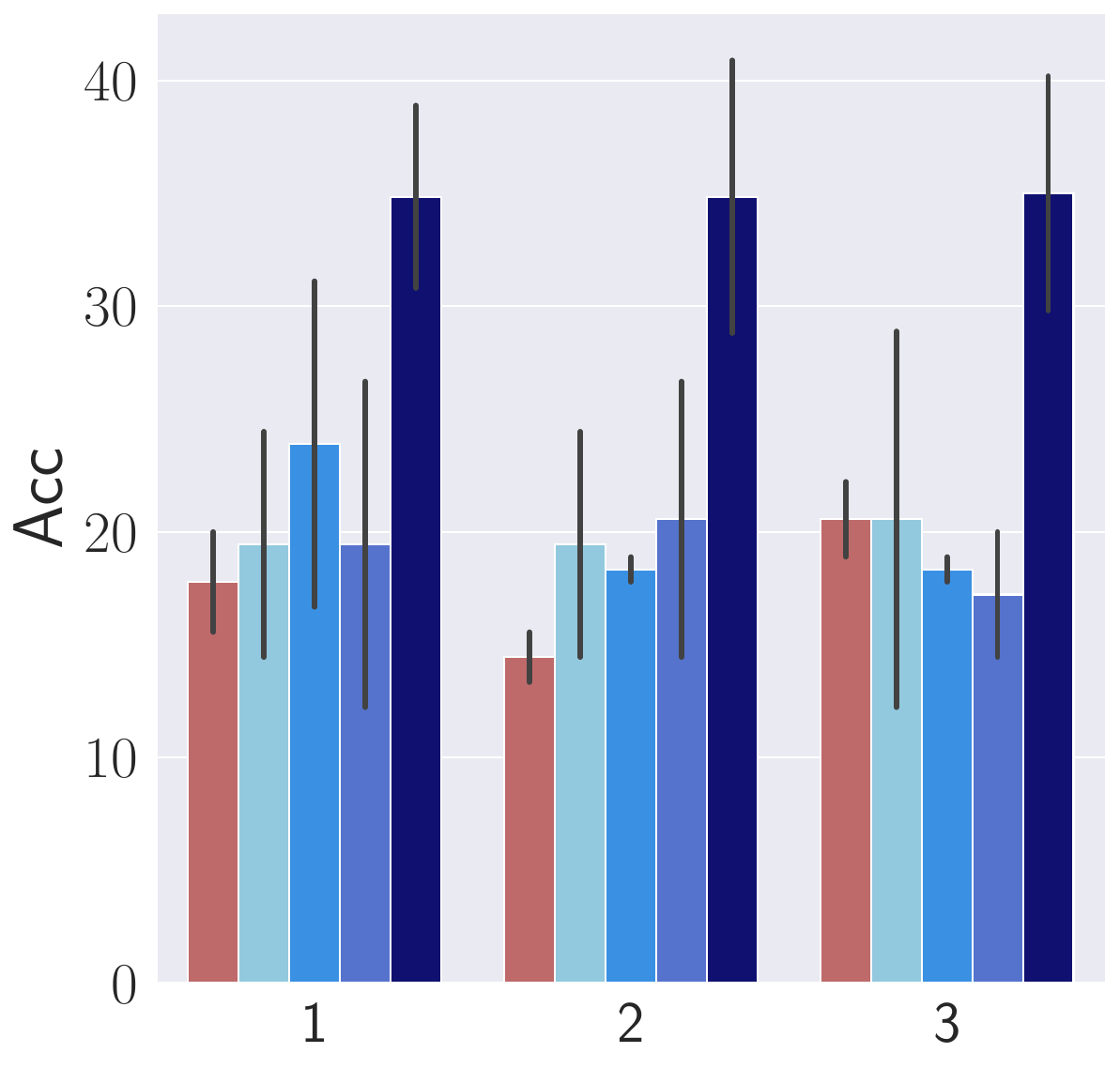}
        \caption{LLaMa 7B}
        \label{fig:llama}
        \end{subfigure}%
         \hspace{1em}%
        \begin{subfigure}[b]{0.23\textwidth}
         \includegraphics[width=\textwidth]{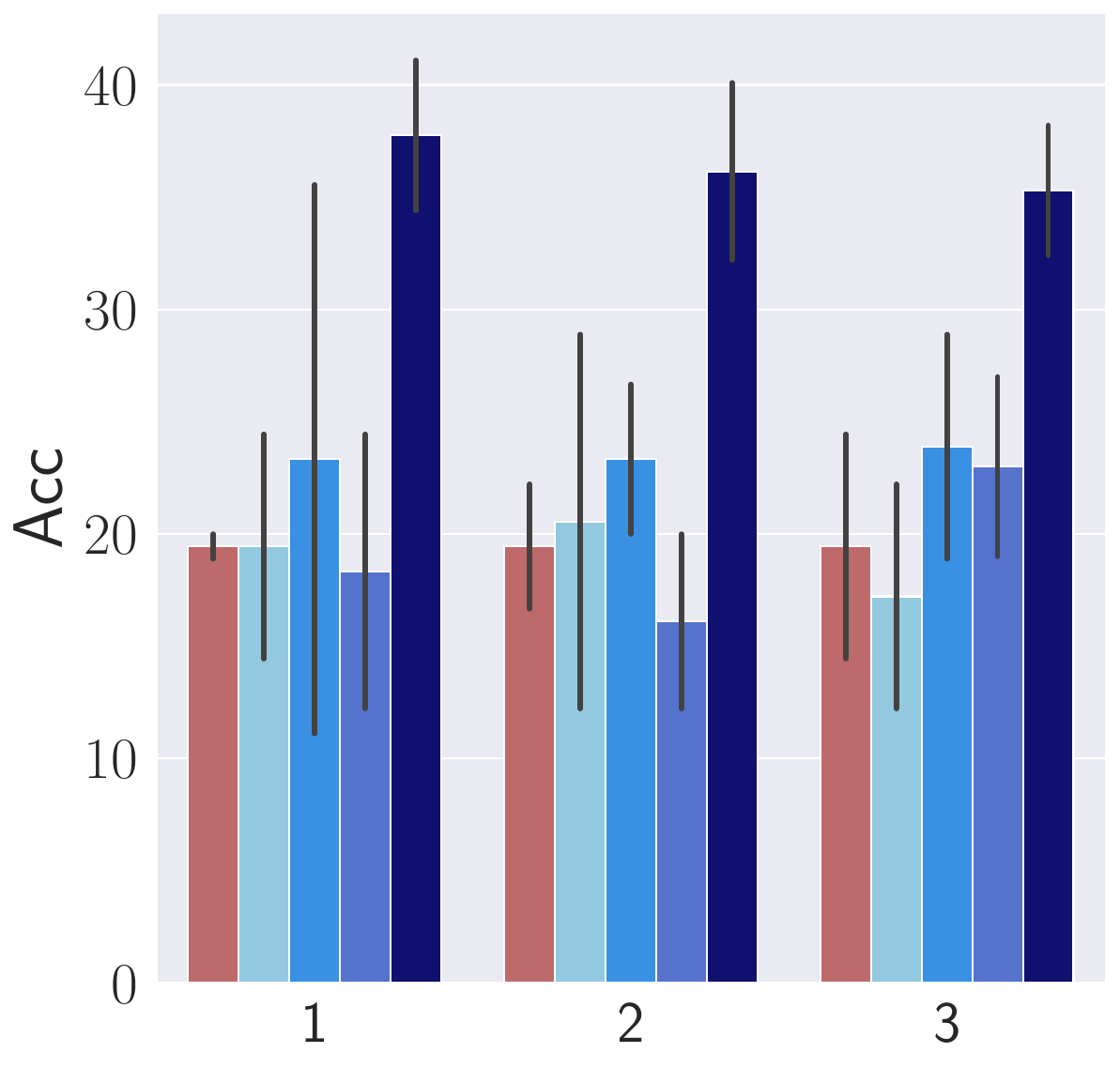}
        \caption{LLaMa 13B}
        \label{fig:llama13}
        \end{subfigure}%
        \hspace{1em}%
         \begin{subfigure}[b]{0.23\textwidth}
         \includegraphics[width=\textwidth]{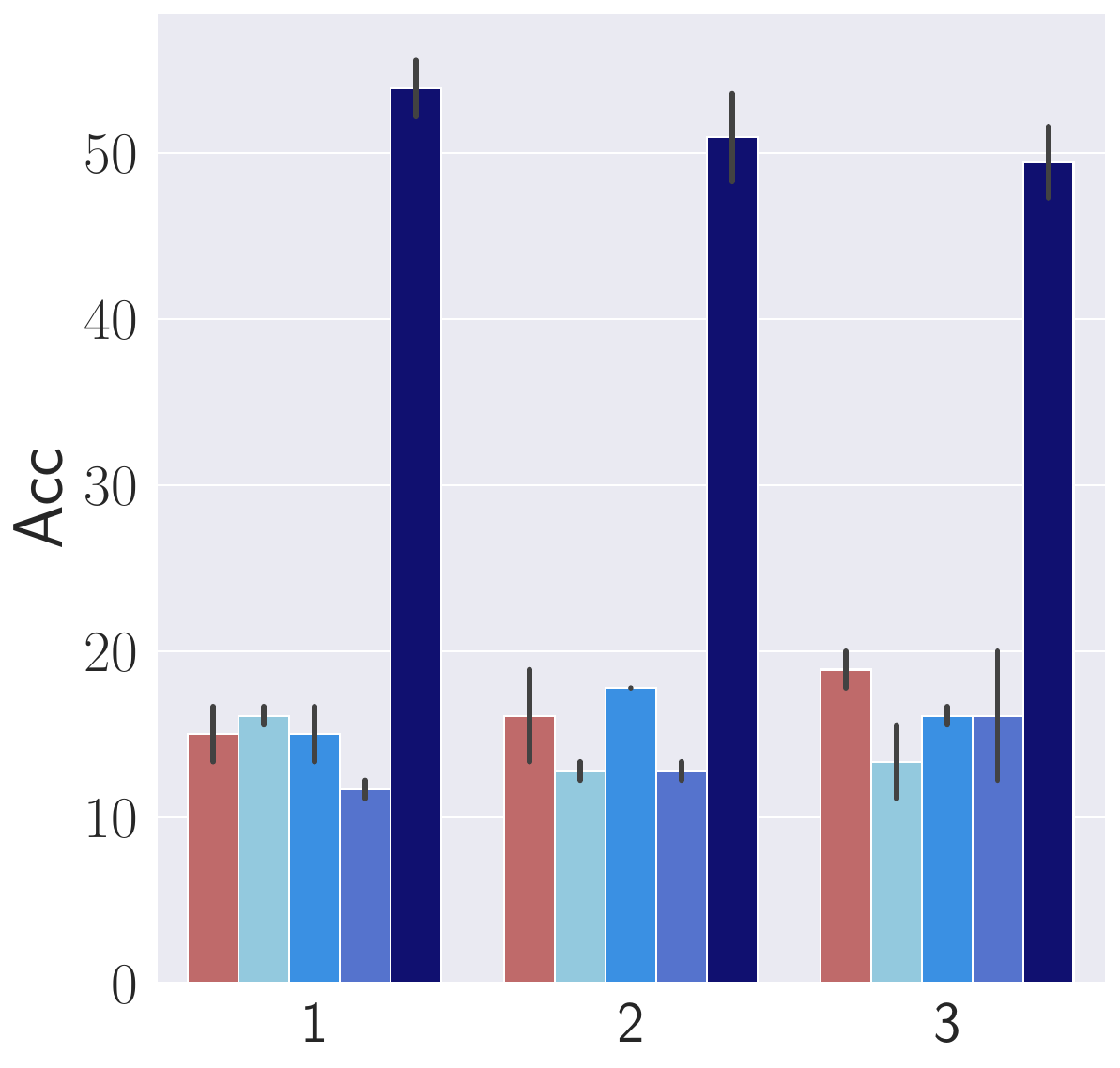}
        \caption{Mistral 7B}
        \label{fig:mistral}
        \end{subfigure}
        \hspace{1em}%
        
         \begin{subfigure}[b]{0.23\textwidth}
         \includegraphics[width=\textwidth]{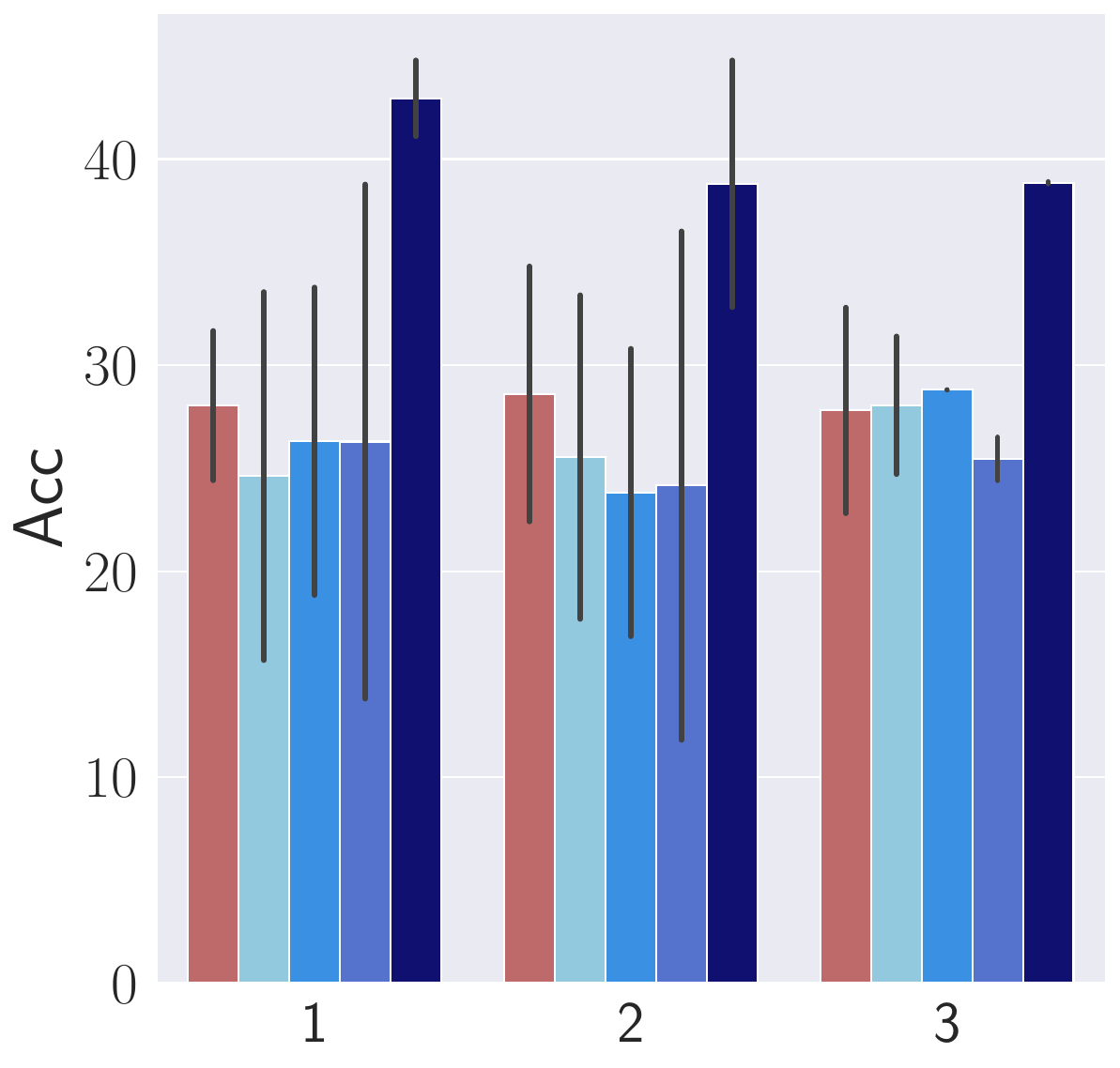}
        \caption{SciT\"ulu 7B}
        \label{fig:scitulu}
        \end{subfigure}%
        \hspace{1em}%
        \begin{subfigure}[b]{0.23\textwidth}
         \includegraphics[width=\textwidth]{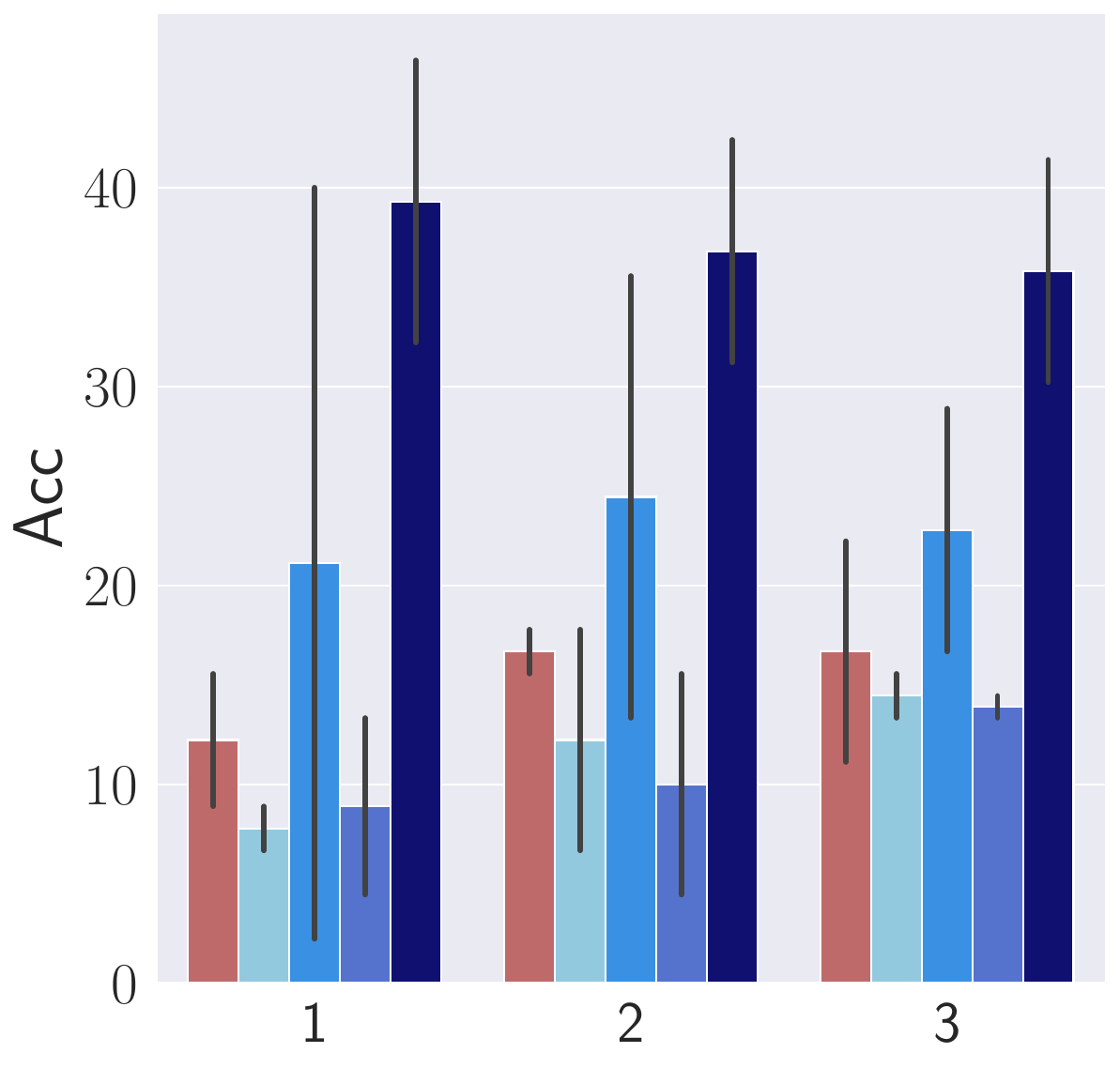}
        \caption{Qwen 7B}
        \label{fig:qwen}
        \end{subfigure}
         \hspace{1em}%

    \caption{Performance of LLMs on using different In Context learning (ICL) methods for Round 3 of our annotation study for fine-grained classification. Error bars indicate using only the target segment (T) as the information source. Acc refers to using GPT-based accuracy}
    \label{fig:models_icl}
\end{figure*}
\begin{figure*}[!htb]
\centering
\includegraphics[height=0.5cm,keepaspectratio]{images/legend2.pdf}\\
    \begin{subfigure}[b]{0.23\textwidth}
        \includegraphics[width=\textwidth]{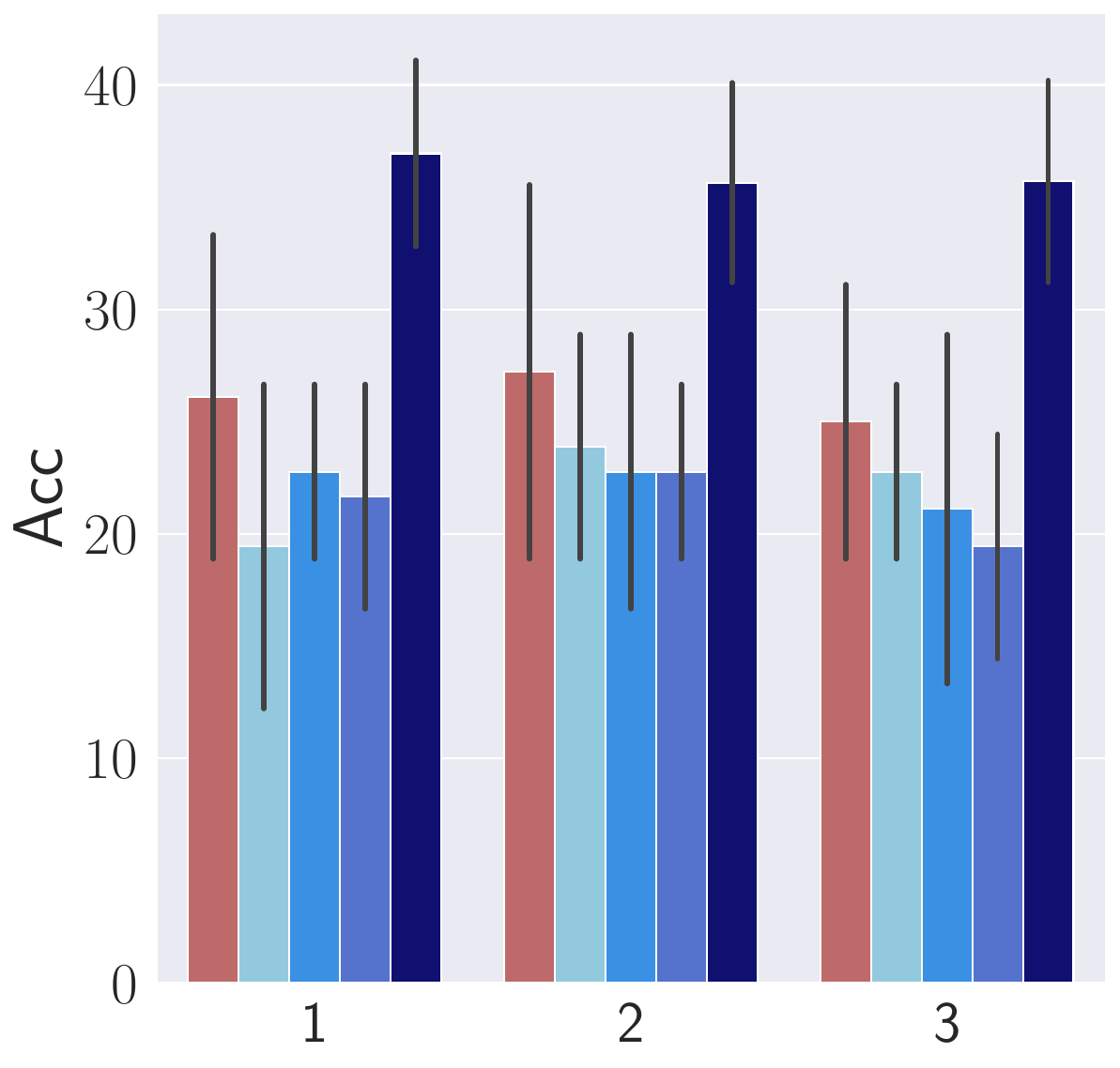}
        \caption{Gemma 7B}
        \label{fig:gemma}
        \end{subfigure}%
        \hspace{1em}%
        \begin{subfigure}[b]{0.23\textwidth}
        \includegraphics[width=\textwidth]{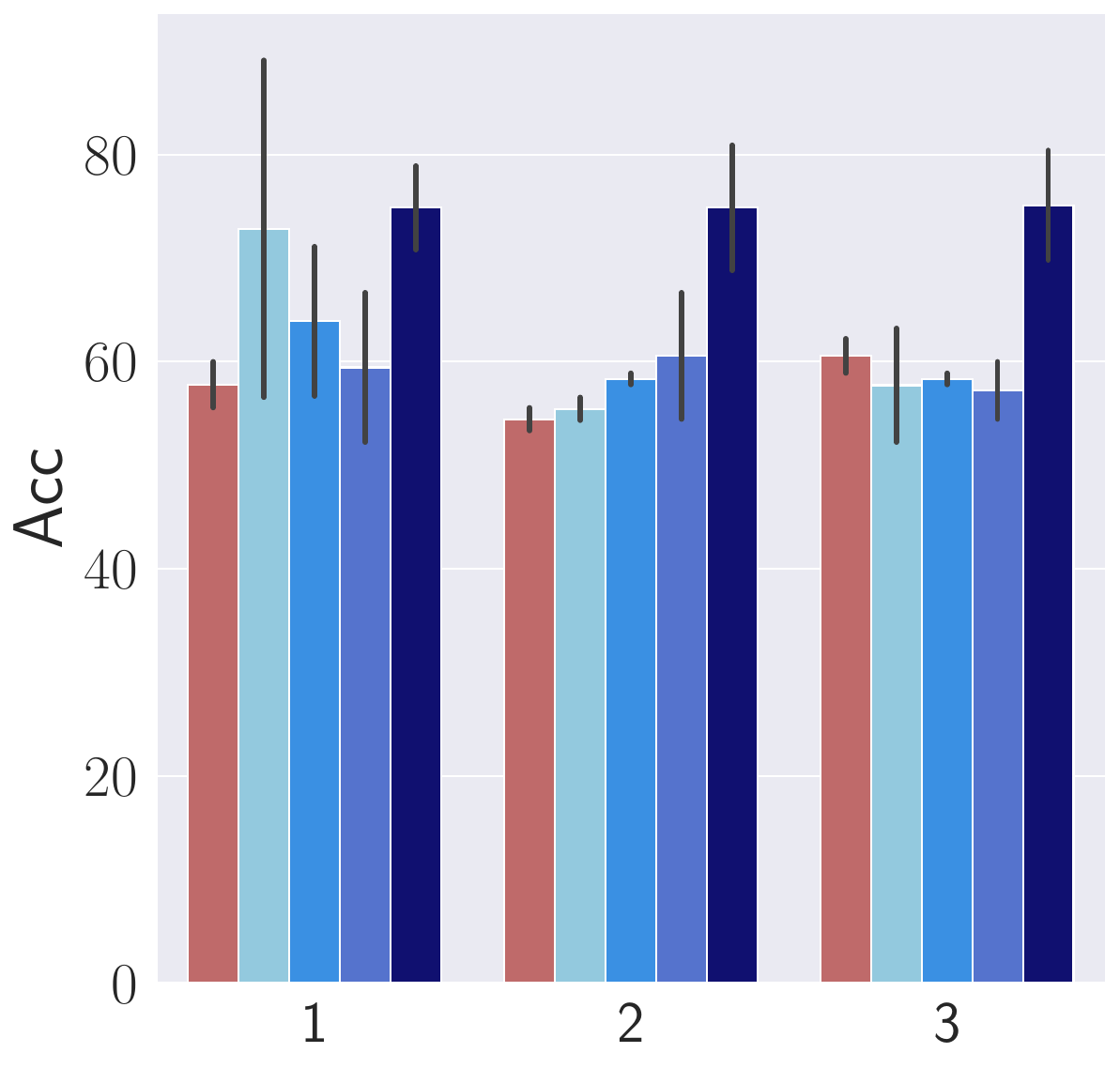}
        \caption{LLaMa 7B}
        \label{fig:llama}
        \end{subfigure}%
         \hspace{1em}%
        \begin{subfigure}[b]{0.23\textwidth}
         \includegraphics[width=\textwidth]{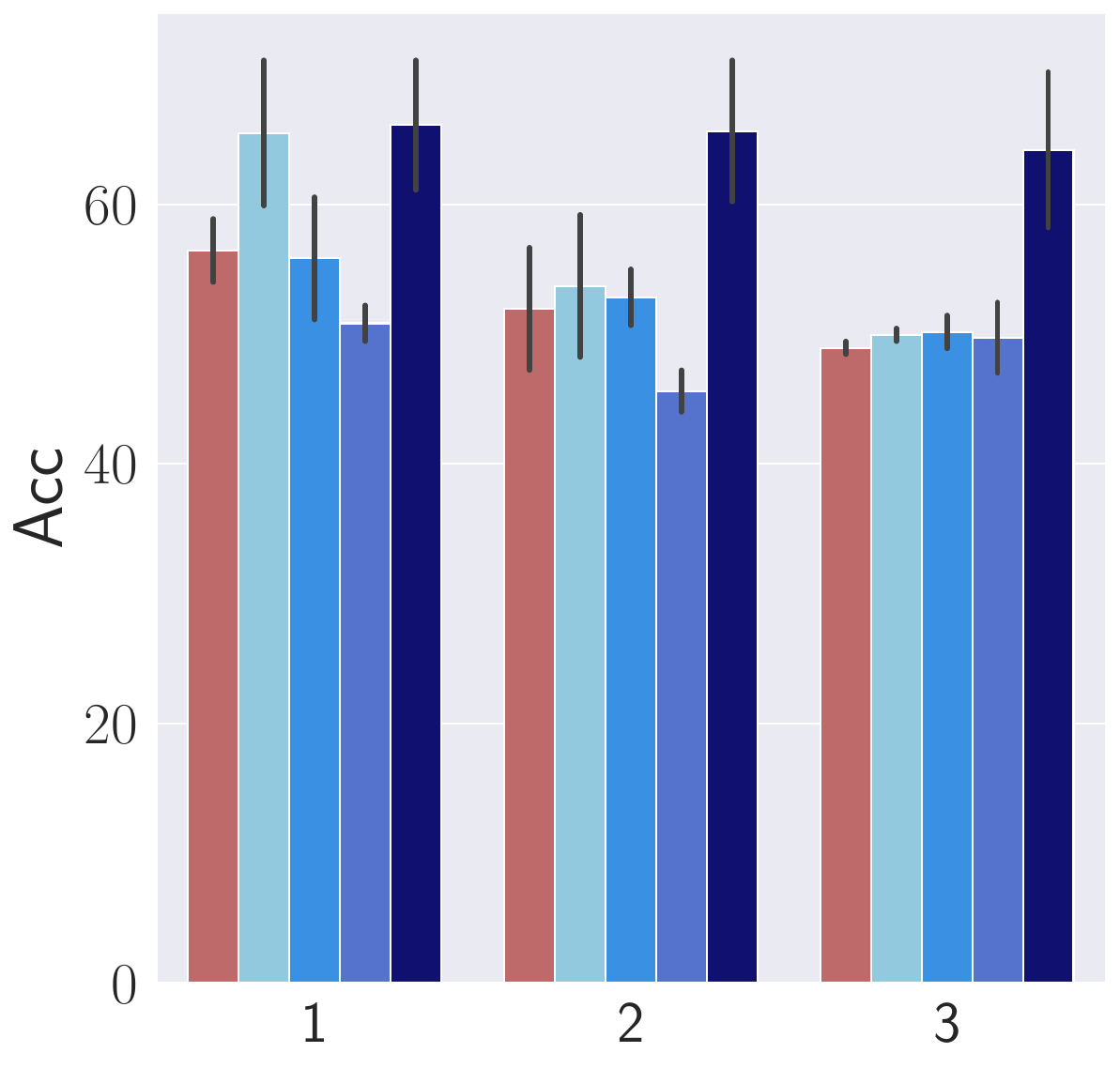}
        \caption{LLaMa 13B}
        \label{fig:llama13}
        \end{subfigure}%
        \hspace{1em}%
         \begin{subfigure}[b]{0.23\textwidth}
         \includegraphics[width=\textwidth]{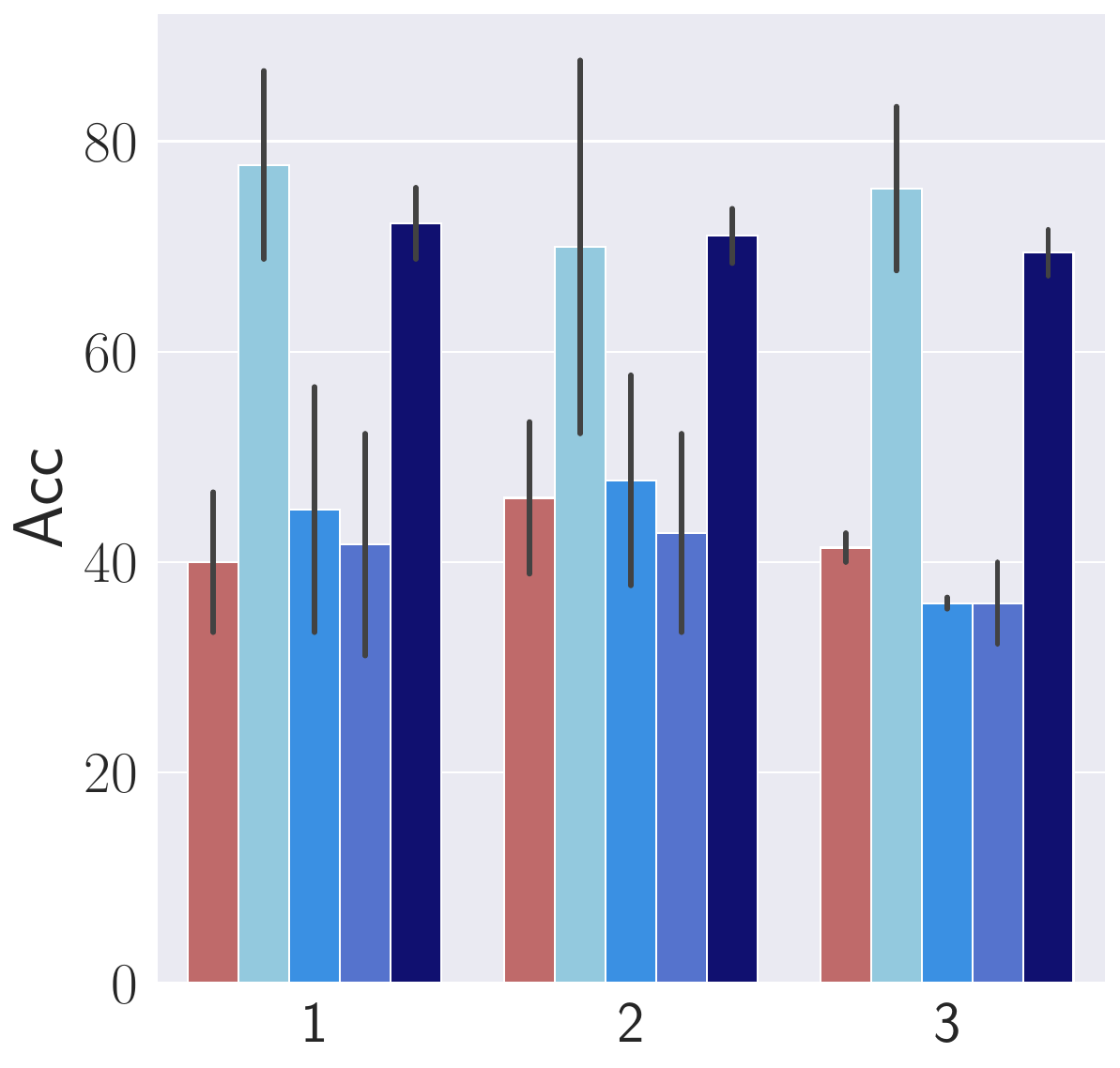}
        \caption{Mistral 7B}
        \label{fig:mistral}
        \end{subfigure}
        \hspace{1em}%
        
         \begin{subfigure}[b]{0.23\textwidth}
         \includegraphics[width=\textwidth]{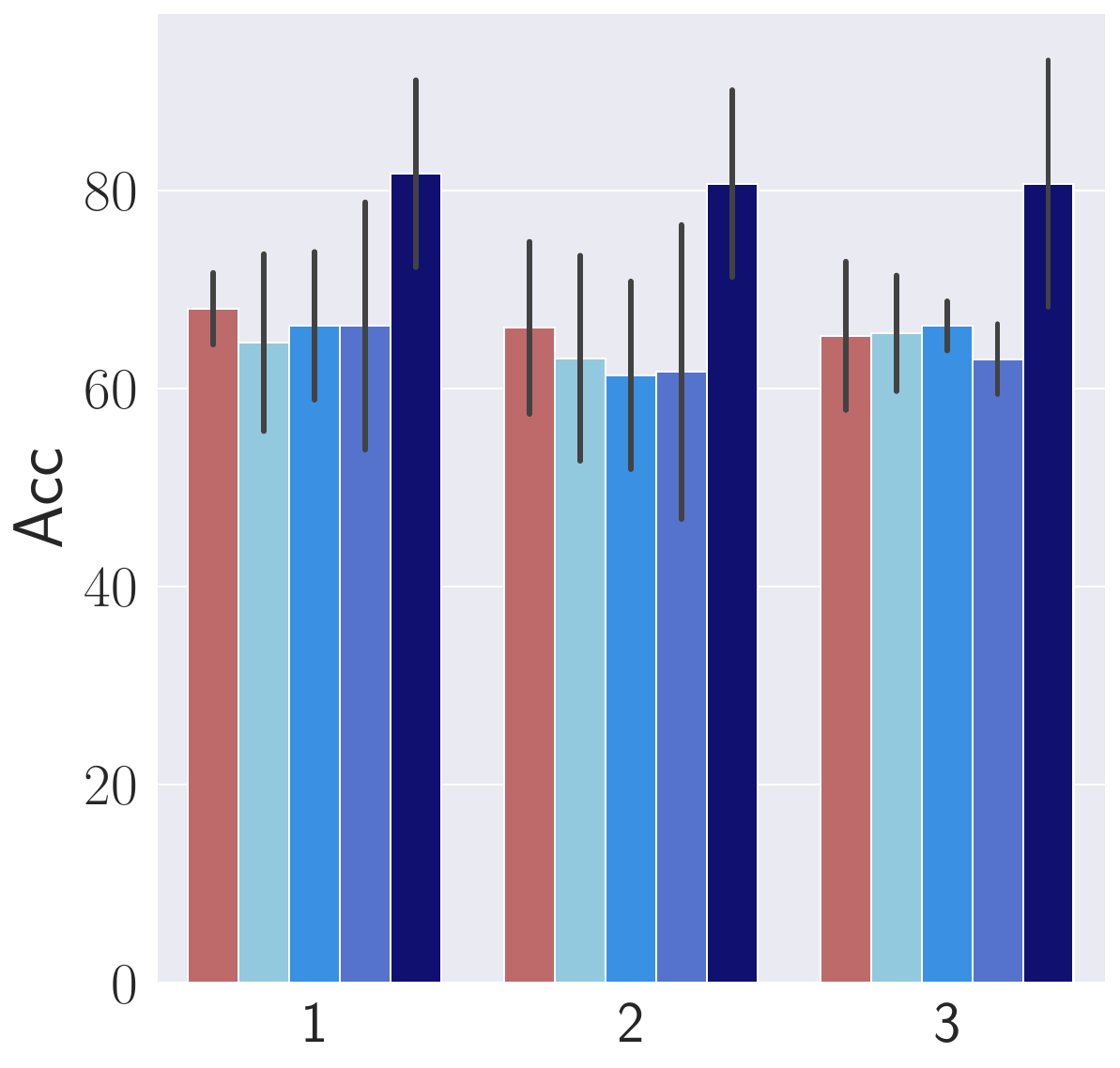}
        \caption{SciT\"ulu 7B}
        \label{fig:scitulu}
        \end{subfigure}%
        \hspace{1em}%
        \begin{subfigure}[b]{0.23\textwidth}
         \includegraphics[width=\textwidth]{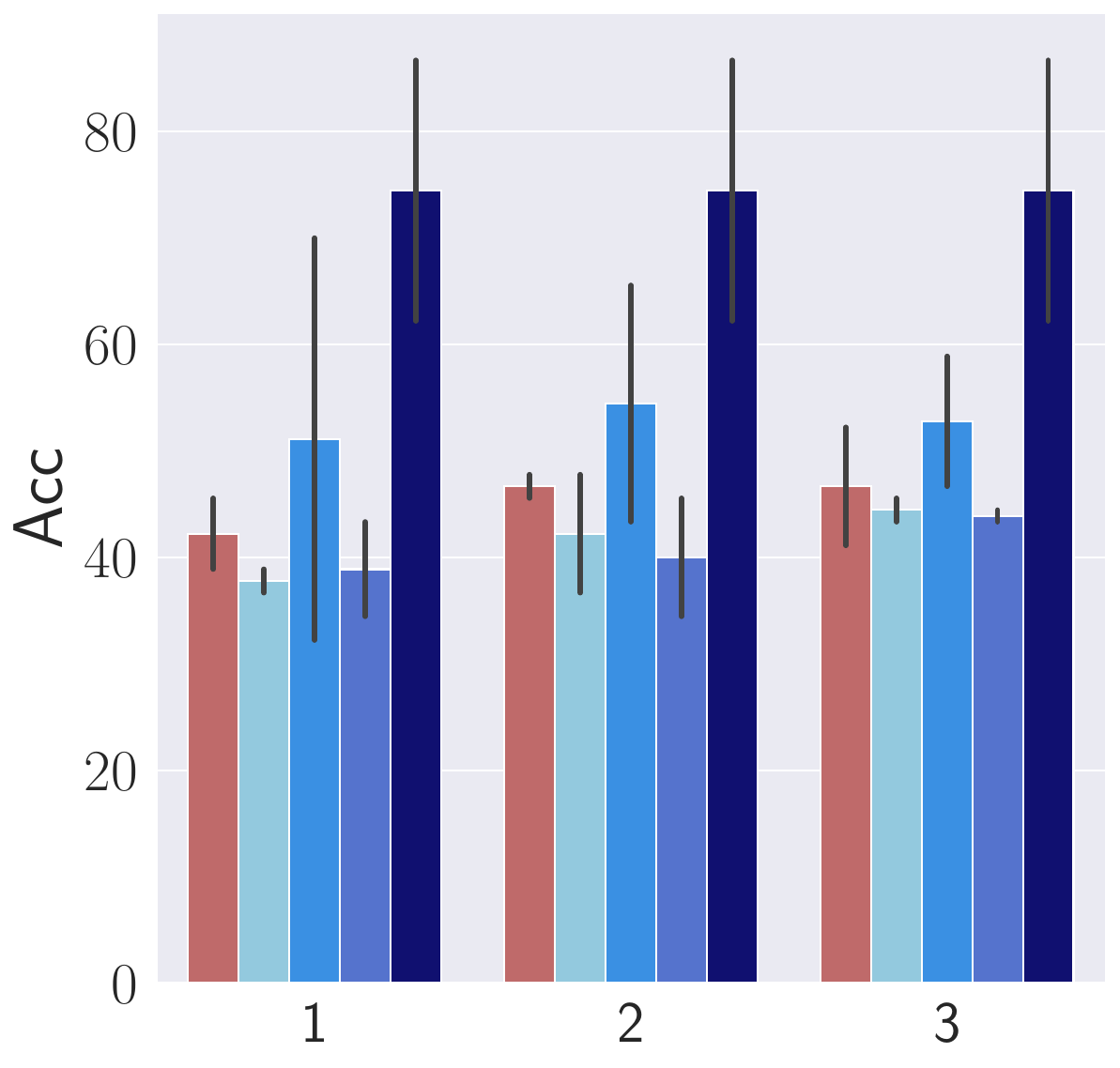}
        \caption{Qwen 7B}
        \label{fig:qwen}
        \end{subfigure}
         \hspace{1em}%

    \caption{Performance of LLMs on using different In Context learning (ICL) methods for Round 3 of our annotation study for the coarse classification task. Error bars indicate using only the target segment (T) as the information source. Acc refers to using GPT-based evaluator accuracy.}
    \label{fig:models_icl_coarse}
\end{figure*}
\begin{figure*}[!ht]
\begin{mybox}{Prompt for GPT-based segment extraction}
\small
\textbf{Prompt:} As the number of scientific publications grows over the years, the need for efficient quality control in peer-reviewing 
becomes crucial. Due to the high reviewing load, the reviewers are often prone to different kinds of bias, which inherently contribute to 
lower reviewing quality and this overall hampers the scientific progress in the field. The ACL peer reviewing guidelines characterize some 
of such reviewing bias.  These are termed lazy thinking. The lazy thinking classes and the reason for them being problematic are provided as a dictionary below: \\

\{'The results are not surprising': 'Many findings seem obvious in retrospect, but this does not mean that the community is already aware of them and can use them as building blocks for future work.',\\ \\
'The results contradict what I would expect': 'You may be a victim of confirmation bias, and be unwilling to accept data contradicting your prior beliefs.',\\ \\
'The results are not novel': 'Such broad claims need to be backed up with references.',\\
'This has no precedent in existing literature': 'Believe it or not: papers that are more novel tend to be harder to publish. Reviewers may be unnecessarily conservative.',\\ \\
'The results do not surpass the latest SOTA', 'SOTA results are neither necessary nor sufficient for a scientific contribution. An engineering paper could also offer improvements on other dimensions (efficiency, generalizability, interpretability, fairness, etc.',\\ \\
'The results are negative': 'The bias towards publishing only positive results is a known problem in many fields, and contributes to hype and overclaiming. If something systematically does not work where it could be expected to, the community does need to know about it.',\\ \\
'This method is too simple': 'The goal is to solve the problem, not to solve it in a complex way. Simpler solutions are in fact preferable, as they are less brittle and easier to deploy in real-world settings.',\\ \\
'The paper doesn't use [my preferred methodology], e.g., deep learning': 'NLP is an interdisciplinary field, relying on many kinds of contributions: models, resource, survey, data/linguistic/social analysis, position, and theory.', \textbf{[...]}\\

\vspace{3pt}
You are given a review, and your task is to identify the segments (1 or more sentences) within the `summary of weaknesses' section that can correspond to any \emph{lazy thinking} class as described above. The output should be a list of these review segments.\\

\textbf{Review:} \hl{paper summary}: This work provides a broad and thorough analysis of how 8 different model families and varying model sizes for a total of 28 models perform on the oLMpics benchmark and the psycholinguistic probing datasets from Ettinger (2020). It finds that all models struggle to resolve compositional questions zero-shot and that attributes such as model size, pretraining objective, etc are not predictive of a model's linguistic capabilities. \hl{summary of strengths}:- The work is well-motivated and clear -A vast selection of models are investigated. \hl{summary of weaknesses}: While the findings are interesting, there is little to no qualitative analysis to provide insight into why these effects might occur -I would expect an analysis such as this to have at least 3 runs with varying random seeds per model to give greater confidence in the model's abilities. A growing body of work indicates that models' linguistic abilities can vary considerably even across initialisations -The exploration and prompt design to adapt the GPT models to the tasks at hand is quite limited. \hl{commnets, suggestions, and typos}: NA

\end{mybox}
\caption{Prompt for GPT based review segment extraction}
\label{fig:gpt_prompt}
\end{figure*}
\subsection{Guidelines used in multiple Annotation Rounds} \label{sec:guidelines}
The existing ARR guidelines~\cite{Rogers_Augenstein_2021} are shown in Tab~\ref{tab:full_arr}, which we used in Round 1 of our annotation study. The guidelines extended using the EMNLP Blog~\cite{LiuCohnEtAl_2020_Advice_on_Reviewing_for_EMNLP} is shown in Table~\ref{tab:arr_guidelines_r2}. The positive examples that we provided the annotators for completing round 3 of our annotation study are shown in Table~\ref{tab:full_arr_positive}.

\subsection{Model and Implementation details} \label{sec:model_details}
We select the LLMs based on multiple criteria: (a) all the models should be open-sourced and privacy-preserving in order to deploy them in real-world reviewing systems, (b) their size should be reasonable in order to perform full fine-tuning using LoRA~\cite{hulora}. c) they must be state-of-the-art on the existing LLM leaderboards~\cite{chiang2024chatbotarenaopenplatform}. Based on these criteria, we select the \textbf{chat} version of the various models shown in Table~\ref{tab:over_models}. We use vLLM for fast inference of all the models.\footnote {\url{https://github.com/vllm-project/vllm}}. We set the temperature as 0 to have consistent predictions throughout and limit the output tokens to 30. 

\subsection{Computational Budget}
We ran all the experiments on Nvidia A100
80GB GPUs. None of the experiments consumed more than 36 hours.
\begin{table}[]
\centering
\resizebox{!}{0.07\textwidth}{\begin{tabular}{lll}
\hline
\textbf{Model}   & \textbf{Size} & \textbf{Link} \\
\hline
LLaMa 2~\cite{touvron2023llama2openfoundation} & 7B   & \url{meta-llama/Llama-2-7b-chat-hf}     \\
LLaMa 2~\cite{touvron2023llama2openfoundation}       & 13B      & \url{meta-llama/Llama-2-13b-chat} \\
Gemma 1.1~\cite{gemmateam2024gemmaopenmodelsbased} & 7B & \url{google/gemma-1.1-7b-it} \\
Mistral v0.1~\cite{jiang2023mistral7b} & 7B & \url{mistralai/Mistral-7B-Instruct-v0.1} \\
Qwen-1.5~\cite{bai2023qwentechnicalreport} & 7B & \url{Qwen/Qwen-7B-Chat} \\
Yi-1.5~\cite{ai2024yiopenfoundationmodels} & 6B & \url{01-ai/Yi-6B-Chat} \\
SciT\"ulu~\cite{wadden2024sciriffresourceenhancelanguage} & 7B & \url{allenai/scitulu-7b} \\ \hline
\end{tabular}}
\caption{Overview of models used in our work along with their sizes and links.}
\label{tab:over_models}
\end{table}

\subsection{Prompt for GPT-based review segment extraction}\label{sec:gpt_prompt} We describe the prompt for GPT-based review segment extraction in Fig~\ref{fig:gpt_prompt}. We provide the classes from the existing ARR guidelines and ask the model to provide us a list of review segments that can likely be related to \emph{lazy thinking}. We show a subset of the classes in the prompt due to space constraints.
\begin{figure*}[!ht]
\begin{mybox}{Prompt Template for Lazy Thinking}
\small
\textbf{Prompt:} As the number of scientific publications grows over the years, the need for efficient quality control in peer-reviewing 
becomes crucial. Due to the high reviewing load, the reviewers are often prone to different kinds of bias, which inherently contribute to 
lower reviewing quality and this overall hampers the scientific progress in the field. The ACL peer reviewing guidelines characterize some 
of such reviewing bias.  These are termed lazy thinking. The lazy thinking classes and the reason for them being problematic are provided as a dictionary below: \\

\{'The results are not surprising': 'Many findings seem obvious in retrospect, but this does not mean that the community is already aware of them and can use them as building blocks for future work.',\\ \\
'The results contradict what I would expect': 'You may be a victim of confirmation bias, and be unwilling to accept data contradicting your prior beliefs.',\\ \\
'The results are not novel': 'Such broad claims need to be backed up with references.',\\
'This has no precedent in existing literature': 'Believe it or not: papers that are more novel tend to be harder to publish. Reviewers may be unnecessarily conservative.',\\ \\
'The results do not surpass the latest SOTA', 'SOTA results are neither necessary nor sufficient for a scientific contribution. An engineering paper could also offer improvements on other dimensions (efficiency, generalizability, interpretability, fairness, etc.',\\ \\
'The results are negative': 'The bias towards publishing only positive results is a known problem in many fields, and contributes to hype and overclaiming. If something systematically does not work where it could be expected to, the community does need to know about it.',\\ \\
'This method is too simple': 'The goal is to solve the problem, not to solve it in a complex way. Simpler solutions are in fact preferable, as they are less brittle and easier to deploy in real-world settings.',\\ \\
\textbf{[...]}

\end{mybox}
\caption{Fixed Prompt for defining lazy thinking}
\label{fig:prompt}
\end{figure*}

\begin{figure*}[!ht]
\begin{mybox}{Prompt Template for Fine-Grained Classification}
\small

\textbf{Task:} Given a full review and a target segment corresponding to that review, you need to classify the target sentence into one of the lazy thinking classes 
eg., 'The topic is too niche', 'The results are negative'.

\vspace{3pt}

\textbf{Full Review:} \texttt{<review>} \\
\textbf{Target Segment:} \texttt{<target segment>}

\end{mybox}

\caption{Prompt for fine-grained classification}
\label{fig:prompt_fine}
\end{figure*}

\begin{figure*}[!ht]
\begin{mybox}{Prompt Template for Coarse-Grained Classification}
\small

\textbf{Task:} Given a full review and a target segment corresponding to that review, you need to classify the target sentence into whether it is `lazy thinking' or not `lazy thinking'

\vspace{3pt}

\textbf{Full Review:} \texttt{<review>} \\
\textbf{Target Segment:} \texttt{<target segment>}

\end{mybox}

\caption{Prompt for coarse-grained classification}
\label{fig:prompt_coarse}
\end{figure*}

\begin{figure*}[!ht]
\begin{mybox}{Prompt Template for Fine-Grained Classification (ICL Based)}
\small

\textbf{Task:} Given a full review and a target segment corresponding to that review, you need to classify the target sentence into one of the lazy thinking classes 
eg., 'The topic is too niche', 'The results are negative'. An example is shown below.

\vspace{3pt}
\textbf{Full Review:} \hl{paper summary}: This work provides a broad and thorough analysis of how 8 different model families and varying model sizes for a total of 28 models perform on the oLMpics benchmark and the psycholinguistic probing datasets from Ettinger (2020). It finds that all models struggle to resolve compositional questions zero-shot and that attributes such as model size, pretraining objective, etc are not predictive of a model's linguistic capabilities. \hl{summary of strengths}:- The work is well-motivated and clear -A vast selection of models are investigated. \hl{summary of weaknesses}: While the findings are interesting, there is little to no qualitative analysis to provide insight into why these effects might occur -I would expect an analysis such as this to have at least 3 runs with varying random seeds per model to give greater confidence in the model's abilities. A growing body of work indicates that models' linguistic abilities can vary considerably even across initialisations -The exploration and prompt design to adapt the GPT models to the tasks at hand is quite limited. \hl{commnets, suggestions, and typos}: NA \\
\textbf{Target Segment:} \hl{The exploration and prompt design to adapt the GPT models to the tasks at hand is quite limited.}\\
\textbf{Class:} \hl{The authors should do extra experiment [X]} \\
\textbf{Full Review:} \texttt{<review>} \\
\textbf{Target Segment:} \texttt{<target segment>}

\end{mybox}

\caption{Prompt for fine-grained classification based on In-Context Learning (ICL) as used in Round 3 of our study}
\label{fig:prompt_fine_icl}
\end{figure*}

\begin{figure*}[!ht]
\begin{mybox}{Prompt Template for Coarse-Grained Classification (ICL Based)}
\small

\textbf{Task:} Given a full review and a target segment corresponding to that review, you need to classify the target sentence into whether it is `lazy thinking' or not `lazy thinking'

\vspace{3pt}
\textbf{Full Review:} \hl{paper summary}: This work provides a broad and thorough analysis of how 8 different model families and varying model sizes for a total of 28 models perform on the oLMpics benchmark and the psycholinguistic probing datasets from Ettinger (2020). It finds that all models struggle to resolve compositional questions zero-shot and that attributes such as model size, pretraining objective, etc are not predictive of a model's linguistic capabilities. \hl{summary of strengths}:- The work is well-motivated and clear -A vast selection of models are investigated. \hl{summary of weaknesses}: While the findings are interesting, there is little to no qualitative analysis to provide insight into why these effects might occur -I would expect an analysis such as this to have at least 3 runs with varying random seeds per model to give greater confidence in the model's abilities. A growing body of work indicates that models' linguistic abilities can vary considerably even across initialisations -The exploration and prompt design to adapt the GPT models to the tasks at hand is quite limited. \hl{commnets, suggestions, and typos}: NA \\
\textbf{Target Segment:} \hl{The exploration and prompt design to adapt the GPT models to the tasks at hand is quite limited.}\\
\textbf{Class:} \hl{Lazy Thinking} \\
\textbf{Full Review:} \texttt{<review>} \\
\textbf{Target Segment:} \texttt{<target segment>}

\end{mybox}

\caption{Prompt for coarse-grained classification based on In-Context Learning (ICL) as used in Round 3 of our study}
\label{fig:prompt_coarse_icl}
\end{figure*}
\newpage
\begin{table*}[t]
    \centering
    \small{
    \begin{tabularx}{\textwidth}{XX}%
    \toprule
    
    \textbf{Heuristics} & \textbf{Description} \\
    \midrule
    \emph{The results are not surprising} & Many findings seem obvious in retrospect, but this does not mean that the community is already aware of them and can use them as building blocks for future work. \\ \midrule
    \emph{The results contradict what I would expect} & You may be a victim of confirmation bias, and be unwilling to accept data contradicting your prior beliefs. \\ \midrule
    \emph{The results are not novel	} & Such broad claims need to be backed up with references. \\ \midrule
    \emph{This has no precedent in existing literature} & Believe it or not: papers that are more novel tend to be harder to publish. Reviewers may be unnecessarily conservative. \\ \midrule
    \emph{The results do not surpass the latest SOTA} & SOTA results are neither necessary nor sufficient for a scientific contribution. An engineering paper could also offer improvements on other dimensions (efficiency, generalizability, interpretability, fairness, etc.) \\  \midrule
    \emph{The results are negative} & The bias towards publishing only positive results is a known problem in many fields, and contributes to hype and overclaiming. If something systematically does not work where it could be expected to, the community does need to know about it. \\ \midrule
    \emph{This method is too simple} & The goal is to solve the problem, not to solve it in a complex way. Simpler solutions are in fact preferable, as they are less brittle and easier to deploy in real-world settings. \\ \midrule
    \emph{The paper doesn't use [my preferred methodology], e.g., deep learning} & NLP is an interdisciplinary field, relying on many kinds of contributions: models, resource, survey, data/linguistic/social analysis, position, and theory. \\ \midrule
    \emph{The topic is too niche} & A main track paper may well make a big contribution to a narrow subfield. \\ \midrule
 \emph{The approach is tested only on [not English], so unclear if it will generalize to other languages} & The same is true of NLP research that tests only on English. Monolingual work on any language is important both practically (methods and resources for that language) and theoretically (potentially contributing to deeper understanding of language in general). \\ \midrule
    \emph{The paper has language errors} & As long as the writing is clear enough, better scientific content should be more valuable than better journalistic skills. \\ \midrule
    \emph{The paper is missing the comparison to the [latest X]} & Per ACL policy, the authors are not obliged to draw comparisons with contemporaneous work, i.e., work published within three months before the submission (or three months before a re-submission). \\ \midrule
    \emph{The authors could also do [extra experiment X]} & It is always possible to come up with extra experiments and follow-up work. This is fair if the experiments that are already presented are insufficient for the claim that the authors are making. But any other extra experiments are in the ``nice-to-have'' category, and belong in the “suggestions” section rather than ``weaknesses.'' \\ \midrule
    \emph{The authors should have done [X] instead} & A.k.a. ``I would have written a different paper.'' There are often several valid approaches to a problem. This criticism applies only if the authors’ choices prevent them from answering their research question, their framing is misleading, or the question is not worth asking. If not, then [X] is a comment or a suggestion, but not a ``weakness.'' \\ \midrule

    \end{tabularx}}
    \caption{Full ARR 2022 guidelines on \textit{lazy thinking} sourced from \citet{Rogers_Augenstein_2021}.}
    \label{tab:full_arr}
\end{table*}

\begin{table*}[t]
    \centering
    \small{
    \begin{tabularx}{\textwidth}{XX}%
    \toprule
    
    \textbf{Heuristics} & \textbf{Description} \\
    \midrule

    \emph{\hlpink{This has no precedent in existing literature}} & The paper’s topic is completely new, such that there’s no prior art or all the prior art has been done in another field. We are interested in papers that tread new ground.  Believe it or not: papers that are more novel tend to be harder to publish. Reviewers may be unnecessarily conservative.  \\ 
  \midrule
    \emph{\hlpink{This method is too simple}} & The paper’s method is too simple. Our goal is not to design the most complex method. Again, think what the paper’s contributions and findings are. Often the papers with the simplest methods are the most cited. If a simple method outperforms more complex methods from prior work, then this is often an important finding. The goal is to solve the problem, not to solve it in a complex way. Simpler solutions are in fact preferable, as they are less brittle and easier to deploy in real-world settings. \\ \midrule
    \emph{\hlpink{The paper doesn't use [my preferred methodology], e.g., deep learning}} & The paper does not use a particular method (e.g., deep learning). No one particular method is a requirement for good work. Please justify why that method is needed. Think about what the paper’s contributions are, and bear in mind that having a diversity of methods used is not a bad thing. NLP is an interdisciplinary field, relying on many kinds of contributions: models, resource, survey, data/linguistic/social analysis, position, and theory. \\ \midrule
    \emph{\hl{The topic is too niche / Narrow Topics}} &The paper’s topic is narrow or outdated. Please be open minded. We do not want the whole community to chase a trendy topic. Look at the paper’s contributions and consider what impact it may have on our community. It is easier to publish on trendy,
‘scientifically sexy’ topics (Smith, 2010). In the
last two years, there has been little talk of anything other than large pretrained Transformers,
with BERT alone becoming the target of over 150
studies proposing analysis and various modifications (Rogers et al., 2020). The ‘hot trend’ forms the prototype for the kind of paper that should be
recommended for acceptance. Niche topics such as
historical text normalization are downvoted (unless,
of course, BERT could somehow be used for that).
\\ \midrule
\emph{\hlpink{The approach is tested only on [not English], so unclear if it will generalize to other languages}} & The paper’s work is on a language other than English. We care about NLP for any language. The same is true of NLP research that tests only on English. Monolingual work on any language is important both practically (methods and resources for that language) and theoretically (potentially contributing to deeper understanding of language in general). \\ \midrule
    \emph{\hl{The paper has language errors / Writing Style}} & As long as the writing is clear enough, better scientific content should be more valuable than better journalistic skills. \\ \midrule
     \emph{\hldb{Non-mainstream approaches}} & Since a ‘mainstream’ *ACL paper currently uses DL-based methods, anything else might look like it does not really
belong in the main track - even though ACL stands for ‘Association for Computational Linguistics’.
That puts interdisciplinary efforts at a disadvantage, and continues the trend for intellectual segregation
of NLP (Reiter, 2007). E.g., theoretical papers
and linguistic resources should not be a priori at a disadvantage just because they do not contain DL
experiments.\\ \midrule
\emph{\hldb{Resource paper}} & The paper is a resource paper. In a field that relies on supervised machine learning as much as NLP, development of datasets is as important as modeling work. \\

\hline

    \end{tabularx}}
    \caption{Lazy thinking classes with \hl{extended names}, \hlpink{extended descriptions} and \hldb{new additions} used in Round 2 of our annotations. The rest of the class names and descriptions are the same as in  Round 1 of our annotation sourced from \citet{Rogers_Augenstein_2021}.}
    \label{tab:arr_guidelines_r2}
\end{table*}

\begin{table*}[t]
    \centering
    \small{
    \begin{tabularx}{\textwidth}{XX}%
    \toprule
    
    \textbf{Heuristics} & \textbf{Positive examples} \\
    \midrule

    \emph{The results are not surprising} & Although the experiments are very comprehensive, this paper lacks technical novelty. \hl{The optimal data selection strategy also seems to offer limited improvement over random data selection}. \\ \midrule
    \emph{The results contradict what I would expect} &  \hl{Although the paper empirically shows that the baseline is not as effective as the proposed method, - I expect more discussion on why using activation values is not a good idea, this contradicts my prior assumption}.-One limitation of this study is that the paper only focuses on single-word cloze queries (as discussed in the paper) \\ \midrule
    \emph{The results are not novel	} & \hl{The novelty of the approach is limited. The attempt to train the document retrieval and outcome prediction jointly is unsuccessful.} \\ \midrule
    \emph{This has no precedent in existing literature} &  There are a lot of problems that I can imagine for real-world large scale models such as GPT3. To mention a few: (1) \hl{Recent works have shown it is possible to continue training language models for either language understanding [1] or additional applications such as code synthesis [2]. Therefore, perhaps these simple approaches are sufficient enough for good generalization.} \\ \midrule
    \emph{The results are negative} & I have two concerns about the experimental results. 1. In the few-shot learning, when the samples are increased from 24 to 1K, the attribute relevance drops by 2 points for positive class as shown in Table 1, and the toxicity metric becomes worse for the detoxification task as shown in Table 2. Please give some explanations. 2. \hl{In human evaluation, the inter-annotator agreement on the sentiment task and the AGNews 482 task is only 0.39 and 0.30 in Fleiss’ $\kappa$, which is low to guarantee a high quality of the evaluation data.}  \\ \midrule
    \emph{This method is too simple} & \hl{There is unfortunately not a whole lot of new content in this paper. The proposed method really just boils down to playing with the temperature settings for the attention of the teacher model; something that is usually just considered a hyperparameter. While I have no problem with what is currently in the paper, I am just not sure that this is enough to form a long paper proposing a new method.} \\ \midrule
    \emph{The topic is too niche / Narrow Topics} & 1- It is not clear how this task and the approach will perform for new information (updates) about an event (even if it is contradictory to what is known about an event) 2- \hl{The approach operates on clusters. New events may not have clusters}.  \\ \midrule
    \emph{The authors could also do [extra experiment X] } & 1. According to Table 3, the performance of BARTword and BARTspan on SST-2 degrades a lot after incorporating text smoothing, why? 2. \hl{Lack of experimental results on more datasets: I suggest conducting experiments on more datasets to make a more comprehensive evaluation of the proposed method. The experiments on the full dataset instead of that in the low-resource regime are also encouraged.}\\

\hline

    \end{tabularx}}
    \caption{Positive examples used during Round 3 of our annotations. The review segment that corresponds to \emph{lazy thinking} is highlighted. We display only the weakness section of the reviews rather than the full review due to space constraints.}
    \label{tab:full_arr_positive}
\end{table*}

\begin{figure*}[!htb]
\centering
    \begin{subfigure}[b]{0.23\textwidth}
        \includegraphics[width=\textwidth]{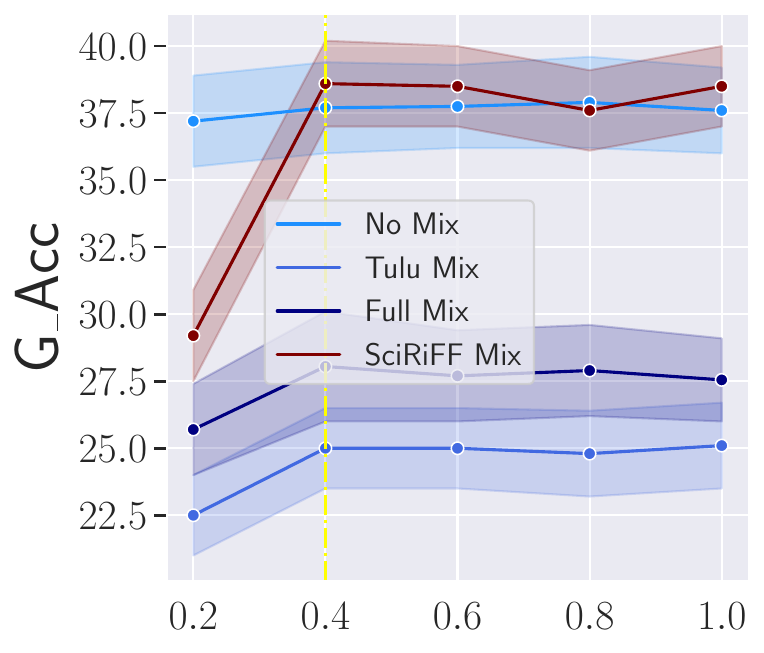}
        \caption{Gemma 7B}
        \label{fig:gemma}
        \end{subfigure}%
        \hspace{1em}%
        \begin{subfigure}[b]{0.23\textwidth}
        \includegraphics[width=\textwidth]{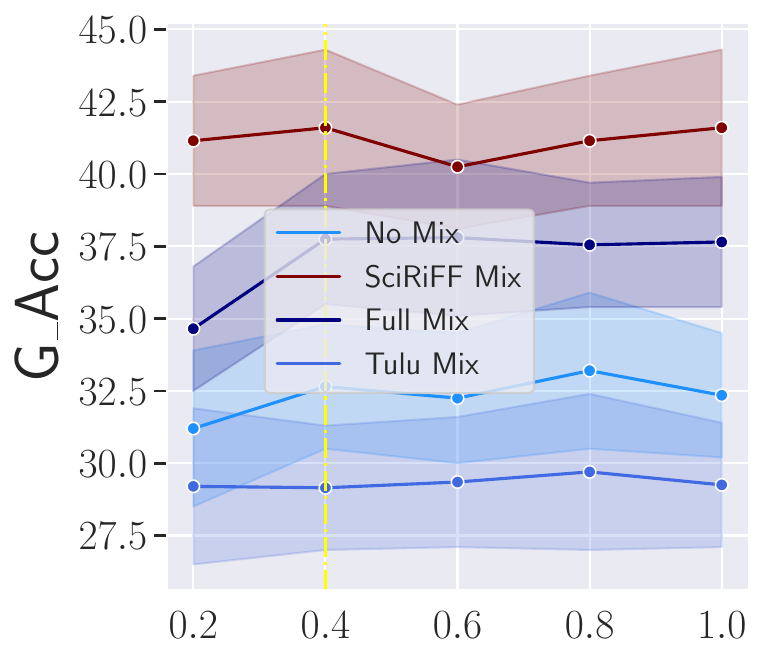}
        \caption{LLaMa 7B}
        \label{fig:llama}
        \end{subfigure}%
         \hspace{1em}%
        \begin{subfigure}[b]{0.23\textwidth}
         \includegraphics[width=\textwidth]{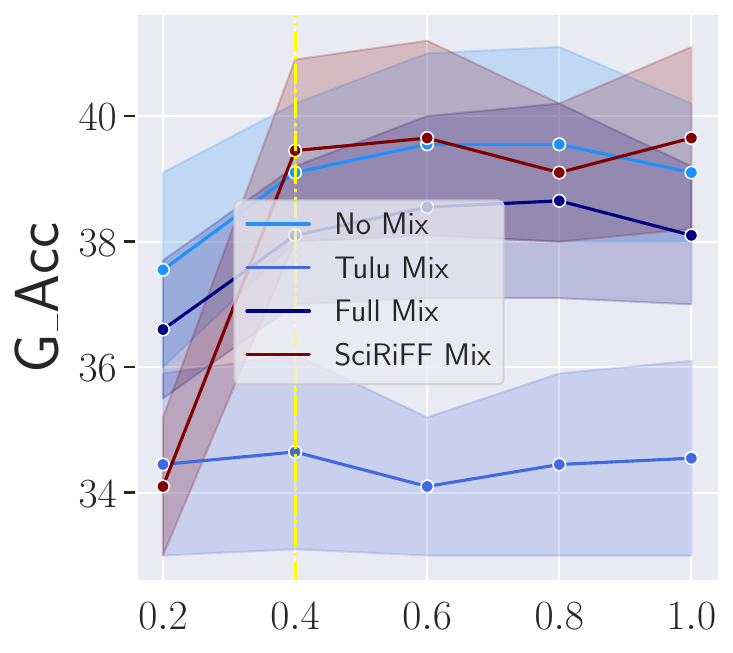}
        \caption{LLaMa 13B}
        \label{fig:llama13}
        \end{subfigure}%
        \hspace{1em}%
         \begin{subfigure}[b]{0.23\textwidth}
         \includegraphics[width=\textwidth]{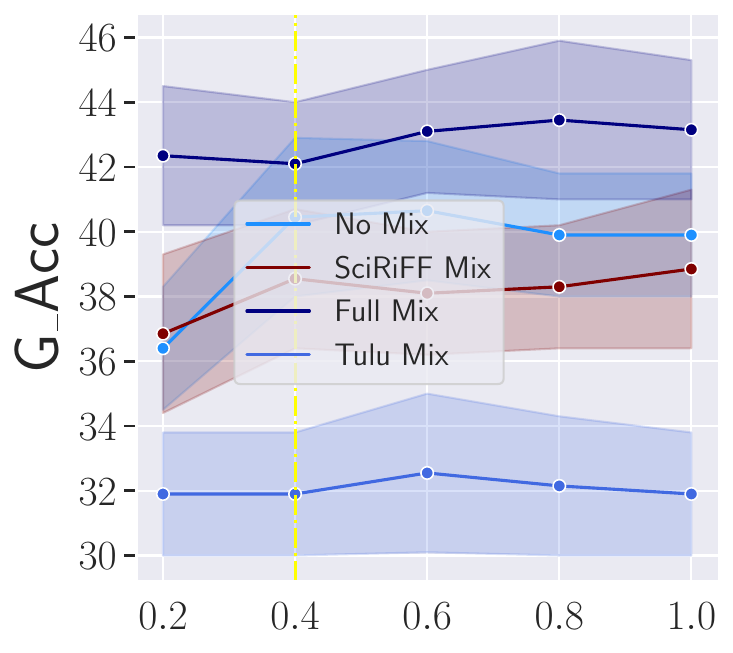}
        \caption{Mistral 7B}
        \label{fig:mistral}
        \end{subfigure}
        \hspace{1em}%

         \begin{subfigure}[b]{0.24\textwidth}
         \includegraphics[width=\textwidth]{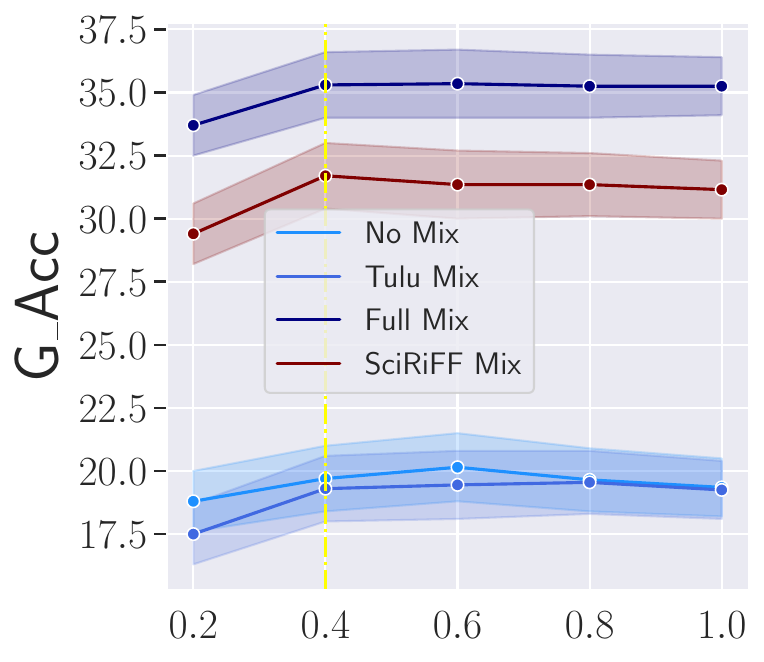}
        \caption{SciT\"ulu 7B}
        \label{fig:scitulu}
        \end{subfigure}
        \begin{subfigure}[b]{0.24\textwidth}
         \includegraphics[width=\textwidth]{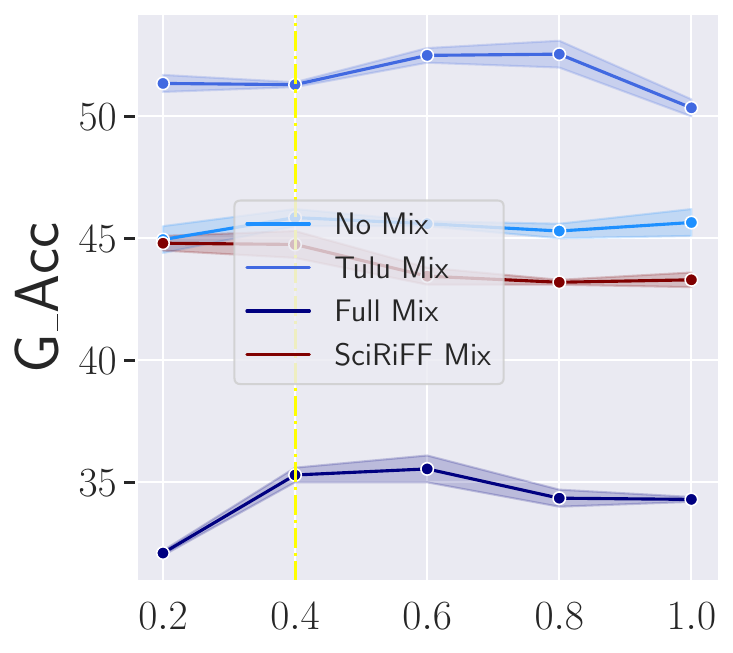}
        \caption{Qwen 7B}
        \label{fig:qwen}
        \end{subfigure}
         \begin{subfigure}[b]{0.24\textwidth}
         \includegraphics[width=\textwidth]{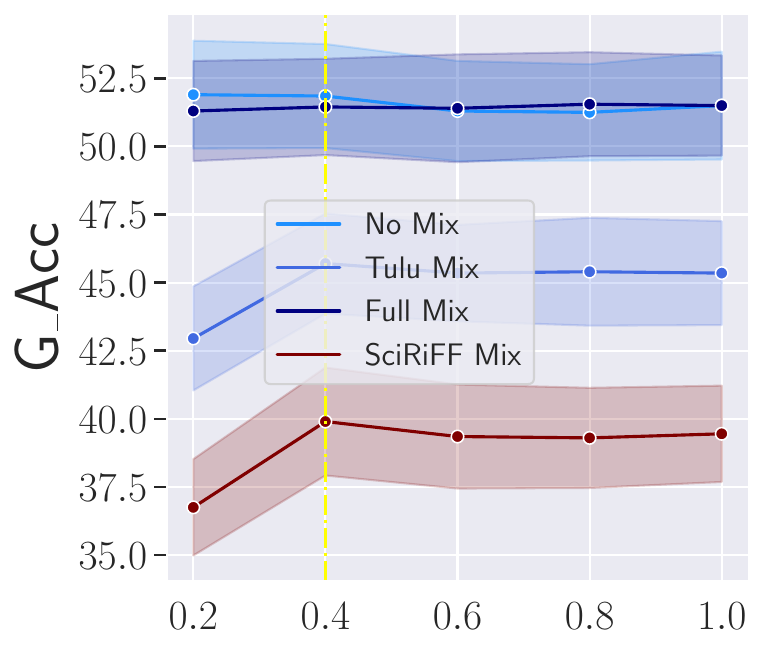}
        \caption{Yi 6B}
        \label{fig:yi}
        \end{subfigure}

    \caption{Performance of instruction-tuned LLMs for \hl{fine-grained classification} on the dev set with multiple percentages of dataset mixes using \textbf{target segment (T)} as the source of information in the prompt.}
    \label{fig:percent_data}
\end{figure*}

\begin{figure*}[!htb]
\centering
    \begin{subfigure}[b]{0.23\textwidth}
        \includegraphics[width=\textwidth]{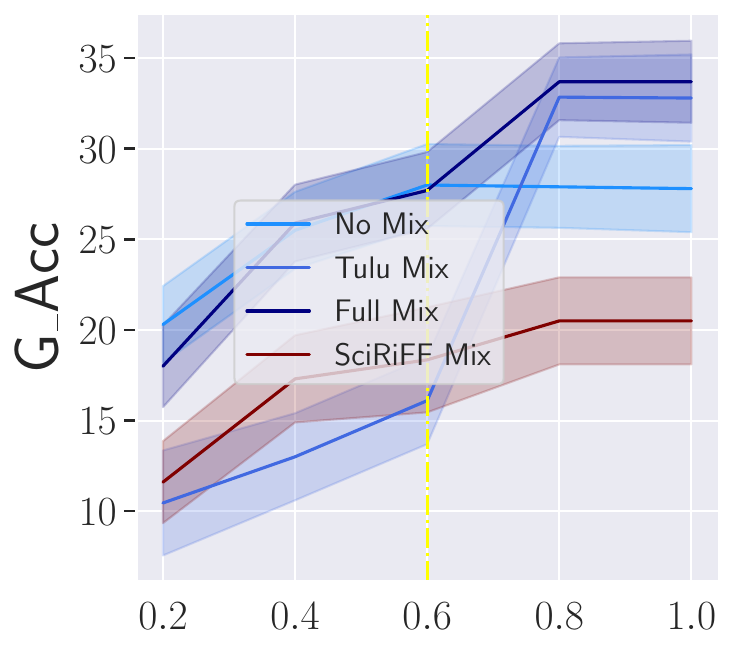}
        \caption{Gemma 7B}
        \label{fig:gemma}
        \end{subfigure}%
        \hspace{1em}%
        \begin{subfigure}[b]{0.23\textwidth}
        \includegraphics[width=\textwidth]{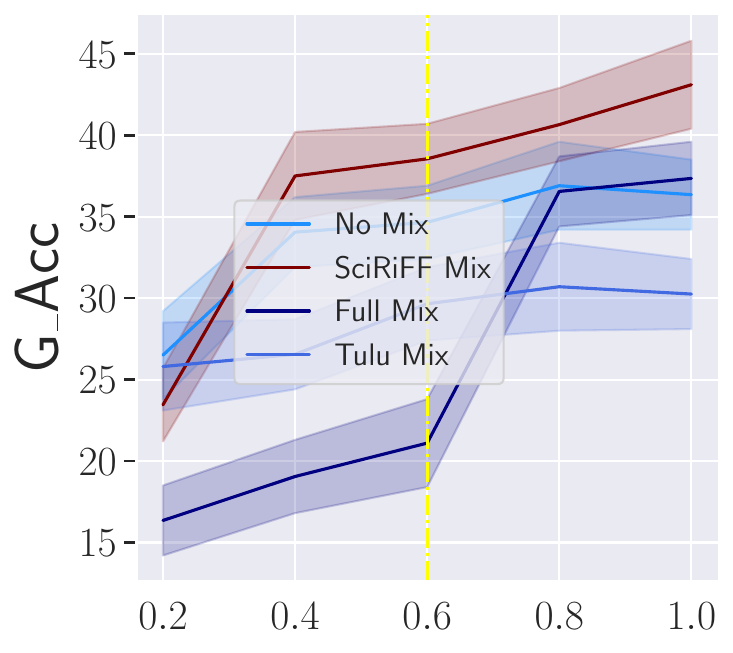}
        \caption{LLaMa 7B}
        \label{fig:llama}
        \end{subfigure}%
         \hspace{1em}%
        \begin{subfigure}[b]{0.23\textwidth}
         \includegraphics[width=\textwidth]{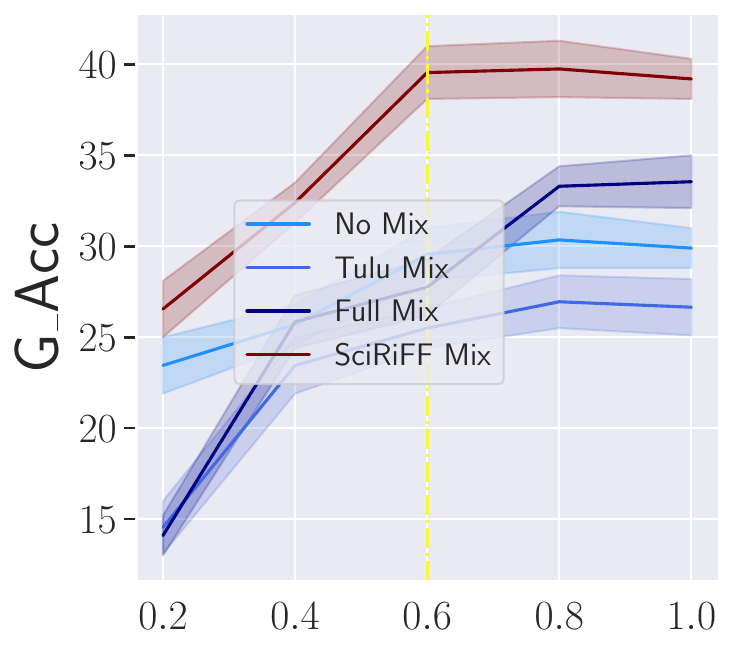}
        \caption{LLaMa 13B}
        \label{fig:llama13}
        \end{subfigure}%
        \hspace{1em}%
         \begin{subfigure}[b]{0.23\textwidth}
         \includegraphics[width=\textwidth]{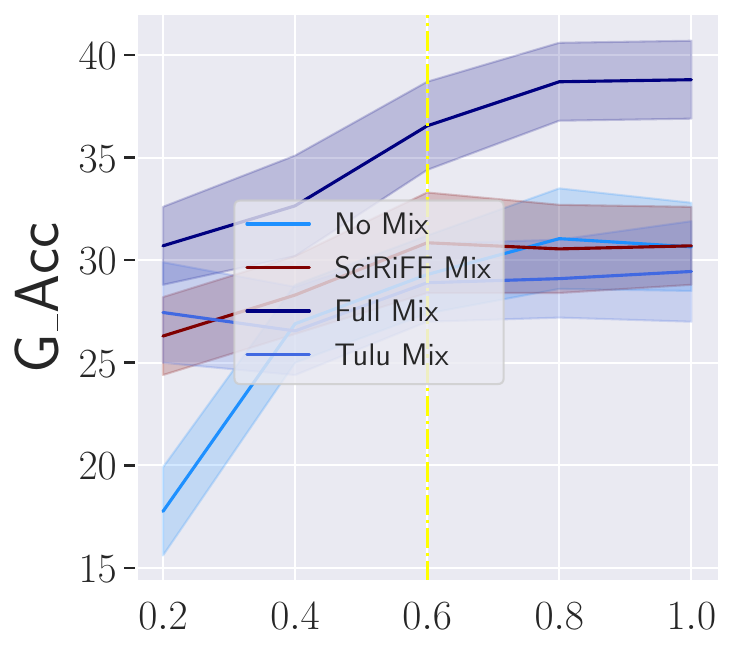}
        \caption{Mistral 7B}
        \label{fig:mistral}
        \end{subfigure}
        \hspace{1em}%

         \begin{subfigure}[b]{0.24\textwidth}
         \includegraphics[width=\textwidth]{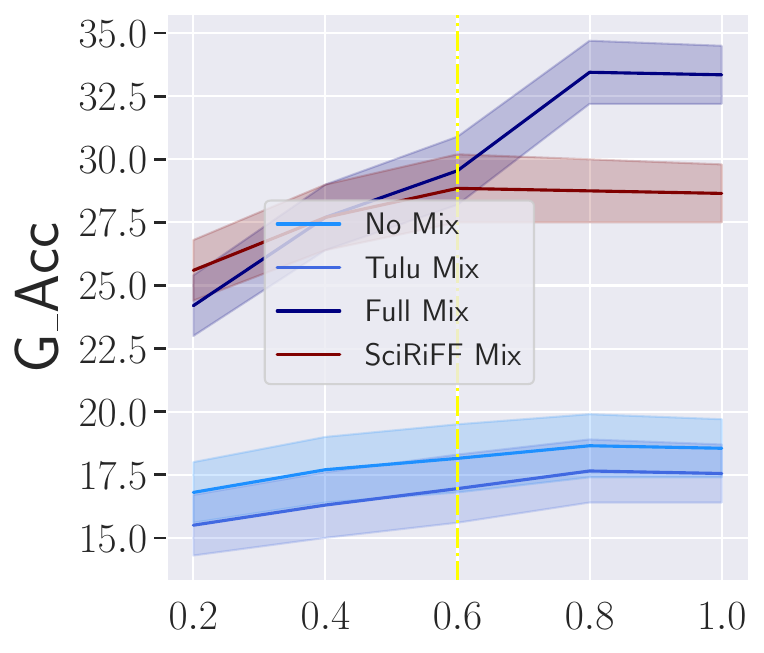}
        \caption{SciT\"ulu 7B}
        \label{fig:scitulu}
        \end{subfigure}
        \begin{subfigure}[b]{0.24\textwidth}
         \includegraphics[width=\textwidth]{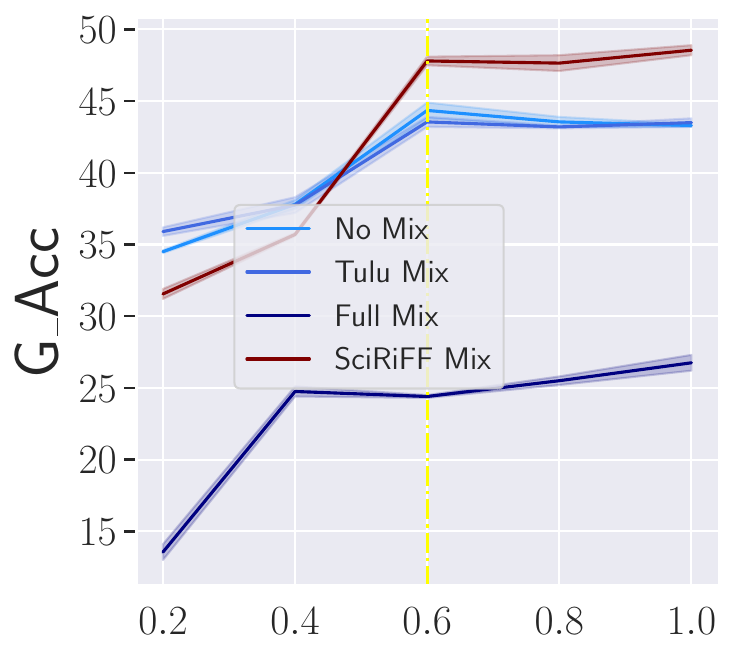}
        \caption{Qwen 7B}
        \label{fig:qwen}
        \end{subfigure}
         \begin{subfigure}[b]{0.24\textwidth}
         \includegraphics[width=\textwidth]{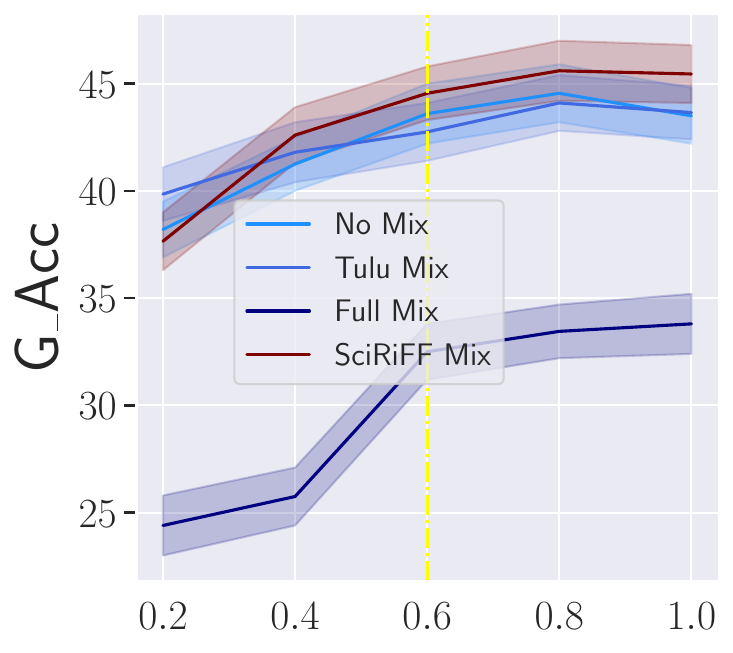}
        \caption{Yi 6B}
        \label{fig:yi}
        \end{subfigure}

    \caption{Performance of instruction-tuned LLMs for \hl{fine-grained classification} on the dev set with multiple percentages of dataset mixes using the  \textbf{combination of review and target segment (RT)} as the source of information in the prompt.}
    \label{fig:percent_data_rt}
\end{figure*}

\begin{figure*}[!htb]
\centering
    \begin{subfigure}[b]{0.23\textwidth}
        \includegraphics[width=\textwidth]{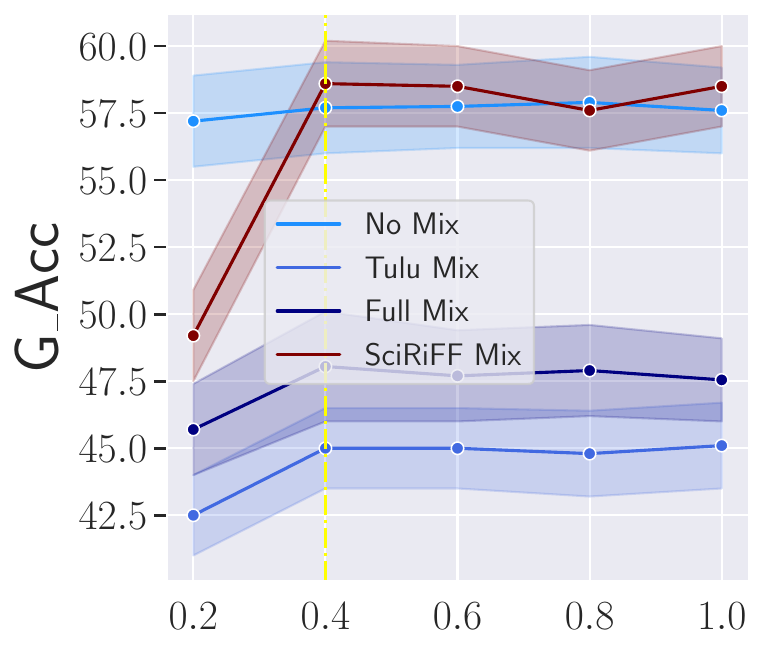}
        \caption{Gemma 7B}
        \label{fig:gemma}
        \end{subfigure}%
        \hspace{1em}%
        \begin{subfigure}[b]{0.23\textwidth}
        \includegraphics[width=\textwidth]{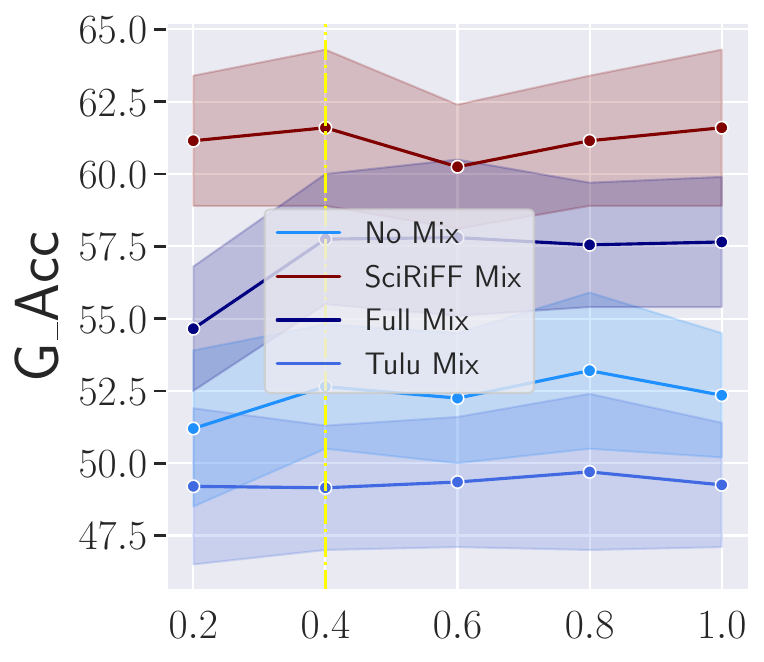}
        \caption{LLaMa 7B}
        \label{fig:llama}
        \end{subfigure}%
         \hspace{1em}%
        \begin{subfigure}[b]{0.23\textwidth}
         \includegraphics[width=\textwidth]{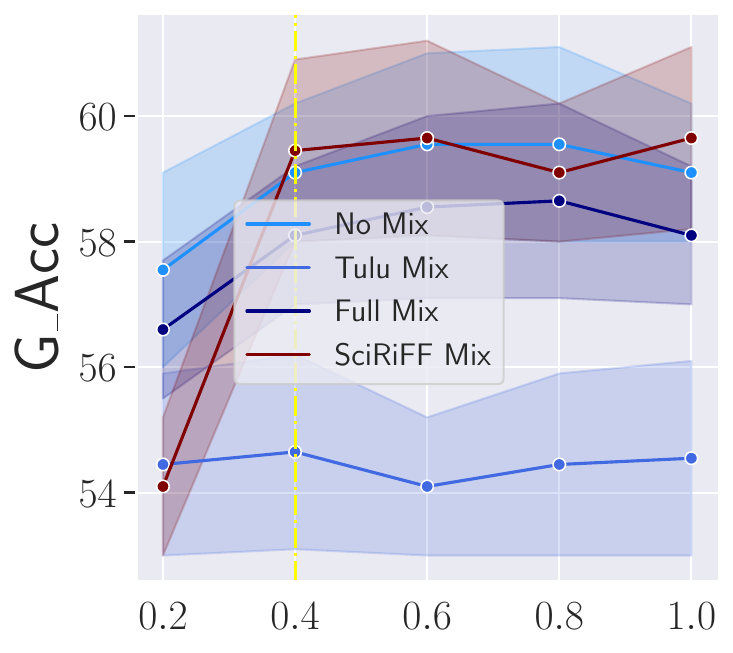}
        \caption{LLaMa 13B}
        \label{fig:llama13}
        \end{subfigure}%
        \hspace{1em}%
         \begin{subfigure}[b]{0.23\textwidth}
         \includegraphics[width=\textwidth]{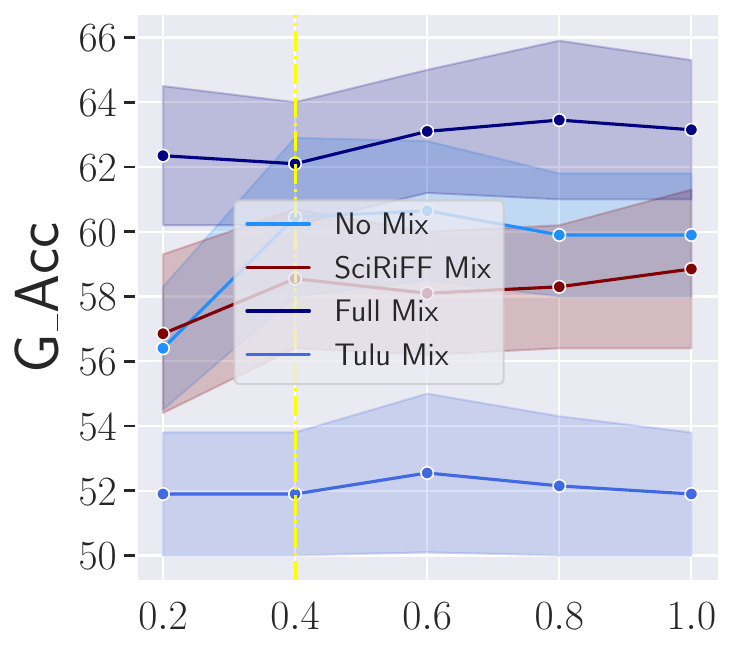}
        \caption{Mistral 7B}
        \label{fig:mistral}
        \end{subfigure}
        \hspace{1em}%

         \begin{subfigure}[b]{0.24\textwidth}
         \includegraphics[width=\textwidth]{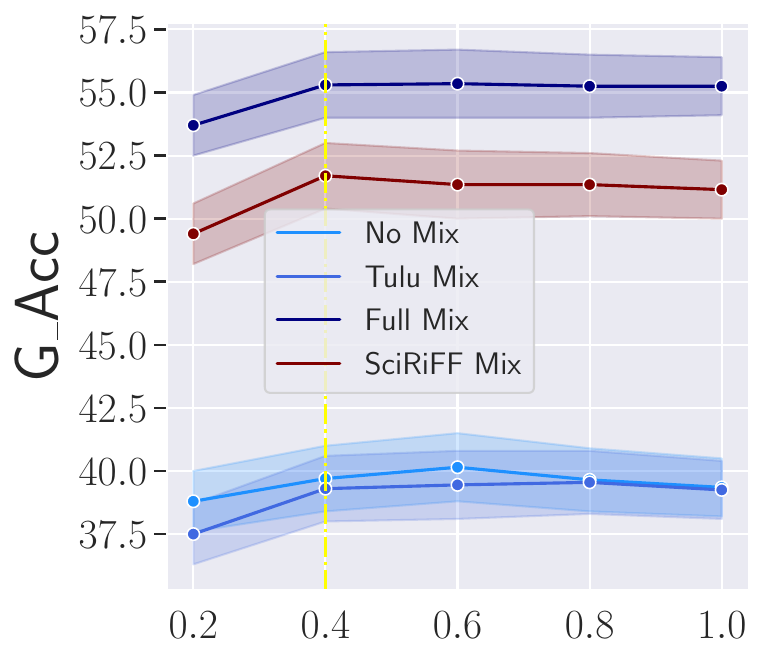}
        \caption{SciT\"ulu 7B}
        \label{fig:scitulu}
        \end{subfigure}
        \begin{subfigure}[b]{0.24\textwidth}
         \includegraphics[width=\textwidth]{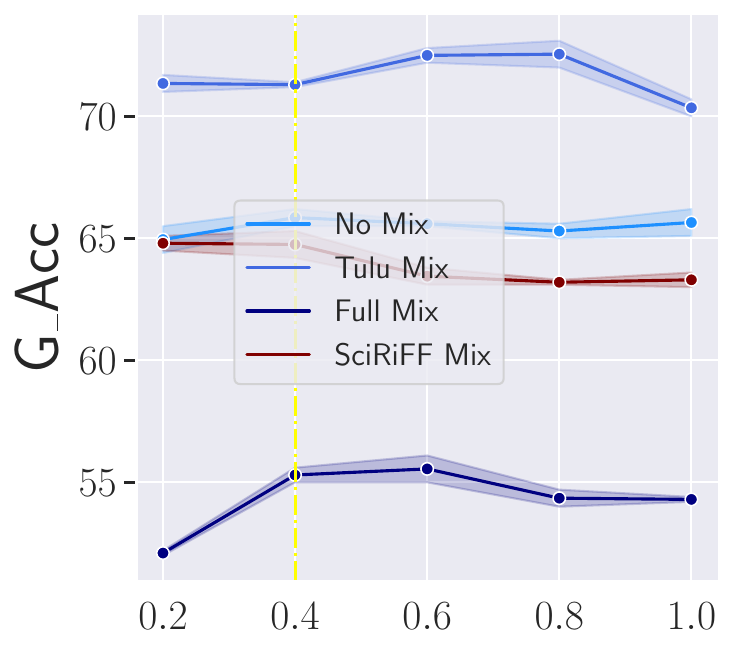}
        \caption{Qwen 7B}
        \label{fig:qwen}
        \end{subfigure}
         \begin{subfigure}[b]{0.24\textwidth}
         \includegraphics[width=\textwidth]{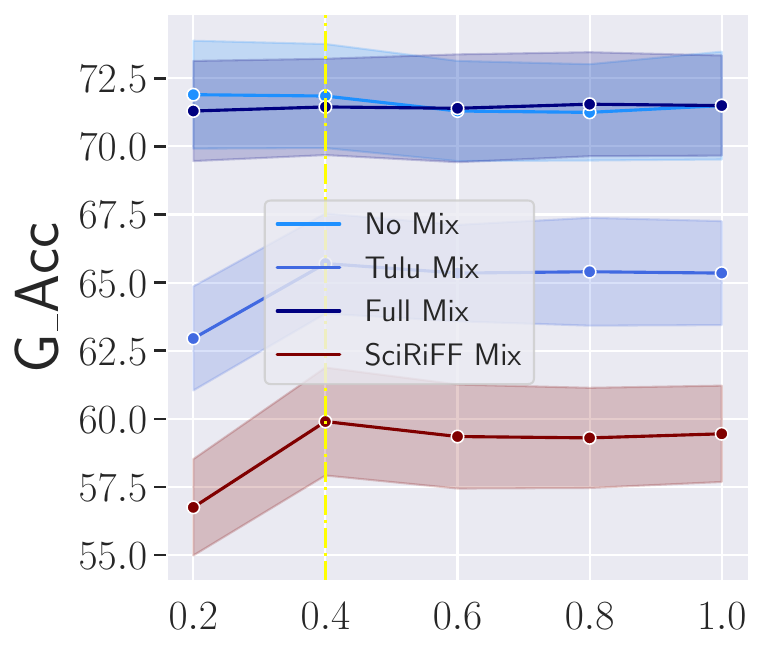}
        \caption{Yi 6B}
        \label{fig:yi}
        \end{subfigure}

    \caption{Performance of instruction-tuned LLMs for \hl{coarse-grained classification} on the dev set with multiple percentages of dataset mixes using \textbf{target segment (T)} as the source of information in the prompt.}
    \label{fig:percent_data_coarse}
\end{figure*}

\begin{figure*}[!htb]
\centering
    \begin{subfigure}[b]{0.23\textwidth}
        \includegraphics[width=\textwidth]{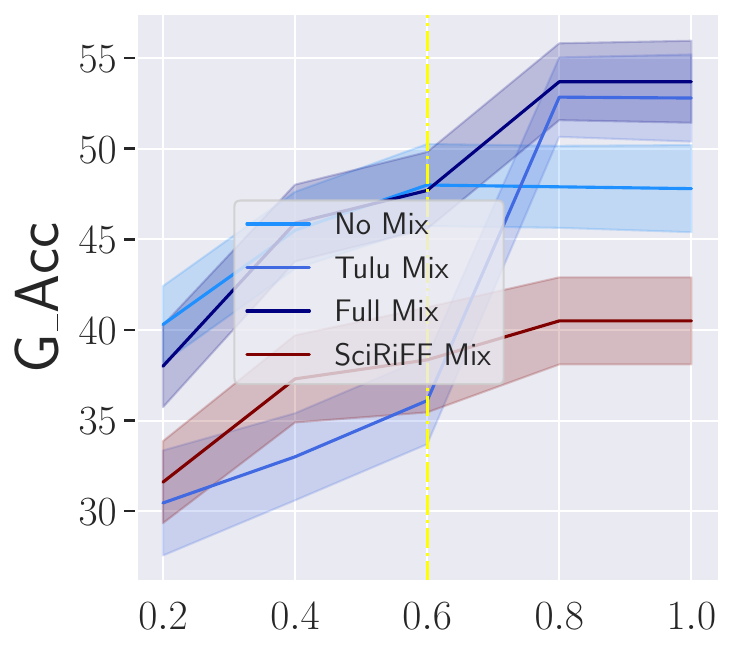}
        \caption{Gemma 7B}
        \label{fig:gemma}
        \end{subfigure}%
        \hspace{1em}%
        \begin{subfigure}[b]{0.23\textwidth}
        \includegraphics[width=\textwidth]{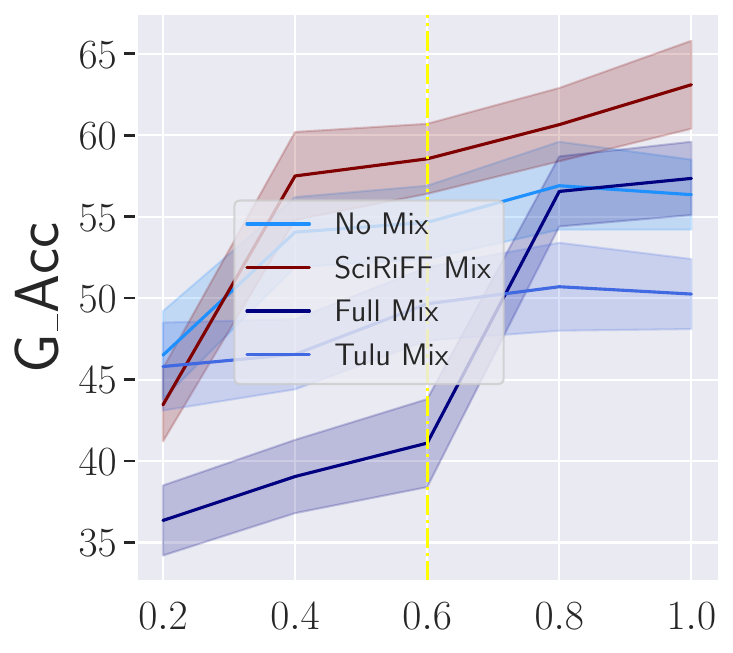}
        \caption{LLaMa 7B}
        \label{fig:llama}
        \end{subfigure}%
         \hspace{1em}%
        \begin{subfigure}[b]{0.23\textwidth}
         \includegraphics[width=\textwidth]{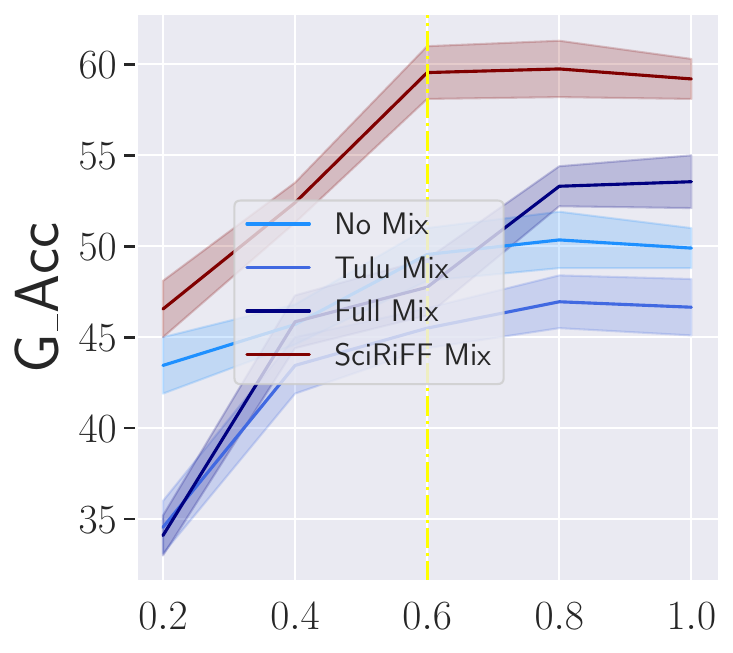}
        \caption{LLaMa 13B}
        \label{fig:llama13}
        \end{subfigure}%
        \hspace{1em}%
         \begin{subfigure}[b]{0.23\textwidth}
         \includegraphics[width=\textwidth]{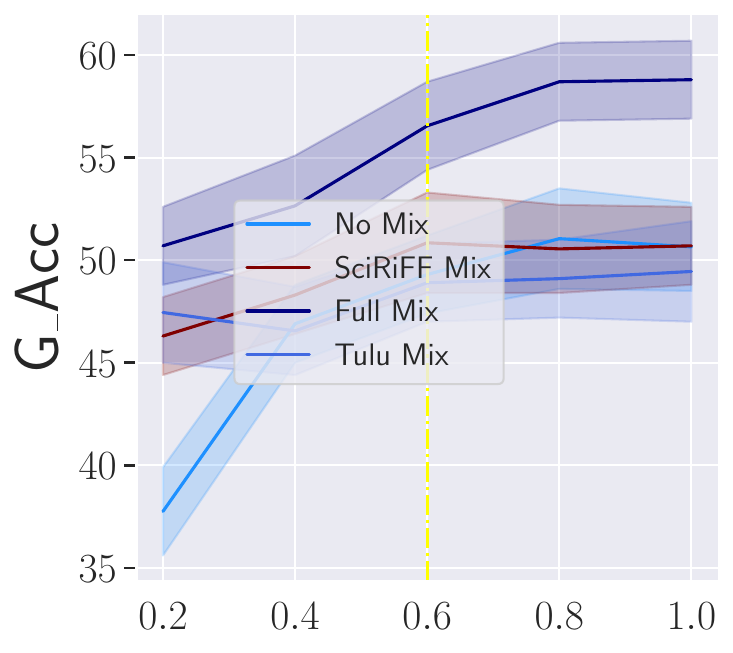}
        \caption{Mistral 7B}
        \label{fig:mistral}
        \end{subfigure}
        \hspace{1em}%

         \begin{subfigure}[b]{0.24\textwidth}
         \includegraphics[width=\textwidth]{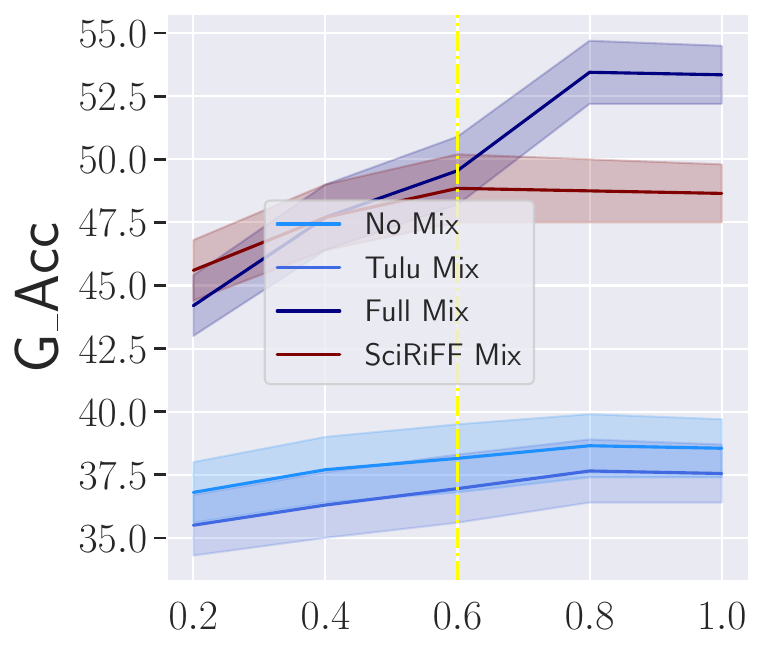}
        \caption{SciT\"ulu 7B}
        \label{fig:scitulu}
        \end{subfigure}
        \begin{subfigure}[b]{0.24\textwidth}
         \includegraphics[width=\textwidth]{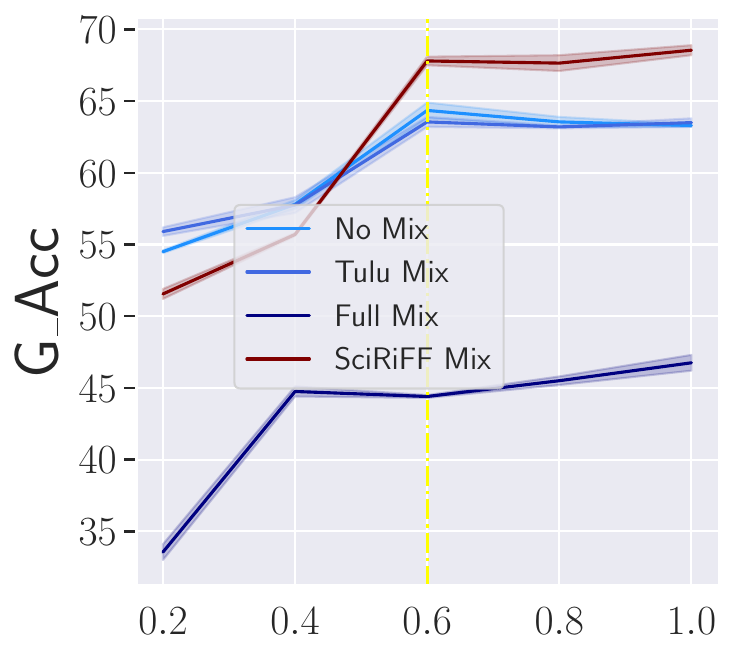}
        \caption{Qwen 7B}
        \label{fig:qwen}
        \end{subfigure}
         \begin{subfigure}[b]{0.24\textwidth}
         \includegraphics[width=\textwidth]{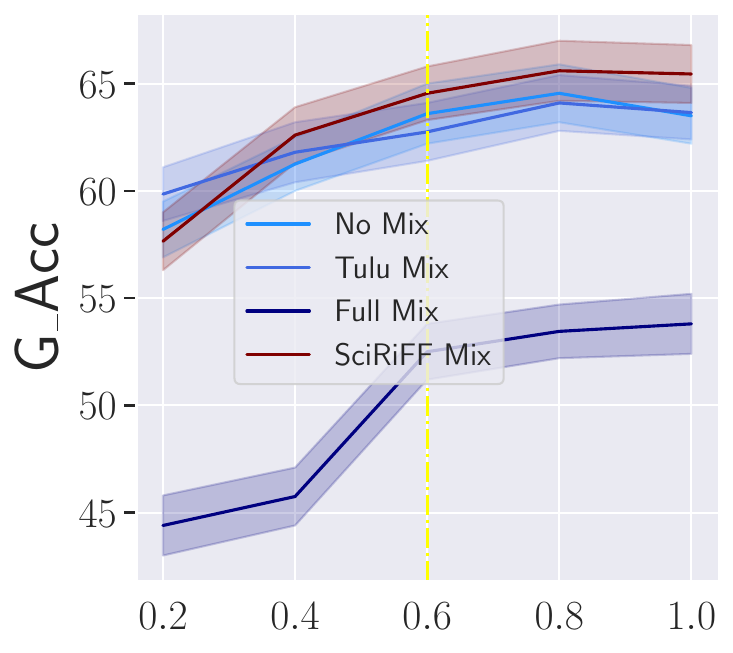}
        \caption{Yi 6B}
        \label{fig:yi}
        \end{subfigure}

    \caption{Performance of instruction-tuned LLMs for \hl{coarse-grained classification} on the dev set with multiple percentages of dataset mixes using the  \textbf{combination of review and target segment (RT)} as the source of information in the prompt.}
    \label{fig:percent_data_rt_coarse}
\end{figure*}

\begin{table*}[]
\centering
\resizebox{!}{0.5\textwidth}{\begin{tabular}{lllllllllll}
\hline
\multicolumn{1}{c}{\multirow{4}{*}{\textbf{Method}}} & \multicolumn{1}{c}{\multirow{4}{*}{\textbf{Models}}} & \multicolumn{9}{c}{\textbf{Fine-grained}}                                                                                                                                                                                                                                                \\ \cline{3-11} 
\multicolumn{1}{c}{}                                 & \multicolumn{1}{c}{}                                 & \multicolumn{6}{c|}{\textbf{S.A}}                                                                                                                              & \multicolumn{3}{c}{\textbf{G.A}}                                                                                    \\ \cmidrule(lr){3-8} \cmidrule(lr){9-11} 
\multicolumn{1}{c}{}                                 & \multicolumn{1}{c}{}                                 & \multicolumn{3}{c}{\textbf{Acc}}                                                  & \multicolumn{3}{c|}{\textbf{F1}}                                             & \multicolumn{3}{c}{\textbf{Acc}}                                                                                      \\ \cmidrule(lr){3-8} \cmidrule(lr){9-11} 
\multicolumn{1}{c}{}                                 & \multicolumn{1}{c}{}                                 & \textbf{R1}   & \textbf{R2 ($\Delta_{R1}$)}  & \multicolumn{1}{c|}{\textbf{R3}}   & \textbf{R1} & \textbf{R2 ($\Delta_{R1}$)} & \multicolumn{1}{l|}{\textbf{R3}} & \multicolumn{1}{l}{\textbf{R1}} & \multicolumn{1}{l}{\textbf{R2 ($\Delta_{R1}$)}} & \multicolumn{1}{c}{\textbf{R3}} \\ \hline
\multirow{3}{*}{}                                    \textsc{Random} &  -                                   & 7.11          & 4.34 \reddown{2.77}          & \multicolumn{1}{l|}{2.46 \reddown{1.88}}          & 7.15        & 2.39 \reddown{4.76}                       & \multicolumn{1}{l|}{1.28}        & -                                & -                                                & -                               \\
                                                     \textsc{Majority} &  -                                  & 11.1          & 7.34 \greenup{3.00}          & \multicolumn{1}{l|}{5.11 \reddown{2.23}}          & 9.05        & 7.17 \reddown{1.88}                       & \multicolumn{1}{l|}{3.45}        & -                                & -                                                & -                               \\

                                                     \hline
\multirow{28}{*}{Zero-Shot}                          & Gemma-1.1 7B (T)                                     & 22.2          & 26.7 \greenup{\textbf{4.50}} & \multicolumn{1}{l|}{18.9}          & 23.1        & 24.8 \greenup{1.7}                       & \multicolumn{1}{l|}{16.7}        & 52.2                             & 58.1 \greenup{5.90}                     & 32.2                            \\
                                                     & Gemma-1.1 7B (TE)                                    & -             & -                            & \multicolumn{1}{l|}{24.4} \greenup{5.5}         & -           & -                           & \multicolumn{1}{l|}{19.4} \greenup{2.7}       & -                                & -                                                & 41.1 \greenup{8.9}                           \\ \cmidrule(lr){3-5} \cmidrule(lr){6-8} \cmidrule(lr){9-11}
                                                     & Gemma-1.1 7B (RT)                                    & 12.2          & 11.6 \reddown{0.60}          & \multicolumn{1}{l|}{14.4}          & 9.51        & 10.4 \greenup{0.89}                       & \multicolumn{1}{l|}{10.8}        & 46.7                             & 51.1 \greenup{4.40}                              & 32.2                            \\
                                                     & Gemma-1.1 7B (RTE)                                   & -             & -                            & \multicolumn{1}{l|}{17.3} \greenup{2.9}          & -           & -                           & \multicolumn{1}{l|}{12.8} \greenup{2.0}       & -                                & -                                                & 32.8  \greenup{0.6}                          \\ \cline{2-11} 
                                                     & LLaMa 7B (T)                                         & 12.2          & 22.2 \greenup{10.0}          & \multicolumn{1}{l|}{11.1}          & 9.51        & 20.3 \greenup{10.79}                       & \multicolumn{1}{l|}{9.12}        & 15.6                             & 30.6 \greenup{15.0}                              & 35.6                            \\
                                                     & LLaMa 7B (TE)                                        & -             & -                            & \multicolumn{1}{l|}{15.6} \greenup{4.5}          & -           & -                           & \multicolumn{1}{l|}{12.4}\greenup{3.28}        & -                                & -                                                & 38.9    \greenup{3.3}                        \\ \cmidrule(lr){3-5} \cmidrule(lr){6-8} \cmidrule(lr){9-11}
                                                     & LLaMa 7B (RT)                                        & 12.2          & 13.2 \greenup{1.00}          & \multicolumn{1}{l|}{12.2}          & 9.51        & 9.89 \greenup{0.38}                       & \multicolumn{1}{l|}{10.1}        & 25.6                             & 33.7 \greenup{7.90}                              & 28.9                            \\
                                                     & LLaMa 7B (RTE)                                       & -             & -                            & \multicolumn{1}{l|}{14.2} \greenup{2.0}          & -           & -                           & \multicolumn{1}{l|}{11.5} \greenup{1.4}        & -                                & -                                                & 30.8    \greenup{1.9}                        \\ \cline{2-11} 
                                                     & LLaMa 13B (T)                                        & 26.7          & 26.7 \greenup{0.00}          & \multicolumn{1}{l|}{11.1}          & 20.4        & 23.8 \greenup{3.4}                        & \multicolumn{1}{l|}{9.11}        & 44.4                             & 45.3 \greenup{0.90}                              & 35.6                            \\
                                                     & LLaMa 13B (TE)                                       & -             & -                            & \multicolumn{1}{l|}{24.4} \greenup{13.3}          & -           & -                           & \multicolumn{1}{l|}{20.8} \greenup{11.7}       & -                                & -                                                & 41.1 \greenup{5.5}                           \\ \cmidrule(lr){3-5} \cmidrule(lr){6-8} \cmidrule(lr){9-11}
                                                     & LLaMa 13B (RT)                                       & 15.6          & 17.6 \greenup{3.00}          & \multicolumn{1}{l|}{10.7}          & 11.7        & 13.8 \greenup{2.1}                        & \multicolumn{1}{l|}{8.89}        & 41.1                             & 40.4 \reddown{1.10}                              & 32.2                            \\
                                                     & LLaMa 13B (RTE)                                      & -             & -                            & \multicolumn{1}{l|}{18.8} \greenup{8.1}          & -           & -                           & \multicolumn{1}{l|}{17.5} \greenup{8.61}        & -                                & -                                                & 34.4             \greenup{2.2}               \\ \cline{2-11} 
                                                     & Mistral 7B (T)                                       & 27.8          & 28.8 \greenup{1.10}          & \multicolumn{1}{l|}{28.2}          & 17.5        & 19.7 \greenup{2.2}                        & \multicolumn{1}{l|}{24.3}        & 47.8                             & 51.1 \greenup{3.30}                              & 54.4                            \\
                                                     & Mistral 7B (TE)                                      & -             & -                            & \multicolumn{1}{l|}{30.0} \greenup{1.8}         & -           & -                           & \multicolumn{1}{l|}{26.1} \greenup{1.8 }        & -                                & -                                                & 55.6 \greenup{1.2}                   \\ \cmidrule(lr){3-5} \cmidrule(lr){6-8} \cmidrule(lr){9-11}
                                                     & Mistral 7B (RT)                                      & 12.2          & 16.6 \greenup{4.40}          & \multicolumn{1}{l|}{22.2}          & 9.51        & 12.8 \greenup{3.29}                        & \multicolumn{1}{l|}{19.4}        & 28.9                             & 35.9 \greenup{6.90}                              & 51.1                            \\
                                                     & Mistral 7B (RTE)                                     & -             & -                            & \multicolumn{1}{l|}{27.8} \greenup{5.6}          & -           & -                           & \multicolumn{1}{l|}{23.2} \greenup{3.8}       & -                                & -                                                & 52.2   \greenup{1.1}                         \\ \cline{2-11} 
                                                     & Qwen 7B (T)                                          & 21.1          & 22.7 \greenup{1.60}          & \multicolumn{1}{l|}{28.9}          & 17.8        & 19.8 \greenup{2.0}                        & \multicolumn{1}{l|}{24.5}        & 46.7                             & 50.0 \greenup{3.30}                              & 44.4                            \\
                                                     & Qwen 7B (TE)                                         & -             & -                            & \multicolumn{1}{l|}{\textbf{31.1}} \greenup{2.2} & -           & -                           & \multicolumn{1}{l|}{\textbf{26.8}} \greenup{2.3}        & -                                & -                                                & \textbf{56.4}     \greenup{12.0}                       \\ \cmidrule(lr){3-5} \cmidrule(lr){6-8} \cmidrule(lr){9-11}
                                                     & Qwen 7B (RT)                                         & 12.2          & 13.3 \greenup{1.10}          & \multicolumn{1}{l|}{26.7}          & 9.51        & 13.4 \greenup{3.89}                        & \multicolumn{1}{l|}{22.8}        & 43.3                             & 42.6 \reddown{1.60}                              & 43.3                            \\
                                                     & Qwen 7B (RTE)                                        & -             & -                            & \multicolumn{1}{l|}{27.8} \greenup{1.1}          & -           & -                           & \multicolumn{1}{l|}{24.7} \greenup{1.9}        & -                                & -                                                & 44.2   \greenup{0.9}                         \\ \cline{2-11} 
                                                     & Yi-1.5 6B (T)                                        & \textbf{35.3}          & \textbf{37.6} \greenup{2.3} & \multicolumn{1}{l|}{26.7}          & \textbf{23.5}        & \textbf{29.7} \greenup{6.2}                       & \multicolumn{1}{l|}{22.1}        & \textbf{56.7}                    & \textbf{60.0} \greenup{3.30}                              & 53.8                            \\
                                                     & Yi-1.5 6B (TE)                                       & -             & -                            & \multicolumn{1}{l|}{30.0} \greenup{3.3}         & -           & -                           & \multicolumn{1}{l|}{24.8} \greenup{2.7}        & -                                & -                                                & 54.9   \greenup{1.1}                         \\ \cmidrule(lr){3-5} \cmidrule(lr){6-8} \cmidrule(lr){9-11}
                                                     & Yi-1.5 6B (RT)                                       & 34.4 & 32.8 \reddown{1.60}          & \multicolumn{1}{l|}{21.2}          & 22.7        & 27.8 \greenup{5.1}                       & \multicolumn{1}{l|}{19.4}        & 51.1                             & 52.2 \greenup{1.20}                              & 51.3                            \\
                                                     & Yi-1.5 6B (RTE)                                      & -             & -                            & \multicolumn{1}{l|}{24.4} \greenup{2.2}          & -           & -                           & \multicolumn{1}{l|}{20.6} \greenup{1.2}        & -                                & -                                                & 52.7   \greenup{0.6}                         \\ \cline{2-11} 
                                                     & SciT\"ulu 7B (T)                                     & 14.4          & 25.3 \greenup{10.9}          & \multicolumn{1}{l|}{22.2}          & 10.5        & 22.5 \greenup{12.0}                        & \multicolumn{1}{l|}{16.2}        & 18.1                             & 29.4 \greenup{\textbf{11.3}}                              & 42.2                            \\
                                                     & SciT\"ulu 7B (TE)                                    & -             & -                            & \multicolumn{1}{l|}{23.3} \greenup{1.1}          & -           & -                           & \multicolumn{1}{l|}{17.8} \greenup{1.6}        & -                                & -                                                & 44.8   \greenup{2.6}                         \\ \cmidrule(lr){3-5} \cmidrule(lr){6-8} \cmidrule(lr){9-11}
                                                     & SciT\"ulu 7B (RT)                                    & 15.6          & 18.3 \greenup{2.70}          & \multicolumn{1}{l|}{18.9}          & 12.8        & 16.7 \greenup{3.9}                        & \multicolumn{1}{l|}{14.9}        & 17.3                             & 23.7 \greenup{6.4}                     & 38.9                            \\
                                                     & SciT\"ulu 7B (RTE)                                   & -             & -                            & \multicolumn{1}{l|}{19.7} \greenup{0.8}          & -           & -                           & \multicolumn{1}{l|}{15.5} \greenup{0.6}        & -                                & -                                                & 41.1  \greenup{2.2}                          \\ \hline
\multirow{14}{*}{Instruction Tuned}                  & Gemma-1.1 7B (T)                                     & 31.4          & 38.8 \greenup{7.40}          & \multicolumn{1}{l|}{34.6}          & 29.8        & 31.6 \greenup{1.8}                        & \multicolumn{1}{l|}{28.1}        & 49.6                             & 53.4 \greenup{3.80}                              & 46.9                            \\
                                                     & Gemma-1.1 7B (RT)                                    & 28.2          & 35.7 \greenup{7.50}          & \multicolumn{1}{l|}{32.8}          & 26.8        & 33.7 \greenup{6.9}                        & \multicolumn{1}{l|}{24.3}        & 44.7                             & 51.2 \greenup{6.50}                              & 42.8                            \\ \cline{2-11} 
                                                     & LLaMa 7B (T)                                         & 43.8          & 47.8 \greenup{2.00}          & \multicolumn{1}{l|}{44.7}          & 36.8        & 37.2 \greenup{0.4}                        & \multicolumn{1}{l|}{37.2}        & 60.8                             & 65.4 \greenup{4.60}                              & 48.9                            \\
                                                     & LLaMa 7B (RT)                                        & 43.2          & 45.3 \greenup{2.10}          & \multicolumn{1}{l|}{41.8}          & 32.1        & 33.8 \greenup{0.7}                        & \multicolumn{1}{l|}{32.3}        & 58.4                             & 61.3 \greenup{2.90}                              & 45.4                            \\ \cline{2-11} 
                                                     & LLaMa 13B (T)                                        & 45.8          & 47.8 \greenup{2.00}          & \multicolumn{1}{l|}{50.5}          & 37.5        & 38.2 \greenup{0.7}                        & \multicolumn{1}{l|}{40.1}        & 50.3                             & 51.2 \greenup{0.90}                              & 54.2                            \\
                                                     & LLaMa 13B (RT)                                       & 41.2          & 45.2 \greenup{4.00}          & \multicolumn{1}{l|}{47.3}          & 32.4        & 36.7 \greenup{3.3}                        & \multicolumn{1}{l|}{42.6}        & 48.2                             & 48.8 \greenup{0.60}                              & 52.2                            \\ \cline{2-11} 
                                                     & Mistral 7B (T)                                       & 35.4          & 37.4 \greenup{2.00}          & \multicolumn{1}{l|}{42.4}          & 28.8        & 30.4 \greenup{1.6}                        & \multicolumn{1}{l|}{36.4}        & 49.2                             & 54.3 \greenup{5.10}                              & 58.3                            \\
                                                     & Mistral 7B (RT)                                      & 31.2          & 35.2 \greenup{4.00}          & \multicolumn{1}{l|}{37.8}          & 24.5        & 27.4 \greenup{2.9}                        & \multicolumn{1}{l|}{32.5}        & 44.2                             & 50.4 \greenup{6.20}                              & 56.3                            \\ \cline{2-11} 
                                                     & Qwen 7B (T)                                          & \textbf{45.9} & \textbf{48.4} \greenup{2.50} & \multicolumn{1}{l|}{59.4}          & 38.6        & 40.2 \greenup{1.6}                        & \multicolumn{1}{l|}{57.2}        & 51.2                             & 55.8 \greenup{4.60}                              & 62.3                            \\
                                                     & Qwen 7B (RT)                                         & 41.2          & 42.4 \greenup{1.20}          & \multicolumn{1}{l|}{47.8}          & 34.3        & 36.4 \greenup{2.1}                        & \multicolumn{1}{l|}{35.3}        & 49.3                             & 50.4 \greenup{1.10}                              & 47.3                            \\ \cline{2-11} 
                                                     & Yi-1.5 6B (T)                                        & 45.1          & 47.8 \greenup{2.70}          & \multicolumn{1}{l|}{47.9}          & 36.4        & 38.2 \greenup{1.8}                        & \multicolumn{1}{l|}{38.2}        & 51.2                    & 53.8 \greenup{2.60}                     & 52.2                   \\
                                                     & Yi-1.5 6B (RT)                                       & 43.2          & 45.3 \greenup{2.10}          & \multicolumn{1}{l|}{46.3}          & 31.7        & 33.5 \greenup{1.8}                        & \multicolumn{1}{l|}{34.6}        & 59.4                             & 61.2 \greenup{1.80}                              & 58.4                            \\ \cline{2-11} 
                                                     & SciT\"ulu 7B (T)                                     & 45.7          & 48.6 \greenup{2.90}          & \multicolumn{1}{l|}{54.3} & 38.6        & 40.2 \greenup{1.6}                       & \multicolumn{1}{l|}{48.5}        & 51.3                             & 52.6 \greenup{1.30}                              & 54.8                            \\
                                                     & SciT\"ulu 7B (RT)                                    & 41.4          & 42.6 \greenup{1.20}          & \multicolumn{1}{l|}{51.4}          & 34.3        & 35.6 \greenup{1.3}                       & \multicolumn{1}{l|}{42.3}        & 49.4                             & 50.2 \greenup{0.80}                              & 51.3                            \\ \hline
\end{tabular}}
\caption{Performance of LLMs on different rounds of annotation. S.A represents the string-matching evaluator, and G.A represents the GPT-based evaluator. `T' represents prompting with only the target sentence, RT represents the combination of the review and the target sentence. Adding demonstrations to the prompt is represented by `E'. R1 represents `Round 1', R2 represents `Round 2', and R3 represents `Round 3' respectively. }
\label{tab:ann_round_llm_fine-huge}
\end{table*}

\begin{table*}[]
\centering
\resizebox{!}{0.5\textwidth}{\begin{tabular}{llllllll|lll}
\hline
\multicolumn{1}{c}{\multirow{4}{*}{\textbf{Method}}} & \multicolumn{1}{c}{\multirow{4}{*}{\textbf{Models}}} & \multicolumn{9}{c}{\textbf{Coarse-grained}}                                                                                                                                                                                                                                                \\ \cline{3-11} 
\multicolumn{1}{c}{}                                 & \multicolumn{1}{c}{}                                 & \multicolumn{6}{c|}{\textbf{S.A}}                                                                                                                              & \multicolumn{3}{c}{\textbf{G.A}}                                                                                    \\ \cmidrule(lr){3-8} \cmidrule(lr){9-11}
\multicolumn{1}{c}{}                                 & \multicolumn{1}{c}{}                                 & \multicolumn{3}{c}{\textbf{Acc}}                                                  & \multicolumn{3}{c|}{\textbf{F1}}                                             & \multicolumn{3}{c}{\textbf{Acc}}                                                                                      \\ \cmidrule(lr){3-8} \cmidrule(lr){9-11}
\multicolumn{1}{c}{}                                 & \multicolumn{1}{c}{}                                 & \textbf{R1}   & \textbf{R2 ($\Delta_{R1}$)}  & \multicolumn{1}{c|}{\textbf{R3}}   & \textbf{R1} & \textbf{R2 ($\Delta_{R1}$)} & \multicolumn{1}{l|}{\textbf{R3}} & \multicolumn{1}{l}{\textbf{R1}} & \multicolumn{1}{l}{\textbf{R2 ($\Delta_{R1}$)}} & \multicolumn{1}{c}{\textbf{R3}} \\ \hline
                                   \textsc{Random}                  &   -                                       & 40.7          & 40.7 \greenup{0.00}          & \multicolumn{1}{l|}{43.3}          & 35.6        & 35.6 \greenup{0.0}                        & \multicolumn{1}{l|}{38.2}        & -             & -                            & -                               \\
                                       \textsc{Majority}              &   -                                     & 51.4          & 51.4 \greenup{0.00}          & \multicolumn{1}{l|}{52.3}          & 44.4        & 48.2 \greenup{3.8}                        & \multicolumn{1}{l|}{49.3}        & -             & -                            & -                               \\
                                        \cline{1-11} 
\multirow{28}{*}{Zero-Shot}                          & Gemma-1.1 7B (T)                                      & 44.3          & 46.1 \greenup{1.8}           & \multicolumn{1}{l|}{55.6}          & 38.7        & 40.2 \greenup{1.5}                       & \multicolumn{1}{l|}{45.3}        & 51.1          & 54.4 \greenup{3.30}          & 57.8                            \\
                                                     & Gemma-1.1 7B (TE)                                     & -             & -                            & \multicolumn{1}{l|}{75.6} \greenup{20.0}          & -           & -                           & \multicolumn{1}{l|}{67.8} \greenup{12.5}        & -             & -                            & 88.9 \greenup{31.1}                            \\ \cmidrule(lr){3-5} \cmidrule(lr){6-8} \cmidrule(lr){9-11}
                                                     & Gemma-1.1 7B (RT)                                     & 48.1          & 50.4 \greenup{2.30}          & \multicolumn{1}{l|}{65.6}          & 42.2        & 45.3 \greenup{3.1}                        & \multicolumn{1}{l|}{54.2}        & 47.4          & 49.1 \greenup{1.70}          & 65.6                            \\
                                                     & Gemma-1.1 7B (RTE)                                    & -             & -                            & \multicolumn{1}{l|}{71.1} \greenup{5.5}          & -           & -                           & \multicolumn{1}{l|}{65.4} \greenup{11.2}        & -             & -                            & 82.2            \greenup{16.6}                \\ \cline{2-11}
                                                     & LLaMa 7B (T)                                          & 57.7          & 60.0 \greenup{2.30}          & \multicolumn{1}{l|}{80.0}          & 52.4       & 55.3 \greenup{0.9}                       & \multicolumn{1}{l|}{67.2}        & 70.0          & 75.0 \greenup{5.00} & 86.1                            \\
                                                     & LLaMa 7B (TE)                                         & -             & -                            & \multicolumn{1}{l|}{84.4} \greenup{4.4}          & -           & -                           & \multicolumn{1}{l|}{70.3} \greenup{3.1}        & -             & -                            & 89.1 \greenup{3.0}                            \\ \cmidrule(lr){3-5} \cmidrule(lr){6-8} \cmidrule(lr){9-11}
                                                     & LLaMa 7B (RT)                                         & 53.3          & 60.0 \greenup{6.70}          & \multicolumn{1}{l|}{73.3}          & 48.4        & 51.6 \greenup{3.2}                        & \multicolumn{1}{l|}{58.3}        & 55.1          & 67.7 \greenup{12.6}          & 76.7                            \\
                                                     & LLaMa 7B (RTE)                                        & -             & -                            & \multicolumn{1}{l|}{75.3}  \greenup{2.0}        & -           & -                           & \multicolumn{1}{l|}{60.2} \greenup{1.9}       & -             & -                            & 81.1 \greenup{4.4}                            \\ \cline{2-11} 
                                                     & LLaMa 13B (T)                                         & 60.2          & 62.2 \greenup{2.00}          & \multicolumn{1}{l|}{71.2}          & 53.4        & 55.7                        & \multicolumn{1}{l|}{64.3}        & 73.1 & 75.4 \greenup{2.30}          & 61.1                            \\
                                                     & LLaMa 13B (TE)                                        & -             & -                            & \multicolumn{1}{l|}{73.1} \greenup{1.9}          & -           & -                           & \multicolumn{1}{l|}{67.2} \greenup{2.9}       & -             & -                            & 71.1 \greenup{10.0}                            \\ \cmidrule(lr){3-5} \cmidrule(lr){6-8} \cmidrule(lr){9-11}
                                                     & LLaMa 13B (RT)                                        & 68.6          & \textbf{70.2} \greenup{1.60} & \multicolumn{1}{l|}{68.8}          & 60.2        & 65.7 \greenup{5.5}                        & \multicolumn{1}{l|}{58.1}        & 69.4          & 70.2 \greenup{0.80}          & 51.2                            \\
                                                     & LLaMa 13B (RTE)                                       & -             & -                            & \multicolumn{1}{l|}{70.3} \greenup{1.5}          & -           & -                           & \multicolumn{1}{l|}{62.2} \greenup{4.1}        & -             & -                            & 61.1 \greenup{9.9}                            \\ \cline{2-11} 
                                                     & Mistral 7B (T)                                        & 57.8          & 58.8 \greenup{1.00}          & \multicolumn{1}{l|}{74.4}          & 51.2        & 53.2 \greenup{2.0}                        & \multicolumn{1}{l|}{65.2}        & 64.8          & 66.3 \greenup{1.50}          & 75.2                            \\
                                                     & Mistral 7B (TE)                                       & -             & -                            & \multicolumn{1}{l|}{86.7} \greenup{4.0}         & -           & -                           & \multicolumn{1}{l|}{70.2} \greenup{5.0}        & -             & -                            & 86.7 \greenup{11.5}                            \\ \cmidrule(lr){3-5} \cmidrule(lr){6-8} \cmidrule(lr){9-11}
                                                     & Mistral 7B (RT)                                       & 55.4          & 57.4 \greenup{2.00}          & \multicolumn{1}{l|}{62.2}          & 48.4        & 50.2 \greenup{1.8}                        & \multicolumn{1}{l|}{54.3}        & 53.8          & 56.0 \greenup{2.20}          & 62.2                            \\
                                                     & Mistral 7B (RTE)                                      & -             & -                            & \multicolumn{1}{l|}{68.8} \greenup{6.6}         & -           & -                           & \multicolumn{1}{l|}{58.4} \greenup{4.1}        & -             & -                            & 68.8 \greenup{6.6}                            \\ \cline{2-11} 
                                                     & Qwen 7B (T)                                           & \textbf{68.9} & \textbf{70.4} \greenup{1.40}          & \multicolumn{1}{l|}{82.7}          & 61.2        & 63.4 \greenup{2.2}                        & \multicolumn{1}{l|}{73.2}        & 74.1          & 76.1 \greenup{1.20}          & 82.7                            \\
                                                     & Qwen 7B (TE)                                          & -             & -                            & \multicolumn{1}{l|}{86.7} \greenup{4.0}          & -           & -                           & \multicolumn{1}{l|}{74.4} \greenup{1.2}        & -             & -                            & 86.7 \greenup{4.0}                            \\ \cmidrule(lr){3-5} \cmidrule(lr){6-8} \cmidrule(lr){9-11}
                                                     & Qwen 7B (RT)                                          & 53.3          & 56.5 \greenup{3.20}          & \multicolumn{1}{l|}{60.2}          & 47.1        & 49.2 \greenup{2.1}                        & \multicolumn{1}{l|}{51.2}        & 53.3          & 56.5 \greenup{3.20}          & 60.2                            \\
                                                     & Qwen 7B (RTE)                                         & -             & -                            & \multicolumn{1}{l|}{62.2} \greenup{2.0}          & -           & -                           & \multicolumn{1}{l|}{55.4}  \greenup{4.2}       & -             & -                            & 62.2 \greenup{2.0}                            \\ \cline{2-11} 
                                                     & Yi-1.5 6B (T)                                         & 64.4          & 68.7 \greenup{4.3}          & \multicolumn{1}{l|}{71.3}          & 62.2        & 66.6 \greenup{4.4}                        & \multicolumn{1}{l|}{68.3}        & 71.1          & 73.4 \greenup{2.3}          & 72.3                            \\
                                                     & Yi-1.5 6B (TE)                                        & -             & -                            & \multicolumn{1}{l|}{74.5} \greenup{3.2}          & -           & -                           & \multicolumn{1}{l|}{72.8} \greenup{4.5}       & -             & -                            & 73.8 \greenup{1.5}                            \\ \cmidrule(lr){3-5} \cmidrule(lr){6-8} \cmidrule(lr){9-11}
                                                     & Yi-1.5 6B (RT)                                        & 63.3          & 68.3 \greenup{5.0}          & \multicolumn{1}{l|}{68.1}          &             &                             & \multicolumn{1}{l|}{69.2}        & 65.1          & 70.4 \greenup{5.30}          & 70.1                            \\
                                                     & Yi-1.5 6B (RTE)                                       & -             & -                            & \multicolumn{1}{l|}{70.1} \greenup{2.0}          & -           & -                           & \multicolumn{1}{l|}{71.6} \greenup{2.4}        & -             & -                            & 72.4 \greenup{2.3}                            \\ \cline{2-11} 
                                                     & SciT\"ulu 7B (T)                                      & 57.8          & 58.3 \greenup{0.50}          & \multicolumn{1}{l|}{51.1}          & 51.2        & 52.6\greenup{1.4}                        & \multicolumn{1}{l|}{45.3}        & 57.8          & 58.3 \greenup{0.50}          & 52.2                            \\
                                                     & SciT\"ulu 7B (TE)                                     & -             & -                            & \multicolumn{1}{l|}{72.2} \greenup{21.1}          & -           & -                           & \multicolumn{1}{l|}{64.3} \greenup{19.0}        & -             & -                            & 72.2 \greenup{20.0}                            \\ \cmidrule(lr){3-5} \cmidrule(lr){6-8} \cmidrule(lr){9-11}
                                                     & SciT\"ulu 7B (RT)                                     & 55.6          & 58.7 \greenup{3.10}          & \multicolumn{1}{l|}{71.1}          &             &                             & \multicolumn{1}{l|}{65.2}        & 55.6          & 58.7 \greenup{3.10}          & 68.8                            \\
                                                     & SciT\"ulu 7B (RTE)                                    & -             & -                            & \multicolumn{1}{l|}{\textbf{88.8}} \greenup{17.7}& -           & -                           & \multicolumn{1}{l|}{76.4} \greenup{11.2}       & -             & -                            & \textbf{91.1} \greenup{22.3}                   \\ \hline
\multirow{14}{*}{Instruction Tuned}                  & Gemma-1.1 7B (T)                                      & 57.8          & 61.4 \greenup{3.60}          & \multicolumn{1}{l|}{81.2}          & 50.2        & 54.6 \greenup{4.4}                       & \multicolumn{1}{l|}{75.3}        & 55.7          & 58.8 \greenup{3.10}          & 91.2                            \\
                                                     & Gemma-1.1 7B (RT)                                     & 55.6          & 59.4 \greenup{3.80}          & \multicolumn{1}{l|}{78.8}          & 53.2        & 52.5 \reddown{0.7}                       & \multicolumn{1}{l|}{71.5}        & 54.9          & 56.6 \greenup{1.70}          & 89.2                            \\ \cline{2-11} 
                                                     & LLaMa 7B (T)                                          & 62.7          & 65.4 \greenup{2.70}          & \multicolumn{1}{l|}{85.4}          & 52.1        & 56.3 \greenup{4.2}                        & \multicolumn{1}{l|}{68.3}        & 75.4          & 79.3 \greenup{3.90}          & 91.1                            \\
                                                     & LLaMa 7B (RT)                                         & 61.2          & 63.1 \greenup{1.90}          & \multicolumn{1}{l|}{81.3}          & 55.4        & 57.4 \greenup{2.0}                       & \multicolumn{1}{l|}{65.2}        & 71.3          & 77.2 \greenup{5.90}          & 88.6                            \\ \cline{2-11} 
                                                     & LLaMa 13B (T)                                         & 74.3          & 74.6 \greenup{0.30}          & \multicolumn{1}{l|}{75.3}          & 65.3        & 67.2 \greenup{1.9}                       & \multicolumn{1}{l|}{69.1}        & 75.2          & 77.8 \greenup{2.60}          & 75.2                            \\
                                                     & LLaMa 13B (RT)                                        & 70.2          & 71.8 \greenup{1.60}          & \multicolumn{1}{l|}{73.3}          & 62.4        & 63.7 \greenup{1.3}                       & \multicolumn{1}{l|}{65.9}        & 71.2          & 72.3 \greenup{1.10}          & 74.2                            \\ \cline{2-11} 
                                                     & Mistral 7B (T)                                       & 60.2          & 62.2 \greenup{2.00}          & \multicolumn{1}{l|}{86.4}          & 54.6        & 56.9 \greenup{2.3}                        & \multicolumn{1}{l|}{68.4}        & 70.2          & 74.3 \greenup{4.10}          & 88.1                            \\  
                                                     & Mistral 7B (RT)                                        & 65.3          & 68.2 \greenup{2.90}          & \multicolumn{1}{l|}{88.2}          & 57.2        & 59.4 \greenup{2.2}                        & \multicolumn{1}{l|}{75.3}        & 68.2          & 70.2 \greenup{2.00}          & 89.2                            \\
                                                     \cline{2-11} 
                                                     & Qwen 7B (T)                                           & \textbf{75.4} & \textbf{76.3} \greenup{0.90} & \multicolumn{1}{l|}{88.4}          & 68.2        & 69.0 \greenup{0.8}                        & \multicolumn{1}{l|}{76.2}        & 74.8          & 76.2 \greenup{1.40}          & 89.4                            \\
                                                     & Qwen 7B (RT)                                          & 73.2          & 74.1 \greenup{0.90}          & \multicolumn{1}{l|}{86.3}          & 65.1        & 68.2 \greenup{3.1}                       & \multicolumn{1}{l|}{74.2}        & 72.8          & 74.0 \greenup{1.20}          & 86.2                            \\ \cline{2-11} 
                                                     & Yi-1.5 6B (T)                                         & 69.5          & 74.2 \greenup{5.30}          & \multicolumn{1}{l|}{78.4}          & 68.2        & 72.1 \greenup{3.9}                        & \multicolumn{1}{l|}{74.1}        & 72.3          & 75.8 \greenup{3.50}           & 79.3                            \\
                                                     & Yi-1.5 6B (RT)                                        & 67.2          & 69.4 \greenup{2.20}          & \multicolumn{1}{l|}{73.2}          & 66.2        & 70.1 \greenup{3.9}                        & \multicolumn{1}{l|}{74.1}        & 69.3          & 72.2 \greenup{2.90}          & 75.2                            \\ \cline{2-11} 
                                                     & SciT\"ulu 7B (T)                                      & 66.3          & 68.4 \greenup{2.10}          & \multicolumn{1}{l|}{\textbf{91.2}} & 58.1        & 60.2 \greenup{2.1}                        & \multicolumn{1}{l|}{80.3}        & \textbf{76.8} & \textbf{78.8} \greenup{2.00} & \textbf{92.3}                   \\
                                                     & SciT\"ulu 7B (RT)                                     & 62.4          & 65.6 \greenup{3.20}          & \multicolumn{1}{l|}{87.2}          & 56.1        & 58.1 \greenup{2.0}                        & \multicolumn{1}{l|}{76.1}        & 72.3          & 73.3 \greenup{1.00}          & 89.9                            \\ \hline
\end{tabular}}
\caption{Performance of LLMs on different rounds of annotation for the coarse classification task. S.A represents the string-matching evaluator, and G.A represents the GPT-based evaluator. `T' represents prompting with only the target sentence, RT represents the combination of the review and the target sentence. Adding demonstrations to the prompt is represented by `E'. R1, R2 and R3 represent `Round 1', `Round 2', and `Round 3' respectively.}
\label{tab:ann_round_llm_coarse}
\end{table*}

\begin{table*}[]
\centering
\resizebox{!}{0.2\textwidth}{\begin{tabular}{l|l|l|ll|ll|ll|ll|ll|ll|ll}
\hline
\multicolumn{1}{l|}{\multirow{2}{*}{\textbf{Method}}}                          & \multicolumn{1}{c|}{\multirow{2}{*}{\textbf{Mode}}} & \multicolumn{1}{c|}{\multirow{2}{*}{\textbf{Mix}}} & \multicolumn{2}{c|}{\textbf{Gemma-1.1 7B}}          & \multicolumn{2}{c|}{\textbf{LLaMa 7B}}              & \multicolumn{2}{c|}{\textbf{LLaMa 13B}}             & \multicolumn{2}{c|}{\textbf{Mistral 7B}}            & \multicolumn{2}{c|}{\textbf{Qwen 7B}}                                         & \multicolumn{2}{c|}{\textbf{Yi-1.5 6B}}                                         & \multicolumn{2}{c}{\textbf{SciT\"ul\"u 7B}}         \\ \cline{4-17} 
\multicolumn{1}{l|}{}                                                        & \multicolumn{1}{c|}{}                               & \multicolumn{1}{c|}{}                              & S                        & \multicolumn{1}{l|}{G}   & S                        & \multicolumn{1}{l|}{G}   & S                        & \multicolumn{1}{l|}{G}   & S                        & \multicolumn{1}{l|}{G}   & S                        & \multicolumn{1}{l|}{G}                             & S                                                  & \multicolumn{1}{l|}{G}     & S                        & G                        \\ \hline
\multirow{2}{*}{Zero-shot}                                                   & T                                                   & -                                                  & 18.1$_{3.2}$             & 23.0$_{4.6}$             & 14.2$_{4.5}$             & 24.5$_{5.6}$             & 17.8$_{2.9}$             & 26.5$_{3.4}$             & 21.0$_{4.6}$             & 23.0$_{3.8}$             & 21.1$_{1.1}$             & 24.5$_{1.1}$                                       & 24.2$_{1.1}$                                       & 30.5$_{1.2}$               & 11.1$_{2.2}$             & 14.0$_{2.4}$             \\
                                                                             & TE                                                  & -                                                  & 22.2$_{3.4}$             & 27.1$_{3.4}$             & 22.2$_{5.4}$             & 28.4$_{5.8}$             & 21.4$_{2.2}$             & 29.3$_{2.4}$             & 18.6$_{4.7}$             & 25.5$_{4.9}$             & 22.2$_{1.1}$             & 28.2$_{1.5}$                                       & 29.1$_{1.8}$                                       & 36.5$_{2.0}$               & 11.2$_{2.4}$             & 20.6$_{2.6}$             \\
                                                                             &                                                     &                                                    & \greenup{4.1}            & \greenup{4.1}            & \greenup{8.0}            & \greenup{3.9}            & \greenup{3.6}            & \greenup{2.8}            & \greenup{2.4}            & \greenup{2.5}            & \greenup{1.1}            & \greenup{3.7}                                      & \greenup{5.9}                                      & \greenup{6.0}              & \greenup{0.1}            & \greenup{6.6}            \\ \hline
\multirow{8}{*}{\begin{tabular}[c]{@{}l@{}}Instruction\\ Tuned\end{tabular}} & \multirow{8}{*}{T}                                  & No Mix                                             & 29.5$_{3.4}$             & 36.5$_{3.4}$             & 25.4$_{4.8}$             & 30.5$_{4.6}$             & 29.7$_{3.3}$             & 35.5$_{3.6}$             & 27.1$_{4.4}$             & 30.5$_{4.4}$             & 27.9$_{1.4}$             & 30.5$_{1.4}$                                       & 44.8$_{0.8}$                                       & 50.5$_{1.4}$               & 15.2$_{2.4}$             & 23.0$_{2.6}$             \\
                                                                             &                                                     &                                                    & \greenup{11.4}           & \greenup{13.5}           & \greenup{11.2}           & \greenup{6.0}            & \greenup{11.9}           & \greenup{11.0}           & \greenup{6.1}            & \greenup{7.5}            & \greenup{6.8}            & \greenup{6.0}                                      & \greenup{20.6}                                     & \greenup{20.0}             & \greenup{4.1}            & \greenup{9.0}            \\ \cline{3-17}
                                                                             &                                                     & SciRiff Mix                                        & 23.5$_{3.4}$             & 29.5$_{3.3}$             & \underline{38.6}$_{4.1}$ & \underline{46.3}$_{4.6}$            & \underline{32.0}$_{3.4}$ & 36.0$_{3.2}$             & 24.0$_{4.2}$             & 26.5$_{4.2}$             & 27.6$_{1.4}$             & 31.5$_{1.6}$                                       & 45.4$_{1.2}$                                       & 51.5$_{1.7}$               & \underline{28.8}$_{2.4}$ & \underline{36.3}$_{2.4}$             \\ 
                                                                             &                                                     &                                                    & \greenup{5.4}            & \greenup{6.5}            & \greenup{10.6}           & \greenup{6.0}            & \greenup{14.2}           & \greenup{10.5}           & \greenup{3.0}            & \greenup{3.5}            & \greenup{6.5}            & \greenup{7.0}                                      & \greenup{21.2}                                     & \greenup{21.0}             & \greenup{17.7}           & \greenup{22.3}            \\ \cline{3-17}
                                                                             &                                                     & T\"ulu Mix                                         & \underline{31.6}$_{3.0}$ & \underline{39.5}$_{3.1}$ & 21.7$_{4.2}$             & 26.5$_{4.4}$             & 28.4$_{2.9}$             & 33.0$_{3.0}$             & 25.7$_{4.4}$             & 26.5$_{4.4}$             & \underline{45.5}$_{1.1}$ & \textcolor{cyan}{\textbf{\underline{51.0}$_{1.3}$}} & 27.2$_{1.0}$                                       & 31.0$_{1.0}$               & 16.5$_{2.2}$             & 21.5$_{2.4}$             \\ 
                                                                             &                                                     &                                                    & \greenup{13.5}           & \greenup{16.5}           & \greenup{7.5}            & \greenup{2.0}            & \greenup{10.6}           & \greenup{6.5}            & \greenup{4.7}            & \greenup{3.5}            & \greenup{24.4}           & \greenup{26.5}                                     & \greenup{3.0}                                      & \greenup{1.5}              & \greenup{5.4}            & \greenup{7.5}            \\ \cline{3-17}
                                                                             &                                                     & Full Mix                                           & 25.4$_{3.0}$             & 30.0$_{3.0}$             & 28.3$_{4.2}$             & 35.5$_{4.0}$             & 31.5$_{3.1}$             & 37.0$_{3.2}$             & \underline{28.6}$_{4.2}$ & \underline{32.3}$_{4.4}$             & 31.9$_{1.0}$             & 35.5$_{1.0}$                                       & \textcolor{cyan}{\textbf{\underline{45.7}}$_{0.8}$} & {\underline{50.0}}$_{0.9}$ & 20.4$_{2.0}$             & 26.5$_{2.2}$             \\ 
                                                                             &                                                     &                                                    & \greenup{7.3}            & \greenup{7.0}            & \greenup{14.1}           & \greenup{11.0}           & \greenup{13.7}           & \greenup{10.5}           & \greenup{7.6}            & \greenup{9.3}            & \greenup{10.8}           & \greenup{11.0}                                     & \greenup{11.5}                                     & \greenup{19.5}             & \greenup{9.3}            & \greenup{12.5}           \\ \hline
\multirow{2}{*}{Zero-shot}                                                   & RT                                                  & -                                                  & 17.2$_{4.1}$             & 21.8$_{4.6}$             & 10.1$_{5.2}$             & 16.3$_{6.1}$             & 14.5$_{3.1}$             & 28.1$_{3.1}$             & 17.8$_{4.1}$             & 22.0$_{4.9}$             & 18.0$_{1.1}$             & 22.8$_{1.9}$                                       & 21.6$_{1.1}$                                       & 29.8$_{1.1}$               & 08.2$_{2.1}$             & 14.4$_{2.1}$             \\
                                                                             & RTE                                                 & -                                                  & 18.7$_{3.4}$             & 25.8$_{3.6}$             & 12.4$_{5.4}$             & 20.4$_{5.1}$             & 21.5$_{2.1}$             & 29.1$_{2.9}$             & 18.4$_{4.1}$             & 23.3$_{4.6}$             & 19.3$_{1.1}$             & 26.6$_{1.1}$                                       & 25.4$_{0.8}$                                       & 31.1$_{0.6}$               & 12.7$_{2.1}$             & 18.8$_{2.2}$             \\
                                                                             &                                                     &                                                    & \greenup{1.5}            & \greenup{4.0}            & \greenup{2.3}            & \greenup{4.1}            & \greenup{7.0}            & \greenup{1.0}            & \greenup{0.6}            & \greenup{1.3}            & \greenup{1.3}            & \greenup{3.8}                                      & \greenup{3.8}                                      & \greenup{1.3}              & \greenup{4.5}            & \greenup{4.4}            \\ \hline
\multirow{8}{*}{\begin{tabular}[c]{@{}l@{}}Instruction\\ Tuned\end{tabular}} & \multirow{8}{*}{RT}                                 & No Mix                                             & 26.8$_{3.1}$             & 28.8$_{3.2}$             & 22.9$_{4.4}$             & 28.8$_{4.4}$             & 28.4$_{1.4}$             & 30.6$_{1.6}$             & 24.1$_{4.4}$             & 27.6$_{4.4}$             & 22.6$_{1.4}$             & 29.6$_{1.1}$                                       & 28.7$_{0.4}$                                       & 36.9$_{0.4}$               & 31.5$_{3.4}$             & 34.4                     \\
                                                                             &                                                     &                                                    & \greenup{9.6}            & \greenup{7.0}            & \greenup{12.8}           & \greenup{12.5}           & \greenup{13.9}           & \greenup{12.5}           & \greenup{6.3}            & \greenup{5.6}            & \greenup{4.6}            & \greenup{6.8}                                      & \greenup{7.1}                                      & \greenup{7.1}              & \greenup{23.3}           & \greenup{20.0}           \\ \cline{3-17}
                                                                             &                                                     & SciRiff Mix                                        & 22.2$_{3.2}$             & 26.0$_{3.2}$             & {33.6}$_{4.1}$ & {42.3}$_{4.6}$ & 31.5$_{3.9}$             & \underline{37.3}$_{3.6}$ & 18.6$_{4.4}$             & 24.4$_{4.2}$             & 22.5$_{0.4}$             & 26.4$_{0.6}$                                       & 38.5$_{1.4}$                                       & 41.3$_{1.7}$               & {24.8}$_{2.4}$ & {32.3}$_{2.4}$ \\ 
                                                                             &                                                     &                                                    & \greenup{5.0}            & \greenup{4.2}            & \greenup{28.5}           & \greenup{30.0}           & \greenup{17.0}           & \greenup{11.2}           & \greenup{0.8}            & \greenup{2.4}            & \greenup{4.5}            & \greenup{3.6}                                      & \greenup{16.9}                                     & \greenup{11.5}             & \greenup{16.6}           & \greenup{17.9}           \\ \cline{3-17}
                                                                             &                                                     & T\"ulu Mix                                         & 25.7$_{3.0}$             & 30.6$_{3.0}$             & 28.6$_{4.2}$             & 33.1$_{4.0}$             & 27.2$_{3.0}$             & 33.1$_{3.0}$             & 21.0$_{3.8}$             & 25.0$_{3.8}$             & 25.1$_{1.2}$             & 31.3$_{1.6}$                                       & 35.8$_{0.8}$                                       & 40.6$_{0.9}$               & 28.7$_{2.0}$             & 35.6$_{2.2}$             \\ 
                                                                             &                                                     &                                                    & \greenup{8.5}            & \greenup{8.8}            & \greenup{18.5}           & \greenup{16.8}           & \greenup{12.7}           & \greenup{5.0}            & \greenup{3.2}            & \greenup{3.0}            & \greenup{7.1}            & \greenup{8.5}                                      & \greenup{14.2}                                     & \greenup{10.8}             & \greenup{20.5}           & \greenup{21.2}           \\ \cline{3-17}
                                                                             &                                                     & Full Mix                                           & 21.4$_{3.2}$             & 27.0$_{3.1}$             & 22.8$_{4.0}$             & 29.0$_{4.0}$             & 25.4$_{3.0}$             & 29.0$_{2.9}$             & {27.6}$_{4.2}$ & {28.3}$_{4.4}$ & 22.7$_{1.1}$             & 28.0$_{1.2}$                                       & 25.4$_{0.8}$                                       & 32.2$_{0.9}$               & 26.8$_{2.2}$             & 31.4$_{2.4}$             \\ 
                                                                             &                                                     &                                                    & \greenup{4.2}            & \greenup{5.2}            & \greenup{11.7}           & \greenup{12.7}           & \greenup{10.9}           & \greenup{0.9}            & \greenup{9.8}           & \greenup{6.3}            & \greenup{4.7}            & \greenup{5.2}                                      & \greenup{3.8}                                      & \greenup{1.4}              & \greenup{18.6}           & \greenup{17.0} \\  \hline          
\end{tabular}}
\caption{Performance of LLMs for \hl{fine-grained classification} with 3-fold cross-validation on \textsc{$\textsc{LazyReview}$} test sets. `S' represents the string-matching evaluator, and `G' represents the GPT-based evaluator reporting \textbf{accuracies}. `T' represents prompting with only the target sentence, RT represents the combination of the review and the target sentence. Adding demonstrations to the prompt is represented by `E'. The best results for this task is highlighted in \textcolor{cyan}{cyan}. Increments are shown with the classic zero-shot setup without exemplars (first row for zero-shots) with T or RT.}
\label{tab:test_set}
\end{table*}

\begin{table*}[]
\centering
\resizebox{!}{0.2\textwidth}{\begin{tabular}{l|l|l|ll|ll|ll|ll|ll|ll|ll}
\hline
\cline{2-17}
\multirow{2}{*}{\textbf{Task}}          & \multicolumn{1}{c}{\multirow{2}{*}{\textbf{Mode}}} & \multicolumn{1}{c}{\multirow{2}{*}{\textbf{Mix}}} & \multicolumn{2}{c}{\textbf{Gemma-1.1 7B}} & \multicolumn{2}{c}{\textbf{LLaMa 7B}} & \multicolumn{2}{c}{\textbf{LLaMa 13B}} & \multicolumn{2}{c}{\textbf{Mistral 7B}} & \multicolumn{2}{c}{\textbf{Qwen 7B}} & \multicolumn{2}{c}{\textbf{Yi-1.5 6B}} & \multicolumn{2}{c}{\textbf{SciT\"ul\"u 7B}} \\ \cline{4-17} 
                               & \multicolumn{1}{c}{}                               & \multicolumn{1}{c}{}                              & S                 & G                 & S                 & G                 & S                  & G                 & S                 & G                & S               & G               & S              & G              & S                 & G                \\ \hline

\multirow{2}{*}{Zero-Shot} & T                                  &  -                 &  57.0$_{2.9}$                & 66.0$_{2.8}$                  &    52.2$_{3.9}$              &   57.8$_{3.9}$                 &  51.8$_{2.6}$                 &   58.6$_{2.8}$                & 51.1$_{3.8}$                 &      61.1$_{3.6}$           &  53.4$_{1.0}$               &   63.5$_{1.0}$             &    57.2$_{1.2}$           &      65.5$_{1.4}$             &   27.3$_{2.6}$        & 32.0$_{2.8}$       \\
& TE                                 &    -               &    63.1$_{3.1}$               &   65.2$_{3.2}$                &    58.2$_{4.2}$              & 60.2$_{4.4}$                   &     59.4$_{3.2}$              &   61.1$_{3.0}$                &    59.1$_{4.0}$              & 63.4$_{4.2}$                &   61.6$_{0.8}$              &   65.7$_{0.7}$             &    62.2$_{1.4}$           &  67.8$_{1.6}$                 & 29.1$_{2.6}$  & 34.8$_{2.9}$               \\
& & & \greenup{6.0} & \reddown{0.8} & \greenup{6.0} & \greenup{2.4} & \greenup{7.6} & \greenup{2.5} & \greenup{8.0} & \greenup{2.3} & \greenup{8.2} & \greenup{2.2} & \greenup{5.0} & \greenup{2.3} & \greenup{1.8} & \greenup{2.8} \\
\hline 
\multirow{8}{*}{\begin{tabular}[c]{@{}l@{}}Instruction\\ Tuned\end{tabular}} & \multirow{8}{*}{T}                                 & No Mix                           &    69.3$_{3.2}$               &       71.5$_{3.2}$            &     69.2$_{4.3}$              &      71.5$_{4.1}$            &         {60.4}$_{2.9}$           &         {65.0}$_{3.0}$          &  61.0$_{4.2}$                 &      65.2$_{4.0}$             &   {63.3}$_{0.8}$             &        {64.0}$_{1.0}$          &    68.2$_{1.0}$           &      71.5$_{1.2}$          &     69.6$_{2.8}$             &   71.5$_{2.9}$               \\
& & & \greenup{12.3} & \greenup{5.5} & \greenup{12.0} & \greenup{13.8} & \greenup{8.6} & \greenup{6.4} & \greenup{9.9} & \greenup{4.1} & \greenup{15.9} & \greenup{0.5} & \greenup{11.0} & \greenup{6.0} & \greenup{42.0} & \greenup{39.5} \\
&                                  & SciRiFF Mix                   &   69.4$_{3.0}$                 &   71.5$_{3.0}$                 &      \underline{69.8}$_{3.0}$              &  \underline{72.5}$_{3.0}$                 &  \underline{68.4}$_{2.9}$           &         \underline{72.0}$_{3.0}$                 &    68.4$_{3.0}$                &   70.5$_{3.0}$                &    60.4$_{3.0}$              &   64.5$_{3.0}$               &   69.4$_{3.0}$              &  71.5$_{3.0}$              &   \underline{\textbf{\textcolor{blue}{69.8$_{3.0}$}}}                &   \underline{\textbf{\textcolor{blue}{73.5$_{3.0}$}}}                \\
& & & \greenup{12.4} & \greenup{5.5} & \greenup{17.6} & \greenup{14.7} & \greenup{16.6} & \greenup{13.4} & \greenup{17.3} & \greenup{9.4} & \greenup{7.0} & \greenup{1.0} & \greenup{12.2} & \greenup{6.0} & \greenup{42.5} & \greenup{41.5} \\
&                                  &  T\"ulu Mix                    &  \underline{69.6}$_{3.2}$                  &   \underline{72.4}$_{3.2}$                 &     65.4$_{3.8}$               &  67.0$_{3.8}$                 &  65.3$_{2.8}$                   &      67.3$_{2.8}$              &   56.2$_{3.8}$                 &   62.5$_{3.8}$                &   \underline{69.3}$_{0.8}$             &        \underline{72.0}$_{1.0}$               &   66.3$_{1.5}$              &   68.0$_{1.7}$             &   65.2$_{2.8}$                 &     67.0$_{2.8}$              \\
& & & \greenup{12.6} & \greenup{6.4} & \greenup{13.2} & \greenup{9.2} & \greenup{13.5} & \greenup{8.7} & \greenup{5.1} & \greenup{1.4} & \greenup{15.9} & \greenup{8.5} & \greenup{9.1} & \greenup{2.5} & \greenup{37.9} & \greenup{35.0} \\
&                                  & Full Mix          &     69.4$_{3.3}$              &   71.5$_{3.4}$                &    69.8$_{4.1}$               &   71.5$_{4.4}$               &   68.4$_{3.4}$                 &    68.2$_{3.6}$               &  \underline{71.3}$_{4.0}$                   &      \underline{72.0}$_{4.2}$           & 63.8$_{0.8}$                &      64.0$_{0.6}$           &   \underline{69.8}$_{1.8}$             &      \underline{72.0}$_{1.2}$         &   {{{60.8}}$_{2.8}$}                &       {{{68.5}}$_{2.7}$}           \\ 
& & & \greenup{12.4} & \greenup{5.5} & \greenup{17.6} & \greenup{13.7} & \greenup{16.6} & \greenup{9.6} & \greenup{20.2} & \greenup{10.9} & \greenup{10.4} & \greenup{0.5} & \greenup{12.6} & \greenup{6.5} & \greenup{33.5} & \greenup{36.5} \\ \hline
\multirow{2}{*}{Zero-Shot} & RT                                 &      -             &  58.2$_{3.1}$                 &  61.3$_{3.1}$                 &   51.5$_{4.2}$               &    55.6$_{4.2}$                &  54.8$_{3.6}$                 &   57.6$_{3.4}$                &   58.8$_{4.2}$               &    60.0$_{4.0}$             &  51.0$_{0.8}$               &   55.0$_{0.7}$             &     50.4$_{1.0}$          &  52.5$_{1.1}$                 & 27.8$_{2.4}$      & 31.8$_{2.6}$           \\           
& RTE                                &    -               &     59.9$_{3.0}$              &    63.4$_{3.0}$               &  56.1$_{4.0}$                &    58.2$_{4.1}$                &     58.4$_{3.8}$              &  61.3$_{3.6}$                 &   57.4$_{4.0}$               &   61.1$_{4.1}$              &      54.5$_{0.7}$           &    57.8$_{0.8}$            &     51.8$_{1.2}$          &    54.5$_{1.4}$               & 32.8$_{2.8}$ &  36.9$_{2.9}$              \\ 
& & & \greenup{1.7} & \greenup{2.1} & \greenup{13.3} & \greenup{11.3} & \greenup{3.6} & \greenup{3.7} & \reddown{1.4} & \greenup{1.1} & \greenup{3.5} & \greenup{2.8} & \greenup{1.4} & \greenup{2.0} & \greenup{5.0} & \greenup{5.1} \\
\hline

\multirow{8}{*}{\begin{tabular}[c]{@{}l@{}}Instruction\\ Tuned\end{tabular}} & \multirow{8}{*}{RT}                                & No Mix                          &  62.8$_{3.2}$                 &   65.9$_{3.1}$                &   64.8$_{4.0}$                &     66.9$_{4.2}$             &  60.2$_{2.7}$                  &    62.6$_{3.0}$               &   65.4$_{4.0}$                &    67.5$_{4.0}$              &   59.2$_{1.0}$              &      61.9$_{0.8}$           &    63.8$_{1.1}$            &    65.0$_{1.2}$           &   64.6$_{2.4}$                &      66.9$_{2.4}$            \\ 
& & & \greenup{4.6} & \greenup{4.6} & \greenup{13.3} & \greenup{11.3} & \greenup{5.4} & \greenup{5.0} & \greenup{6.6} & \greenup{7.5} & \greenup{8.2} & \greenup{6.9} & \greenup{13.4} & \greenup{12.5} & \greenup{36.8} & \greenup{35.1} \\

&                                 & SciRiFF Mix                    &    63.4$_{3.2}$                &      66.3$_{3.0}$              &  66.5$_{3.8}$                  &    66.9$_{3.9}$               & 58.2$_{3.0}$                    &    60.6$_{2.8}$                &     66.4$_{3.4}$               &        68.1$_{3.3}$           &  58.5$_{1.0}$                &    60.6$_{0.8}$              &     64.3$_{1.2}$            &  66.2$_{1.4}$              &    64.3$_{2.6}$                &     66.9$_{2.8}$              \\ 
& & & \greenup{5.2} & \greenup{5.0} & \greenup{1.7} & \greenup{11.3} & \greenup{3.4} & \greenup{3.0} & \greenup{7.6} & \greenup{8.1} & \greenup{7.5} & \greenup{5.6} & \greenup{13.9} & \greenup{13.7} & \greenup{36.5} & \greenup{35.1} \\

&                                 & T\"ulu Mix                    &   64.3$_{3.1}$                 &    66.3$_{3.2}$                &    65.4$_{4.2}$                &    67.5$_{4.3}$               &  62.8$_{2.8}$                   &    66.9$_{2.8}$                &     63.9$_{2.8}$               &    66.9$_{2.8}$               &  59.2$_{2.8}$                &   61.0$_{2.8}$               &      62.4$_{2.8}$           &   66.3$_{2.8}$             &   63.8$_{2.8}$                 &       66.9$_{2.8}$            \\ 
& & & \greenup{6.1} & \greenup{5.0} & \greenup{13.9} & \greenup{11.9} & \greenup{8.0} & \greenup{9.3} & \greenup{5.1} & \greenup{6.9} & \greenup{8.2} & \greenup{6.0} & \greenup{12.0} & \greenup{13.8} & \greenup{36} & \greenup{35.1} \\

&                                 & Full Mix         &   64.3$_{3.3}$                &   66.9$_{3.5}$                &   65.3$_{4.2}$                &   66.9$_{4.3}$               &     65.4$_{3.2}$               &     66.9$_{3.4}$              &   64.8$_{4.3}$                &   66.9$_{4.4}$               &  56.4$_{0.4}$               &     58.1$_{0.8}$            &    62.8$_{1.8}$            &  65.6$_{1.6}$             & 64.5$_{2.4}$                  &      66.9$_{2.6}$            \\ 
& & & \greenup{6.1} & \greenup{5.6} & \greenup{13.8} & \greenup{11.3} & \greenup{9.6} & \greenup{9.3} & \greenup{6.0} & \greenup{6.9} & \greenup{5.0} & \greenup{3.1} & \greenup{12.4} & \greenup{13.1} & \greenup{36.7} & \greenup{35.1} \\
\hline
\end{tabular}}
\caption{Performance of LLMs on 3-fold cross-validation on \textsc{$\textsc{LazyReview}$} test sets for \hl{coarse-grained classification}. S represents the string-matching evaluator, and G represents the GPT-based evaluator reporting \textbf{accuracies}. `T' represents prompting with only the target sentence, RT represents the combination of the review and the target sentence. Adding demonstrations to the prompt is represented by `E'. The best results for this task is highlighted in \textcolor{blue}{blue}. Increments are shown with the classic zero-shot setup without exemplars (first row for zero-shots) with T or RT.}
\label{tab:test_set_coarse}
\end{table*}

\begin{table}[!t]
\resizebox{!}{0.16\textwidth}{\begin{tabular}{lllll|lll}
\hline
\multirow{2}{*}{\textbf{Model}}              & \multirow{2}{*}{\textbf{Class}} & \multicolumn{3}{l}{\textbf{Fine-grained}}                                                         & \multicolumn{3}{l}{\textbf{Coarse-grained}}   \\ \cline{3-8} 
                                             &                                 & \multicolumn{1}{c}{\textbf{P}} & \multicolumn{1}{c}{\textbf{R}} & \multicolumn{1}{c}{\textbf{F1}} & \textbf{P} & \textbf{R} & \textbf{F1} \\ \hline
\multirow{2}{*}{gemma-1.1-7b-it}             & Correct                         & 0.56                           & 1.00                           & 0.71                            & 0.95       & 0.92       & 0.93        \\
                                             & Incorrect                       & 1.00                           & 0.91                           & 0.95                            & 1.00       & 0.92       & 0.95        \\ \hline
\multirow{2}{*}{Llama-7B-chat}               & Correct                         & 1.00                           & 1.00                           & 1.00                            & 0.95       & 0.98       & 0.97        \\
                                             & Incorrect                       & 1.00                           & 1.00                           & 1.00                            & 1.00       & 1.00       & 1.00        \\ \hline
\multirow{2}{*}{Llama-13b-chat}              & Correct                         & 0.88                           & 1.00                           & 0.93                            & 0.95       & 0.96       & 0.96        \\
                                             & Incorrect                       & 1.00                           & 0.97                           & 0.99                            & 1.00       & 1.00       & 1.00        \\ \hline
\multirow{2}{*}{Mistral-7B-instruct}         & Correct                         & 0.83                           & 1.00                           & 0.91                            & 0.92       & 0.98       & 0.95        \\
                                             & Incorrect                       & 1.00                           & 0.97                           & 0.91                            & 1.00       & 1.00       & 1.00        \\ \hline
\multirow{2}{*}{Qwen-7B-chat}                & Correct                         & 0.86                           & 1.00                           & 0.92                            & 1.00       & 1.00       & 1.00        \\
                                             & Incorrect                       & 1.00                           & 0.97                           & 0.99                            & 1.00       & 1.00       & 1.00        \\ \hline
\multirow{2}{*}{Yi-6B-chat}                  & Correct                         & 1.00                           & 0.92                           & 0.96                            & 1.00       & 1.00       & 1.00        \\
                                             & Incorrect                       & 0.97                           & 1.00                           & 0.98                            & 1.00       & 1.00       & 1.00        \\ \hline
\multirow{2}{*}{SciT\"ulu 7B} & Correct                         & 1.00                           & 0.95                           & 0.98                            & 0.91       & 0.94       & 0.98        \\
                                             & Incorrect                       & 1.00                           & 1.00                           & 0.98                            & 1.00       & 1.00       & 1.00        \\ \hline
\end{tabular}}
\caption{Manual evaluation of the string-based evaluator for each model in terms of Precision (P), Recall (R), and F1 scores.}
\label{tab:str_eval}
\end{table}

\begin{table}[!t]
\resizebox{!}{0.16\textwidth}{\begin{tabular}{lllll|lll}
\hline
\multirow{2}{*}{\textbf{Model}}                       & \multirow{2}{*}{\textbf{Class}} & \multicolumn{3}{l}{\textbf{Fine-grained}}                                                                  & \multicolumn{3}{l}{\textbf{Coarse-grained}}                                                                        \\ \cline{3-8} 
                                             &                        & \multicolumn{1}{c}{\textbf{P}} & \multicolumn{1}{c}{\textbf{R}} & \multicolumn{1}{c}{\textbf{F1}} & \multicolumn{1}{c}{\textbf{P}} & \multicolumn{1}{c}{\textbf{R}} & \multicolumn{1}{c}{\textbf{F1}} \\ \hline
\multirow{2}{*}{Gemma-1.1-7b-it}             & Correct                & 0.69                           & 0.96                           & 0.81                            & 1.00                           & 1.00                           & 1.00                            \\
                                             & Incorrect              & 0.89                           & 0.42                           & 0.57                            & 1.00                           & 1.00                           & 1.00                            \\ \hline
\multirow{2}{*}{LlaMa-7B-chat}               & Correct                & 0.76                           & 1.00                           & 0.87                            & 1.00                           & 1.00                           & 1.00                            \\
                                             & Incorrect              & 0.89                           & 0.42                           & 0.57                            & 1.00                           & 1.00                           & 1.00                            \\ \hline
\multirow{2}{*}{LlaMa-13b-chat}              & Correct                & 0.76                           & 1.00                           & 0.87                            & 1.00                           & 1.00                           & 1.00                            \\
                                             & Incorrect              & 1.00                           & 0.44                           & 0.61                            & 1.00                           & 1.00                           & 1.00                            \\ \hline
\multirow{2}{*}{Mistral-7B-instruct}         & Correct                & 0.82                           & 1.00                           & 0.90                            & 1.00                           & 1.00                           & 1.00                            \\
                                             & Incorrect              & 1.00                           & 0.36                           & 0.53                            & 1.00                           & 1.00                           & 1.00                            \\ \hline
\multirow{2}{*}{Qwen-7B-chat}                & Correct                & 0.89                           & 1.00                           & 0.94                            & 1.00                           & 1.00                           & 1.00                            \\
                                             & Incorrect              & 1.00                           & 0.42                           & 0.59                            & 1.00                  & 1.00                           & 1.00                            \\ \hline
\multirow{2}{*}{Yi-6B-chat}                  & Correct                & 0.88                           & 1.00                           & 0.93                            & 1.00                           & 1.00                           & 1.00                            \\
                                             & Incorrect              & 1.00                           & 0.76                           & 0.87                            & 1.00                           & 1.00                           & 1.00                            \\ \hline
\multirow{2}{*}{SciT\"ulu 7B} & Correct                & 0.83                           & 1.00                           & 0.94                            & 1.00                           & 1.00                           & 1.00                            \\
                                             & Incorrect              & 1.00                           & 0.86                           & 0.88                            & 1.00                           & 1.00                           & 1.00                            \\ \hline
\end{tabular}}
\caption{Manual evaluation of the GPT-3.5 evaluator for each model in terms of Precision (P), Recall (R), and F1 scores.}
\label{tab:gpt_35_manual}
\end{table}

\subsection{Metric alignment study} \label{sec:metrics}
To verify the \hl{reliability of our metrics}, we follow \citet{holtermann2024evaluatingelementarymultilingualcapabilities} and perform an alignment study. We sample 50 responses from various models and evaluate them with both string matching and GPT-based methods. We report the results in Tables~\ref{tab:str_eval} and \ref{tab:gpt_35_manual} respectively. We find that the string-based evaluator is precise for correct answers but tends to underestimate them, while the GPT-based evaluator is better at identifying incorrect answers but may overestimate correct ones. As such, the string-based evaluator serves as a lower bound while the GPT-based evaluator serves as an upper bound. Manual analysis shows that the GPT-based evaluator struggles with similar classes, like "not surprising" versus "not novel," while the string-based evaluator misses correct predictions such as "The writing needs to be clear" and "Language errors/Writing style." However, for the coarse-grained classification, both the evaluators perform equally well.
\subsection{Prompts for Coarse-grained and Fine-grained classification} \label{sec:prompt_details}
We use a fixed prompt template to describe the \emph
{lazy thinking} as shown in Fig~\ref{fig:prompt}. We then add the task-specific prompt for fine-grained and coarse-grained classification as shown in Fig~\ref{fig:prompt_fine} and Fig~\ref{fig:prompt_coarse}, respectively. For the first round, we add the classes in the fixed template from the initial ARR guidelines (Table~\ref{tab:full_arr}). In the second round, the classes are added from the enhanced guidelines in Table~\ref{tab:arr_guidelines_r2}. We use the full review and target segment as input for the `RT' setup and only the target segment for the `T' setup as described in Sec~\S\ref{sec:experiments}. The prompt templates for In-Context Learning as used in Round 3 of our annotation study, shown in Figures~\ref{fig:prompt_fine_icl} and \ref{fig:prompt_coarse_icl} for fine-grained and coarse-grained classification, respectively. 

\subsection{In-context Learning Experiment for Round 3} \label{sec:incontext_app}
We use different types of methods to generate in-context learning exemplars, namely Static, Random, Top-K, and Vote-K, as described in Sec~\S\ref{sec:incontext}. We use 1,2, and 3 exemplars for this experiment. The performance of the models for the fine-grained and coarse-grained classification are shown in Fig~\ref{fig:models_icl} and \ref{fig:models_icl_coarse}, respectively. We observe that the performances donot change on increasing the number of exemplars or using different strategies. 

\subsection{Training Details to perform instruction tuning} \label{sec:training}
\subsubsection{Instructions} We use the same instructions from Round 2 of our annotation study due to higher inter-annotator agreement (Sec~\S\ref{sec:guidelines}). During fine-tuning, we follow the \textsc{SciRIFF} template using Jinja~\cite{jinja}, providing task descriptions and data as prompts, with output labels as responses. Two instruction formats are used: one with just the target segment (T) and another with both the target segment and full review (RT). The output is either a fine-grained \emph{lazy thinking} class or a coarse label (\emph{not lazy thinking} or \emph{lazy thinking}). Demonstration instances are excluded to ensure consistency with \textsc{SciRIFF} and \textsc{T\"ulu V2}.

\subsubsection{Hyper-parameters for LoRa training} \label{sec:hyper_parameters}
We train all the models for $3$ epochs. For LoRa, we use the rank of $64$, alpha of $16$, and dropout of $0.1$. The models are trained with a cosine learning rate scheduler and a warmup ratio of $0.03$. We use BF16 and TF32 to ensure training precision. The learning rate is set as $1e-4$.

\subsubsection{Hyper-parameter search} \label{sec:hyper-parameter}
We train models using various data proportions: 0.2, 0.4, 0.6, 0.8, and 1.0 from the different mixes. We then evaluate their performance on multiple validation sets. The performance results are plotted in two ways: using only the target segment (T) in Fig~\ref{fig:percent_data} and using a combination of review and target segment (RT) in Fig~\ref{fig:percent_data_rt} corresponding to the fine-grained classification task. For the T setup, where only the target segment is used, the models achieve optimal performance with 0.2 to 0.4 of the data. Performance tends to stagnate with more data indicative of reaching the maxima and is in line with the findings in \citet{wadden2024sciriffresourceenhancelanguage}. In the RT setup, where both the target segment and review are used, the models perform best with 0.6 to 0.8 of the data. This is because the combined prompts provide more context and take longer for the model to fully grasp, allowing it to reach peak performance with a larger dataset. As a result, we select \textbf{0.3} as the optimal data proportion for the T setup and \textbf{0.7} for the RT setup, balancing performance and avoiding the issues observed with different data sizes. We obtain similar findings for coarse-grained classification as shown in Fig~\ref{fig:percent_data_coarse} (`T') and Fig~\ref{fig:percent_data_rt_coarse} (`RT'), respectively.

\subsubsection{Instruction Tuning Training Setup for Annotation Rounds} 
We train models with the mixes identified in the experiments with 3-fold cross-validation for optimal performance. We use the same instructions that are used in the annotations and zero-shot prompting for finetuning the models. We use a data proportion of 0.3 to train models within the target segment-based prompting setup (T) and 0.8 for the RT (full review with target segment) setup from the best mizes identified fro each model. In line with the results on the test sets, we use \textsc{SciRIFF Mix} to train LLaMa and SciT\"ulu models, \textsc{T\"ulu Mix} for the Gemma and Qwen models and Full Mix for Mistral and Yi models.

\subsection{Confusion matrices for Annotation Labels and Confidences across annotation rounds} \label{sec:ann_conf}
\begin{figure*}[!htb]
\centering
    \begin{subfigure}[b]{0.25\textwidth}
        \includegraphics[width=\textwidth]{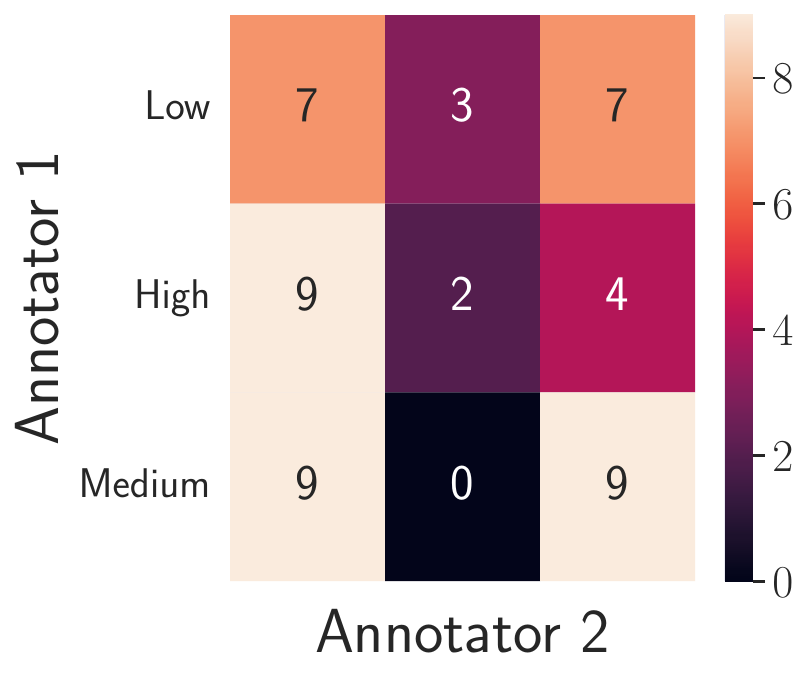}
        \caption{Round 1}
        \label{fig:r1_conf}
        \end{subfigure}%
        \hspace{1em}%
        \begin{subfigure}[b]{0.25\textwidth}
        \includegraphics[width=\textwidth]{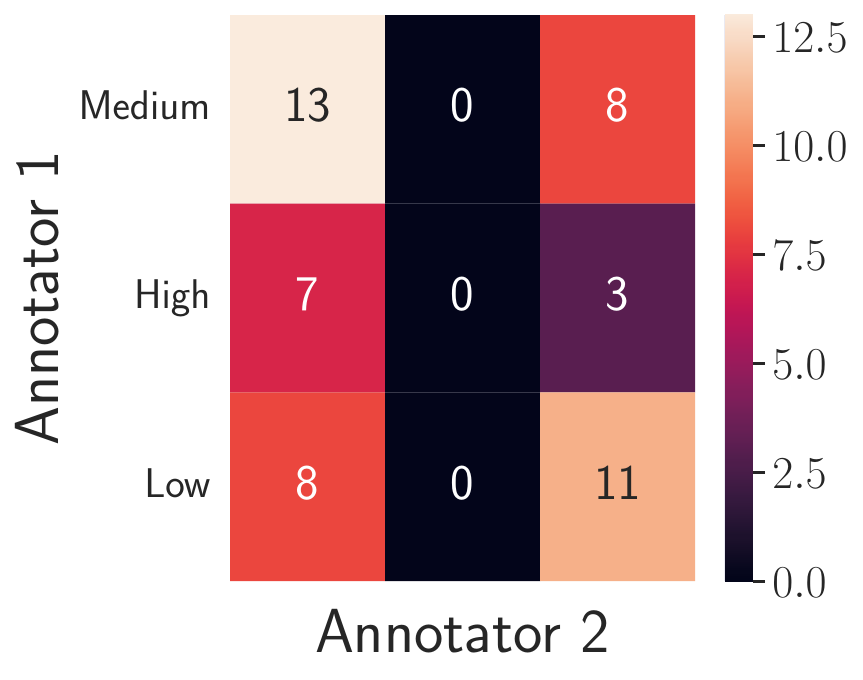}
        \caption{Round 2}
        \label{fig:r2_conf}
        \end{subfigure}%
         \hspace{1em}%
        \begin{subfigure}[b]{0.25\textwidth}
         \includegraphics[width=\textwidth]{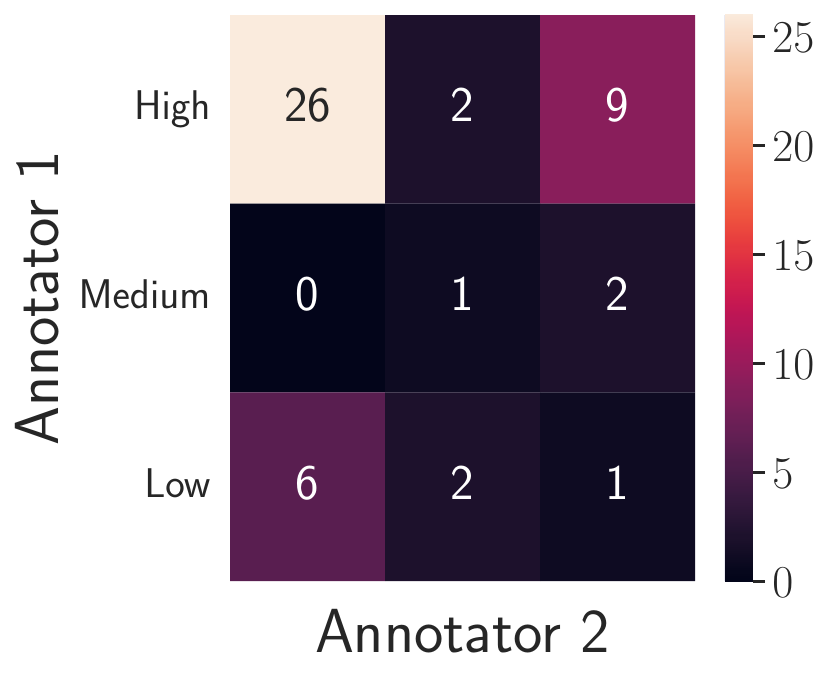}
        \caption{Round 3}
        \label{fig:r3_conf}
        \end{subfigure}

    \caption{Confusion matrices for the confidence levels among annotators across multiple rounds of annotation.}
    \label{fig:conf_matrices}
\end{figure*}

We show the confusion matrices for annotation labels and confidences in Figures~\ref{fig:round_annotation} and \ref{fig:conf_matrices} respectively.

\subsection{Results for the 3-fold cross validation} \label{sec:3_fold}
The results for the 3-fold cross-validation for fine-grained and coarse-grained classification is shown in Tables~\ref{tab:test_set} and \ref{tab:test_set_coarse} respectively. The random and majority baseline scores for fine-grained classification are $4.11_{5.2}$ and $6.7_{5.4}$ respectively. The random and majority baseline scores for coarse-grained classification are $44.11_{3.2}$ and $47.7_{3.2}$ respectively. Thus, all the models significantly outperform the majority and random baselines for both the tasks.

\subsection{Results for all the experiments across the annotation rounds} \label{sec:full_results} We report the full results for the fine-grained and coarse-grained classification in Tables~\ref{tab:ann_round_llm_fine-huge} and \ref{tab:ann_round_llm_coarse} respectively.

\subsection{Instructions to Annotators for Rewriting Reviews and Evaluation Protocol} \label{sec:instruction}
As discussed in Sec~\S\ref{sec:experiments}, we form two control groups to rewrite reviews. The first control group is asked to rewrite reviews based on the current ARR guidelines. While the other control group is also provided with \emph{lazy thinking} annotations. We provide both the control groups with the paper and the reviews corresponding to that paper. Finally, a senior Ph.D and PostDoc evaluate the rewrites pair-wise based on various measures.

\noindent \textbf{Instructions for re-writing based on current ARR guidelines.} Given the review and the corresponding manuscript, your task is to re-write the reviews to comply with the current ARR guidelines.\footnote{https://aclrollingreview.org/reviewertutorial}. Please feel free to edit or remove the content from any of the reviewing sections such as `summary of weaknesses', `comments, suggestions and typos' if applicable. 

\noindent \textbf{Instructions for re-writing based on current ARR guidelines and \emph{lazy thinking} annotations.} Given the review and the corresponding manuscript, you have been provided annotated instances of \emph{lazy thinking}.  The definition of lazy thinking is as follows `Lazy thinking, in the context of NLP research paper reviews, refers to the practice of dismissing or criticizing research papers based on superficial heuristics or preconceived notions rather than thorough analysis. It is characterized by reviewers raising concerns that lack substantial supporting evidence and are often influenced by prevailing trends within the NLP community.' According to ARR guidelines, reviewers are explicitly discouraged from using such heuristics in their review reports. In line with the ARR guidelines, your task is to re-write the reviews to comply with the guidelines. Please feel free to edit or remove the content from any of the reviewing sections such as `summary of weaknesses', `comments, suggestions and typos' if applicable. 

\noindent \textbf{Instruction for evaluating the rewrites.} You are provided a paper along with a pair of reviews written for the same paper. Your task is to perform pair-wise comparison of the reviews based on: \textbf{(1)} Constructiveness- reviewers should provide actionable feedback for the authors to improve on the paper, \textbf{(2)} Justified -
reviewers should clearly state the reason for their
arguments rather than putting out vague statements.
Additionally, we introduce a measure, \textbf{(3)}Adherence, which
judges how well the reviews are written based on
the current ARR guidelines.
\subsection{Distribution of review segment lengths in \textsc{LazyReview}} \label{sec:seg_length} 
\begin{figure}[!t]
\centering
        \includegraphics[width=\linewidth]{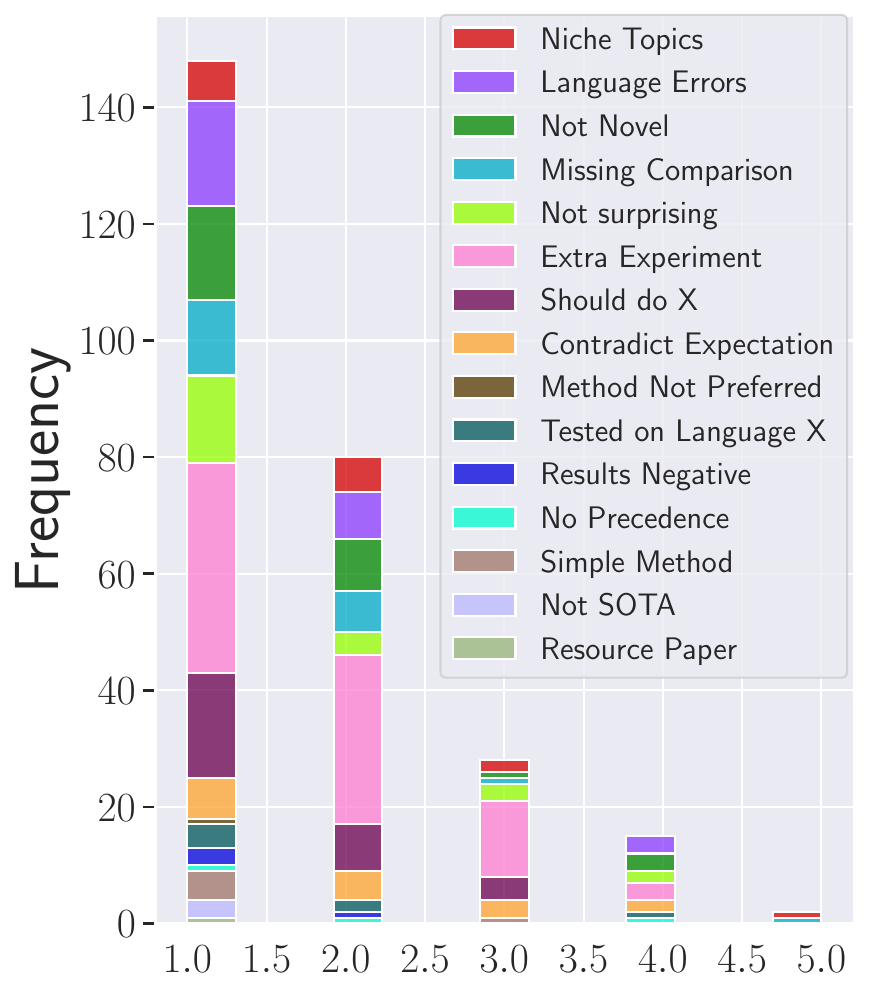}
        \caption{Segment lengths}
        \label{fig:sent_length}
\end{figure}

We plot the distribution of review segment lengths in Fig~\ref{fig:sent_length}. We find that `Extra Experiments' is the most common with variable segment lengths. 
\subsection{Analysis on the Silver Annotations} \label{sec:silver}
\begin{figure}[!t]
    \centering
    \includegraphics[width=\linewidth]{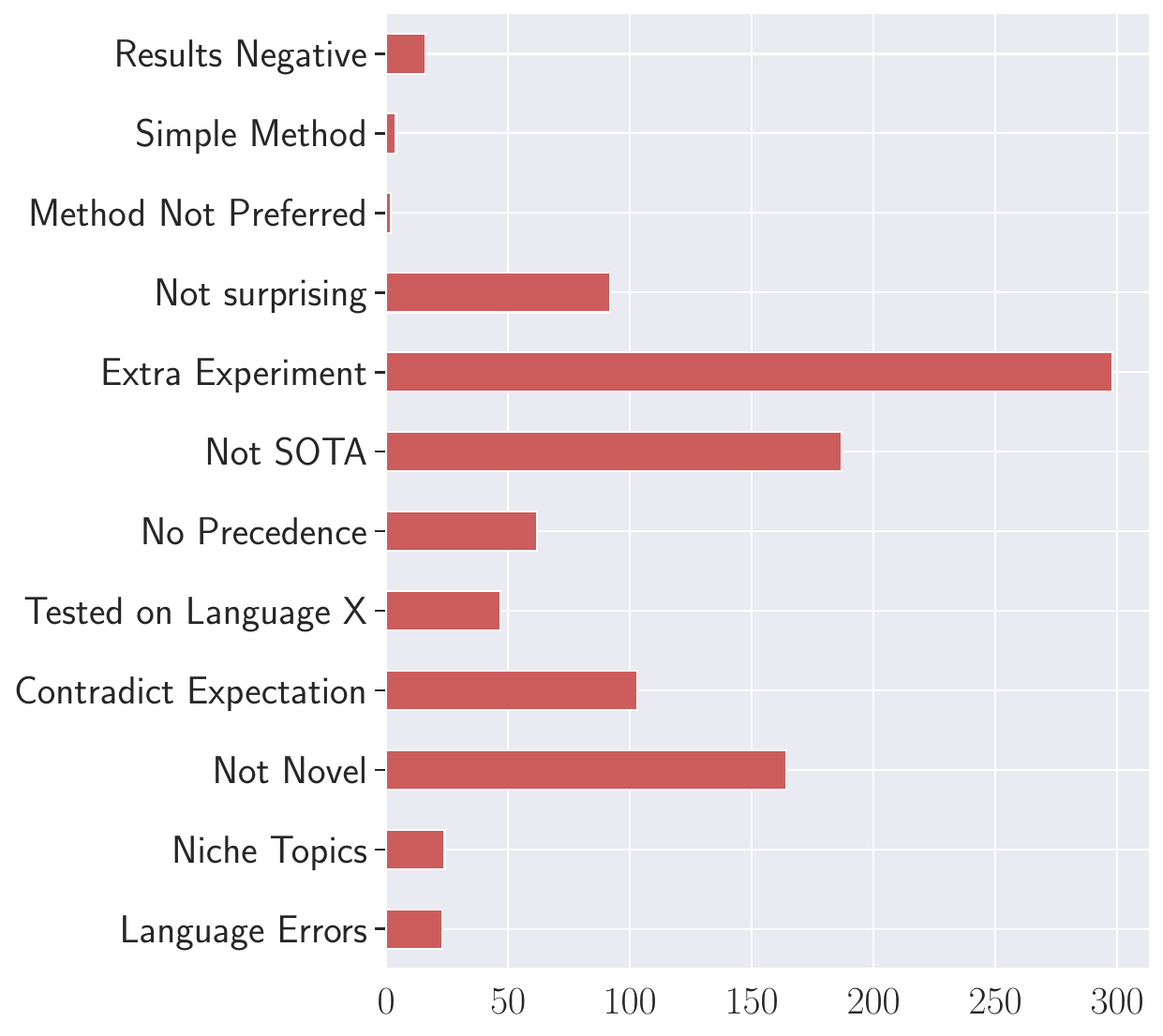}
    \caption{Distribution of \emph{lazy thinking} labels in the silver annotated data}
    \label{fig:silver_annotations}
\end{figure}

As discussed in RQ3 (cf. \S\ref{sec:experiments}), we use the best-performing model (Qwen) to annotate the remaining 1,276 review segments in our dataset, generating silver annotations. One of the annotators reviews a sample of 100 segments, yielding a Cohen's $\kappa$ of 0.56 with the silver annotations, which is substantial given the subjectivity of this task and the domain. The class label distribution is presented in Fig~\ref{fig:silver_annotations}, where `Extra Experiment' remains the most frequent \emph{lazy thinking} class. Additionally, we see increases in `Not SOTA' and `Not Novel' instances, likely reflecting the fast-paced advancements in ML that quickly render state-of-the-art methods obsolete~\cite{sevilla2022compute}. The notion of `novelty' is also frequently debated, with reviewers often using it to reject papers without sufficient justification~\cite{kumar2023novelty}. 
\subsection{Analysis on performance of model for different classes} \label{sec:further_analysis}
\begin{figure}
    \centering
    \includegraphics[width=0.6\linewidth]{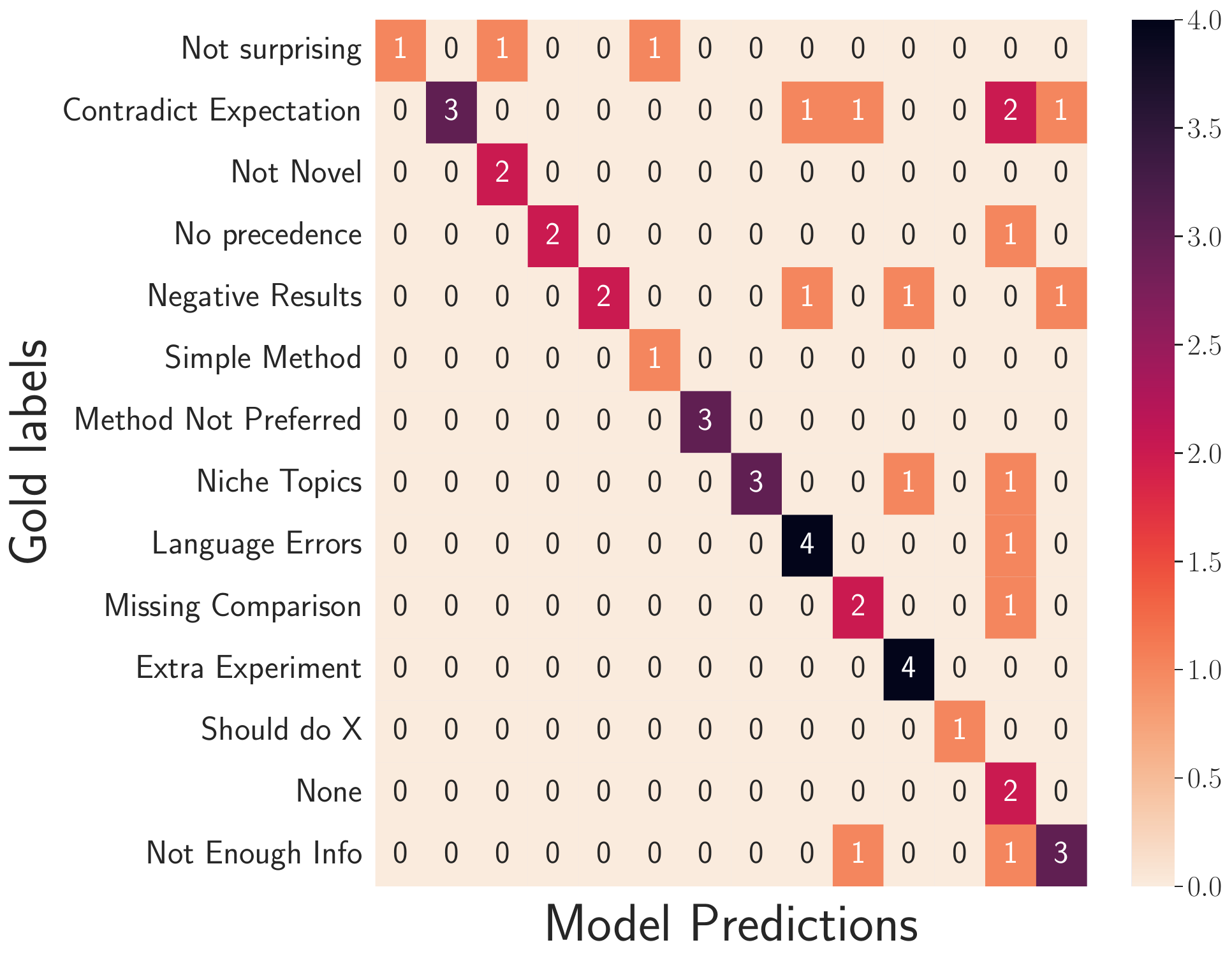}
    \caption{Performance analysis of Qwen on Round 3 annotation data}
    \label{fig:round3_model}
\end{figure}

\begin{figure*}[!htb]
\centering
    \begin{subfigure}[b]{0.3\textwidth}
        \includegraphics[width=\textwidth]{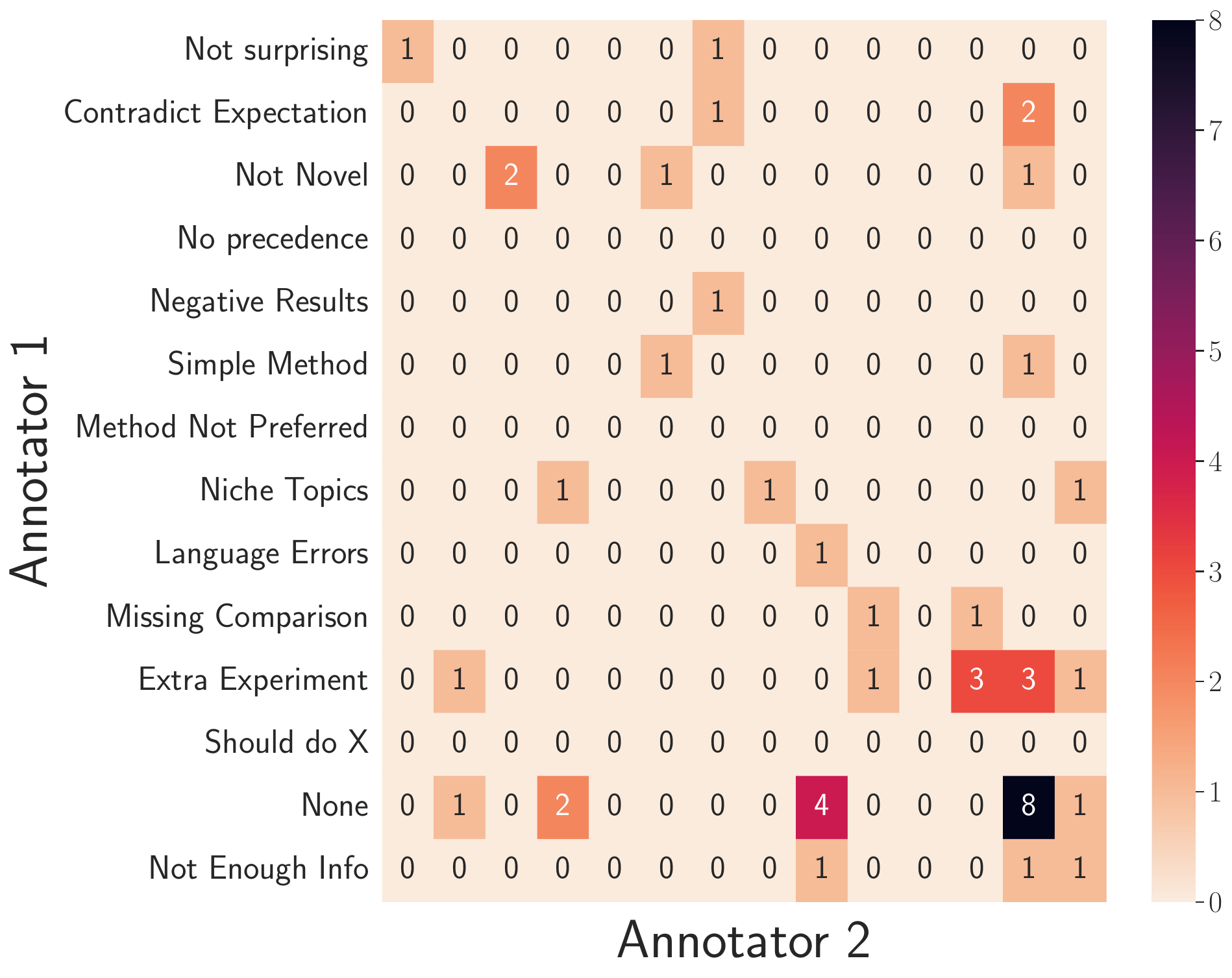}
        \caption{Round 1}
        \label{fig:r1_mat}
        \end{subfigure}%
        \hspace{1em}%
        \begin{subfigure}[b]{0.3\textwidth}
        \includegraphics[width=\textwidth]{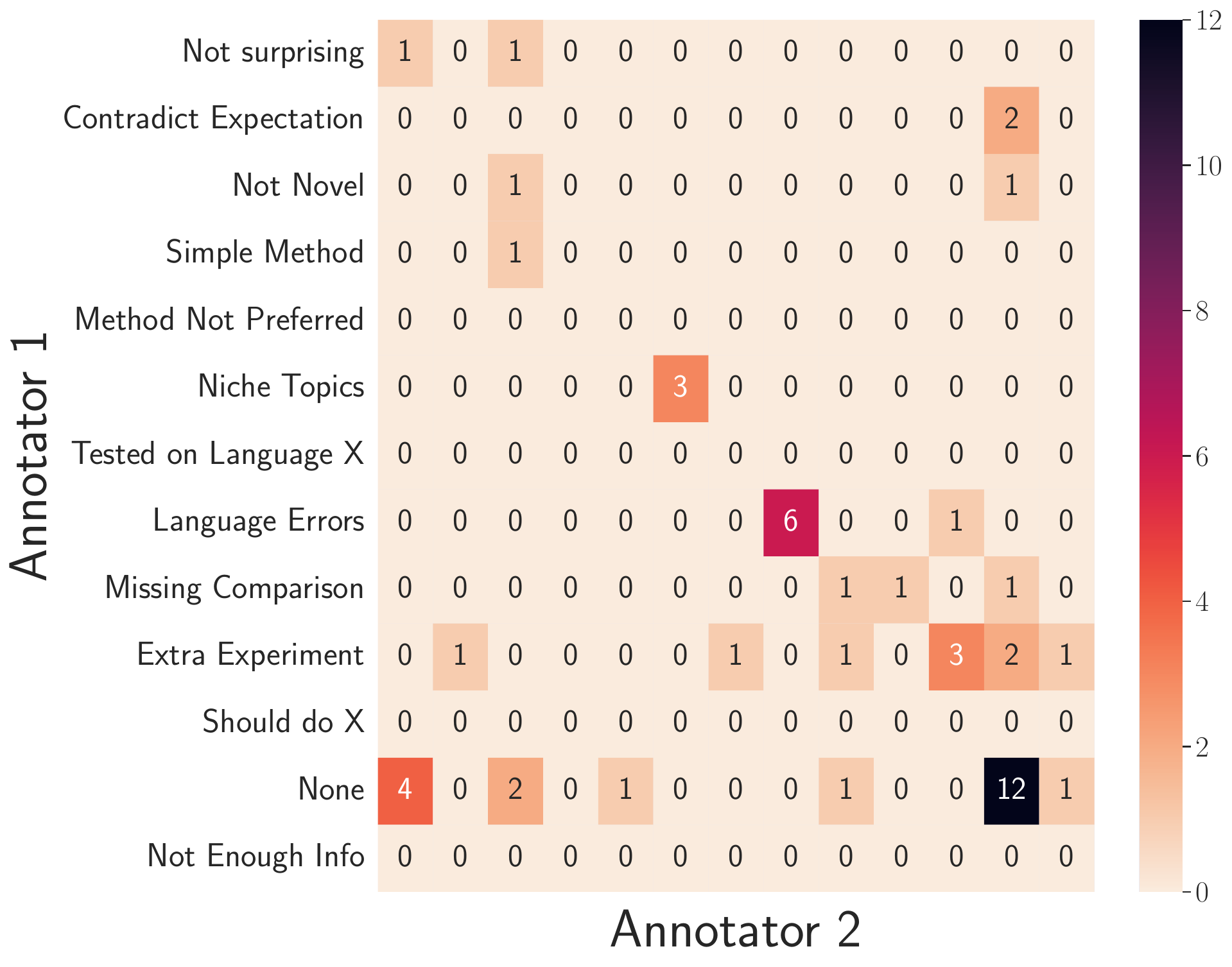}
        \caption{Round 2}
        \label{fig:r2_mat}
        \end{subfigure}%
         \hspace{1em}%
        \begin{subfigure}[b]{0.3\textwidth}
         \includegraphics[width=\textwidth]{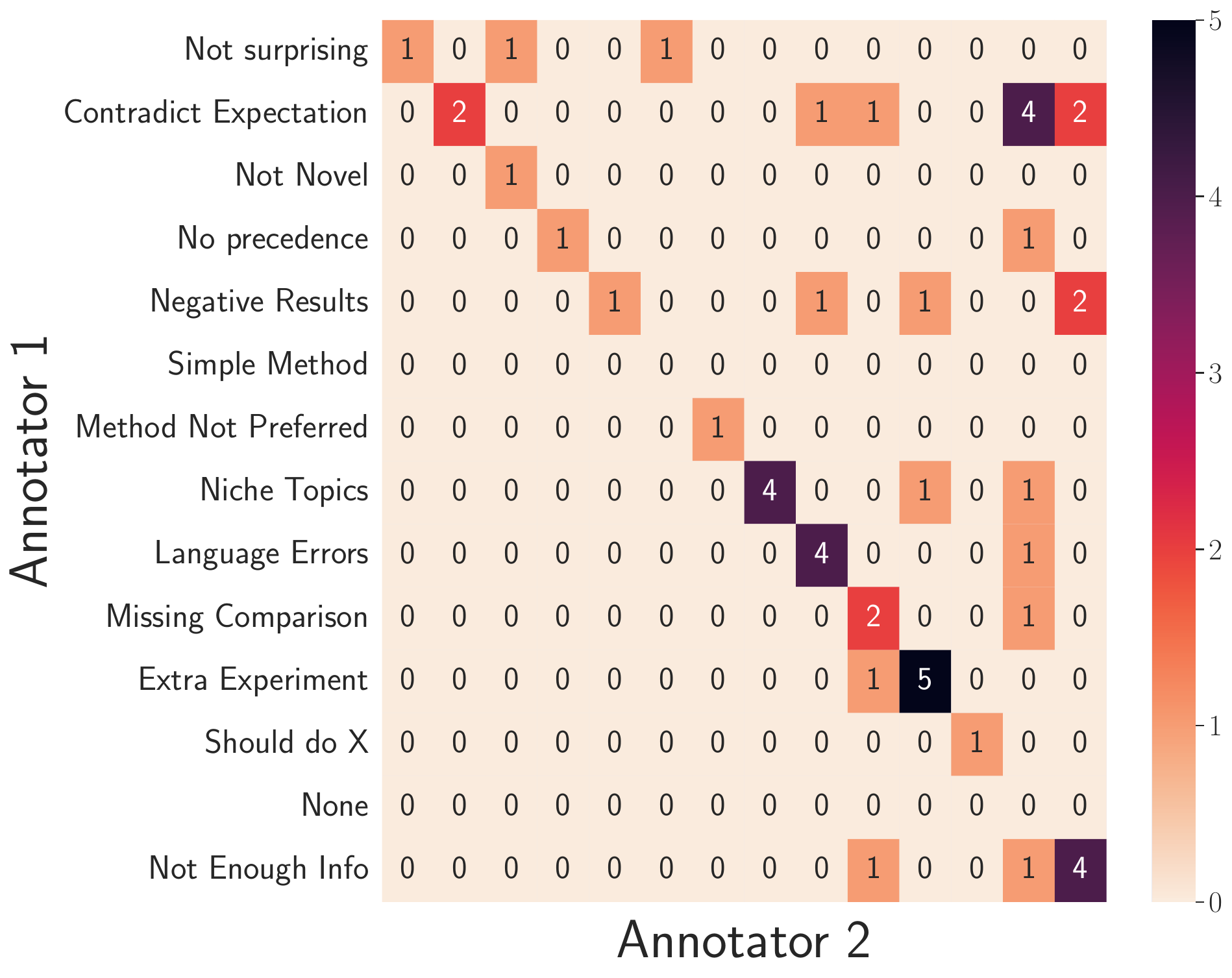}
        \caption{Round 3}
        \label{fig:r3_mat}
        \end{subfigure}

    \caption{Confusion matrices for the \emph{lazy thinking} classes from the ARR guidelines and EMNLP Blog used for annotation across multiple rounds. The class names are shortened due to space constraints.}
    \label{fig:round_annotation}
\end{figure*}

We analyze the performance of the best-performing model, Qwen, on the Round 3 annotations of our dataset. Since this portion yielded the highest inter-annotator agreement, we specifically focus on how the model’s predictions distribute across classes here. The confusion matrix is shown in Fig~\ref{fig:round3_model}. The model particularly struggles with the classes `Contradict expectations' and `Negative results', which mirror the difficulties observed in the human annotations from Round 3 (Fig~\ref{fig:r3_mat}). On the other hand, `Language errors' and `Niche topics' remain among the easiest classes for the automated methods to identify.


\end{document}